\PassOptionsToPackage{numbers, compress}{natbib}
\PassOptionsToPackage{table}{xcolor}
\documentclass{article}

\usepackage[preprint]{neurips_2026}

\usepackage[utf8]{inputenc}
\usepackage[T1]{fontenc}
\usepackage{hyperref}
\usepackage{url}
\usepackage{xurl}
\usepackage{booktabs}
\usepackage{amsfonts}
\usepackage{microtype}
\usepackage{xcolor}
\usepackage{array}
\usepackage{tabularx}
\usepackage{multirow}
\usepackage{makecell}
\usepackage{amsmath}
\usepackage{enumitem}
\usepackage{graphicx}
\usepackage{subcaption}  % for subfigure environment
\usepackage{caption}
\usepackage{comment}
\usepackage{colortbl} 
\usepackage{xcolor}
\usepackage{listings}
\usepackage{multicol}
\usepackage{wrapfig}
\usepackage{booktabs}
\usepackage{tabularx}
\usepackage{xcolor}
\usepackage{colortbl}
\usepackage{array}
\usepackage{booktabs}
\usepackage{tabularx}
\usepackage{pifont}
\usepackage{makecell}
\definecolor{gold}{RGB}{255, 215, 0}
\definecolor{silver}{RGB}{192, 192, 192}
\definecolor{bronze}{RGB}{205, 127, 50}
\definecolor{simhigh}{RGB}{198,227,190}
\definecolor{simmed}{RGB}{255,243,176}
\definecolor{simlow}{RGB}{255,205,186}
\definecolor{truegreen}{RGB}{198,227,190}
\definecolor{falsered}{RGB}{255,205,186}

\usepackage{booktabs} 
\usepackage{makecell}
\usepackage{tabularx}
\usepackage{xcolor}
\usepackage[table]{xcolor}
\definecolor{rowhl}{RGB}{235,245,255}
\newcolumntype{L}[1]{{\raggedright\arraybackslash}p{#1}}
\definecolor{rowhl}{RGB}{219,234,254}
\newcolumntype{L}[1]{>{\raggedright\arraybackslash}p{#1}}
\newcolumntype{C}[1]{>{\centering\arraybackslash}p{#1}}

\usepackage{booktabs}
\usepackage{changepage}
\usepackage{tabularx}
\usepackage{xcolor}
\usepackage{colortbl}
\usepackage{array}
\usepackage{pifont}
\usepackage{colortbl}
\usepackage{xcolor}
\definecolor{rowblue}{RGB}{219,234,254}
\definecolor{rowgreen}{RGB}{220,252,231}
\definecolor{rowamber}{RGB}{254,243,199}

\definecolor{rowhl}{RGB}{219,234,254}   % blue  — AssetOpsBench
\definecolor{yr24}{RGB}{254,243,199}    % amber — 2024 rows
\definecolor{yr25}{RGB}{220,252,231}    % green — 2025 rows

\definecolor{tableheadbg}{RGB}{28, 54, 94}
\definecolor{tableheadfg}{RGB}{255, 255, 255}
\definecolor{groupbg}{RGB}{240, 245, 252}
\definecolor{groupfg}{RGB}{28, 54, 94}

\definecolor{clrcomment}{HTML}{888780}
\definecolor{clrkw1}{HTML}{534AB7}
\definecolor{clrkw2}{HTML}{0F6E56}
\definecolor{clrfn1}{HTML}{185FA5}
\definecolor{clrfn2}{HTML}{0F6E56}
\definecolor{clrparam}{HTML}{993C1D}
\definecolor{clrbg}{HTML}{F7F6F3}
\definecolor{clrframe1}{HTML}{CECBF6}
\definecolor{clrframe2}{HTML}{9FE1CB}

\lstdefinestyle{track1}{
  basicstyle=\ttfamily\scriptsize,
  backgroundcolor=\color{clrbg},
  commentstyle=\color{clrcomment},      % removed \itshape
  keywordstyle=\color{clrkw1}\bfseries,
  stringstyle=\color{clrparam},
  breaklines=true,
  breakatwhitespace=false,              % prevents mid-token breaks
  frame=single,
  framerule=0.8pt, rulecolor=\color{clrframe1},
  aboveskip=0pt, belowskip=0pt,
  xleftmargin=6pt, xrightmargin=6pt,
}

\lstdefinestyle{track2}{
  basicstyle=\ttfamily\scriptsize,
  backgroundcolor=\color{clrbg},
  commentstyle=\color{clrcomment},      % removed \itshape
  keywordstyle=\color{clrkw2}\bfseries,
  stringstyle=\color{clrparam},
  breaklines=true,
  breakatwhitespace=false,              % prevents mid-token breaks
  frame=single,
  framerule=0.8pt, rulecolor=\color{clrframe2},
  aboveskip=0pt, belowskip=0pt,
  xleftmargin=6pt, xrightmargin=6pt,
}

\lstdefinestyle{track1freeze}{
  style=track1,
  keywordstyle=\color{clrcomment}\bfseries,
  framerule=0.6pt,
  rulecolor=\color{clrcomment},
}

\lstdefinestyle{track2freeze}{
  style=track2,
  keywordstyle=\color{clrcomment}\bfseries,
  framerule=0.6pt,
  rulecolor=\color{clrcomment},
}

\newcolumntype{Y}{>{\raggedright\arraybackslash}X}
\newcommand{\assetops}{\textsc{AssetOpsBench}}
\newcommand{\assetopslive}{\textsc{AssetOpsBench-Live}}
\newcommand{\code}[1]{\texttt{#1}}

\newcommand{\tmatch}{\textit{t-match}}

\definecolor{headerblue}{RGB}{50, 102, 173}
\definecolor{rowblue}{RGB}{230, 241, 251}
\definecolor{rowgreen}{RGB}{225, 245, 238}
\definecolor{rowpurple}{RGB}{238, 237, 254}
\definecolor{headertext}{RGB}{255, 255, 255}
\definecolor{shadeA}{RGB}{245, 247, 250}
\definecolor{shadeB}{RGB}{237, 242, 248}
\definecolor{accentblue}{RGB}{31, 78, 140}
\definecolor{cgreen}{RGB}{198,227,190}
\definecolor{cpink}{RGB}{255,182,182}
\newcommand{\cyes}{\cellcolor{cgreen}1}
\newcommand{\cno}{\cellcolor{cpink}0}
\newcommand{\chal}{\cellcolor{cpink}1}
\newcommand{\cnoh}{\cellcolor{cgreen}0}
\definecolor{rowred}{RGB}{254,226,226}

\title{Results and Retrospective Analysis of the CODS 2025 \assetops{} Challenge}

\author{%
  Dhaval Patel$^{1}$\thanks{ Corresponding author  \texttt{pateldha@us.ibm.com}} \quad
  Chathurangi Shyalika$^{2}$ \quad
  Suryanarayana Reddy Yarrabothula$^{3,4}$ \\
  \bfseries
  Ling Yue$^{5}$ \quad
  Shuxin Lin$^{1}$ \quad
  Nianjun Zhou$^{1}$ \quad
  James Rayfield$^{1}$\\[1pt]
  $^{1}$IBM  ~ ~ $^{2}$Artificial Intelligence Institute at University of South Carolina \\
  $^{3}$Steel Authority of India Limited ~ ~  
  $^{4}$Indian Institute of Technology, Bhilai \\
  $^{5}$Rensselaer Polytechnic Institute  ~~
}
\begin{document}

\maketitle

\begin{abstract}
We present a retrospective analysis of the CODS 2025 \assetops{} 
challenge. The challenge evaluated multi-agent AI systems on 
long-horizon Industry 4.0 tasks under hidden-scenario, 
privacy-preserving conditions. Submitted agents operated through 
the entire Sensing $\rightarrow$ Reasoning $\rightarrow$ Actuation 
pipeline, with separate tracks isolating planning and execution 
capabilities. Despite the specialist expertise typically required 
in this domain, the registration artifact records 349 declared 
member slots across 149 teams, and the server log records 300 
submission attempts, 234 of which reached \texttt{Finished} status. 
The majority came from undergraduate teams and early-stage 
startups. We analyze the submission corpus along five 
complementary dimensions that aggregate leaderboard standings 
alone cannot address: participation, submission behavior, ranking 
robustness, computational cost, and strategy attribution. The 
analysis surfaces concrete weaknesses in composite-metric design, 
public-to-hidden rank alignment, and ranking stability. Most 
strikingly, public and hidden execution scores fail to correlate 
($\rho = -0.13$, $n=13$, $p=0.71$), indicating that public 
standing does not predict hidden robustness. A trustworthy-benchmark 
checklist published after the challenge independently validates 
most of our infrastructure by design and flags precisely the 
scorer-robustness gaps we surface. We release the scenarios and 
scoring traces and distill the analysis into portable diagnostics 
for future agentic benchmarks.
\end{abstract}

\section{Introduction}
\label{sec:intro}

%Recent advances in LLM-based agents have produced systems 
%capable of complex, multi-step industrial tasks through 
%reasoning, tool use, and multi-agent coordination. Yet moving 
%these systems from laboratory settings to real-world 
%deployment has made \textit{evaluation itself a central 
%scientific challenge}. The behaviors that matter most in 
%deployment are the hardest to measure: tool-use robustness, 
%privacy-preserving execution, and multi-step orchestration 
%are difficult to benchmark cheaply, hard to release publicly, 
%and highly sensitive to metric design. A competition that 
%gets metric design wrong can reward superficial prompt 
%engineering while leaving the harder problems unmeasured.

Recent advances in LLM-based agents have produced systems 
capable of accomplishing complex, multi-step industrial tasks through 
reasoning, tool use, and multi-agent coordination. However, moving 
these systems from laboratory settings to real-world 
deployment has made \textit{evaluation itself a central 
scientific challenge}. This challenge is amplified by the limitations of benchmark-style evaluation, which can misrepresent real-world capability by favoring narrowly specified and easily optimized tasks~\citep{bandelgeneral, kapoor2026openworld}. The 
behaviors that matter most in deployment are the hardest to 
measure such as tool-use robustness, privacy-preserving execution, 
and multi-step orchestration are difficult to benchmark 
cheaply, hard to release publicly, and highly sensitive to 
metric design. A competition that gets the metric design wrong 
can reward superficial prompt engineering while leaving the 
harder problems unmeasured.

\begin{figure*}[t]
  \centering
  \includegraphics[width=\linewidth]{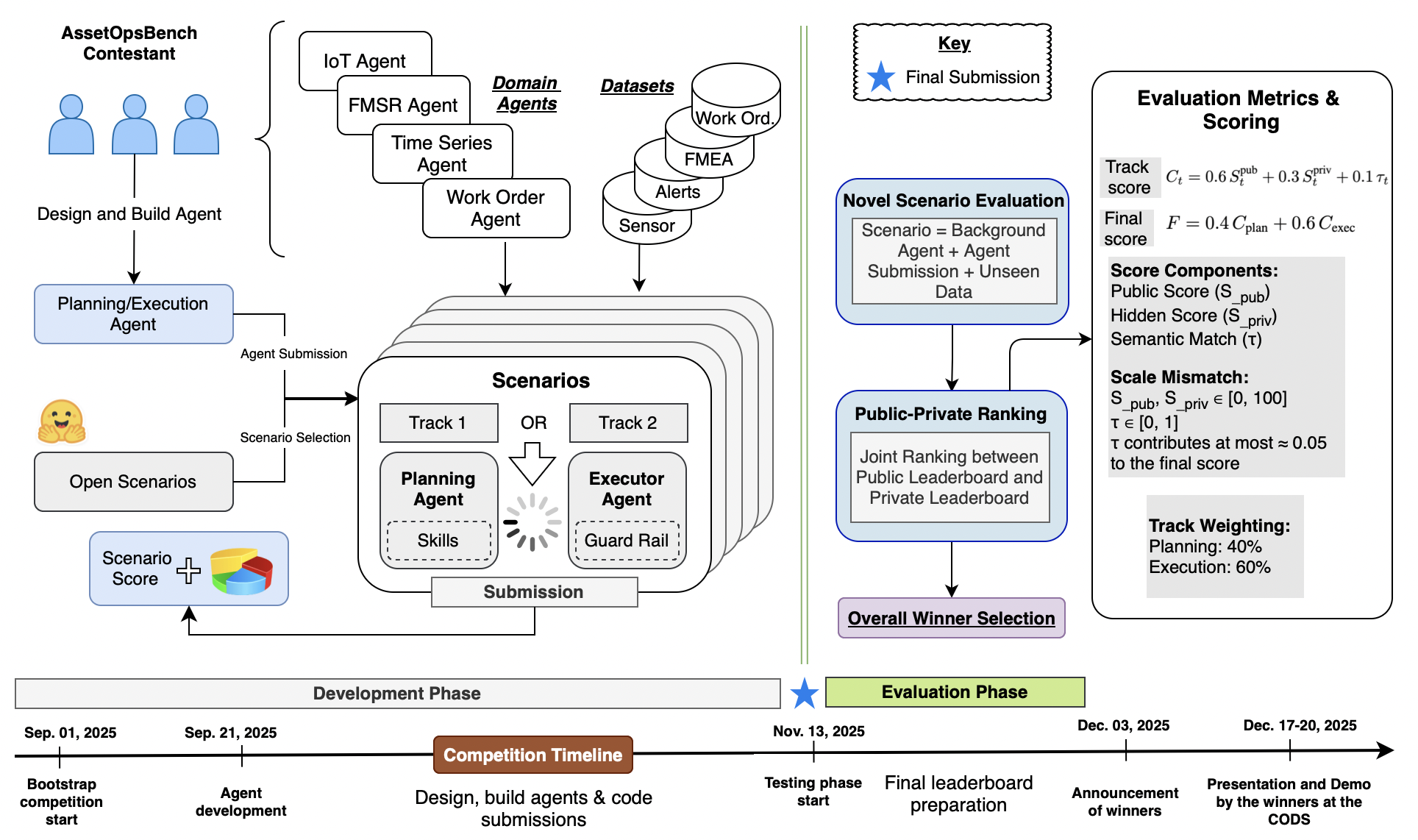}
  \caption{%
    CODS~2025 \assetops{} competition framework. Submissions are evaluated across Planning and Execution tracks against four domain agents on multimodal industrial data. The blue star marks the transition from
  the open Development to the hidden Evaluation phase.%
  }
  \vspace{-0.2in}
  \label{fig:framework}
\end{figure*}

Competition-based evaluation offers a principled alternative. 
By combining blind submission, hidden test sets, and 
large-scale participation, competitions expose failure modes 
that static benchmarks miss, including progressive reasoning 
failures and adaptive strategies that emerge only under 
iterative evaluation~\citep{debenedetti2024dataset}. 
Recent competition-driven benchmarks show that evaluation design is central, enabling reliability-aware datasets~\citep{bhimji2025fair}, large-scale stress testing~\citep{dingtechnical}, and contamination-resistant protocols~\citep{balunovicmatharena}. Previous retrospectives in nearest-neighbor search~\citep{simhadri2024results}, theorem proving~\citep{tsoukalas2024putnambench}, power systems~\citep{Marot2021LearningTR,yagoubiml4cfd}, systems neuroscience~\citep{turishcheva2024retrospective}, and security~\citep{debenedetti2024dataset} show that the most valuable competition papers explain what a leaderboard measures, not just who placed first. We follow this principle by analyzing a large-scale agentic AI competition in an industrial setting, asking what this evaluation design measured, what it failed to measure, and how its incentives shaped the submitted systems.

The CODS~2025 \assetops{} challenge is, to our knowledge, 
the first competition-track benchmark to combine agentic 
evaluation, an industrial physical-asset domain, and 
privacy-constrained deployment. It was hosted at the 
Conference on Data Science \& Management 
(CODS-COMAD)~\citep{cods2025}, one of Asia's premier data 
science venues. The challenge builds on two prior works: \assetops{}, an
industrial benchmark spanning predictive maintenance, fault diagnosis, work-order generation, 
and root-cause analysis for physical assets such as chillers and 
air-handling units~\citep{patel2025assetopsbench}; and 
\assetopslive{}, which extends this benchmark by deploying it as 
a privacy-preserving Codabench competition with six-dimensional 
LLM-as-judge scoring and clustered failure-mode 
feedback~\citep{cemri2025multi, patel2026assetopsbench}. 
Figure~\ref{fig:framework} presents the competition framework, 
highlighting the four domain-specific agents (i.e., IoT, FMSR, Time Series, Work Order) and the end-to-end 
evaluation timeline. Hosted in Codabench~\citep{xu2022codabench}, 
the organizer artifacts analyzed in this paper include 149 registered
teams, 349 declared member slots, and a 300-row submission-attempt log
across the two tracks (Planning and Execution). The best selected entries of development phase
were then evaluated under blind conditions on hidden industrial
scenarios. Codabench designated it as a spotlight competition in its yearly 
newsletter~\citep{codabench2025newsletter}, reflecting the 
reliability and scale of the evaluation infrastructure.

We therefore treat the challenge as a competition retrospective, 
not just as a leaderboard report. Our analysis combines the final 
rank sheets, a 300-attempt server log, 149-team registration 
forms, best-submission exports, scoring traces, and available 
top-submission artifacts. These artifacts support four concrete 
observations. The public planning leaderboard saturates at 
$72.73\%$. Public and hidden execution scores fail to correlate 
($\rho = -0.13$, $n=13$, $p=0.71$), so the public signal does 
not predict hidden robustness. The released \tmatch{} term 
contributes at most $0.05$ points per track because it is 
combined on a different numerical scale. The strongest execution 
systems are guardrail engineers, not architectural innovators as  
they improve response selection, cleanup, fallback, and context 
control rather than introducing new agent architectures. The rest of the paper describes the competition setup, the resulting leaderboard behavior, and the design lessons these outcomes suggest for future agent competitions.

\section{Competition Overview}
\label{sec:setup}

% ── Figure 1: Competition framework ──────────────────────────

% ── 2.1 Competition Design ───────────────────────────────────
\subsection{Challenge Design and Tracks}
\label{sec:design}

% Deploying LLM-based agents in industrial settings requires two capabilities that are easy to conflate but hard to improve jointly: the ability to \textit{plan} a coherent multi-agent workflow from a task description, and the ability to \textit{execute} that workflow reliably under real tool-use conditions. The CODS~2025 \assetopslive{} challenge was designed to measure each capability independently by holding the other fixed.

This section describes the key building blocks of the
competition framework shown in Figure~\ref{fig:framework}.
The public \assetops{} benchmark comprises 141 industrial
scenarios spanning 99 single-agent and 42 multi-agent cases~\citep{patel2025assetopsbench}. The scenarios are
hosted on Hugging Face~\citep{assetopsbench2026} and serve
as the shared evaluation set for all submissions. The four
domain-specific agents (IoT, FMSR, TSFM, and WO),
together with their associated multimodal datasets, are
packaged inside a Docker container \citep{assetopsbench_docker}, ensuring that every
participant executes against an identical, controlled
environment regardless of local infrastructure.

The competition was hosted on Codabench~\citep{assetopsbench_codabench_competition}, with technical documentation
and prebuilt Docker images released through a public starter
repository~\citep{assetopsbench_codabench2026}. This setup supported
long-running agent executions while keeping industrial data and hidden
scenarios within the evaluation environment.

\begin{figure}[h!]
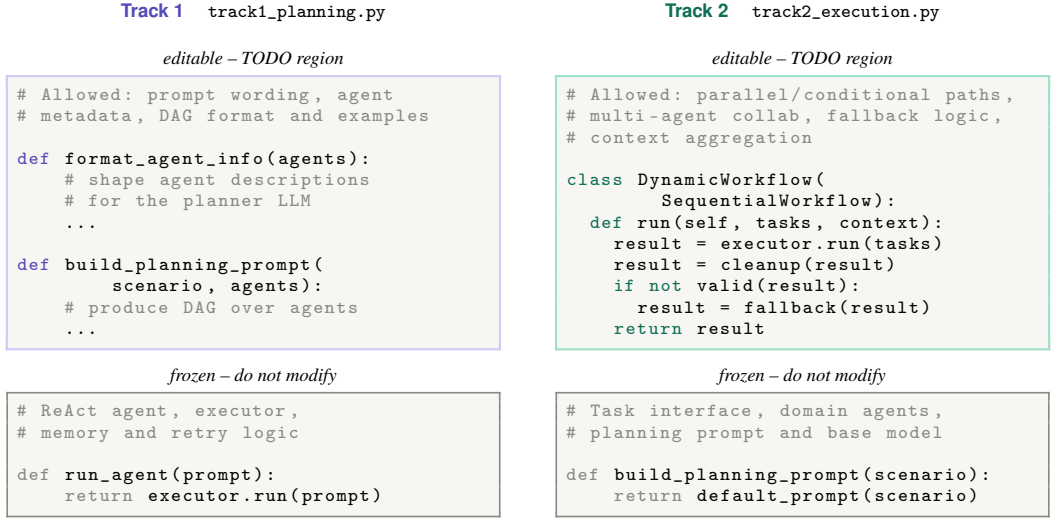

\vspace{-0.1in}
\begin{minipage}[t]{0.48\linewidth}
  \centering
  {\scriptsize\sffamily\textcolor{clrkw1}{\textbf{Track~1}}
   \quad\texttt{track1\_planning.py}}\\[4pt]
  \lstset{style=track1}
  \begin{lstlisting}[language=Python,
     title={\scriptsize\textit{editable -- TODO region}}]
# Allowed: prompt wording, agent
# metadata, DAG format and examples

def format_agent_info(agents):
    # shape agent descriptions
    # for the planner LLM
    ...

def build_planning_prompt(
        scenario, agents):
    # produce DAG over agents
    ...
  \end{lstlisting}
  \lstset{style=track1freeze}
  \begin{lstlisting}[language=Python,
     title={\scriptsize\textit{frozen -- do not modify}}]
# ReAct agent, executor,
# memory and retry logic

def run_agent(prompt):
    return executor.run(prompt)
  \end{lstlisting}
\end{minipage}
\hfill
\begin{minipage}[t]{0.48\linewidth}
  \centering
  {\scriptsize\sffamily\textcolor{clrkw2}{\textbf{Track~2}}
   \quad\texttt{track2\_execution.py}}\\[4pt]
  \lstset{style=track2}
  \begin{lstlisting}[language=Python,
     title={\scriptsize\textit{editable -- TODO region}}]
# Allowed: parallel/conditional paths,
# multi-agent collab, fallback logic,
# context aggregation

class DynamicWorkflow(
        SequentialWorkflow):
  def run(self, tasks, context):
    result = executor.run(tasks)
    result = cleanup(result)
    if not valid(result):
      result = fallback(result)
    return result
  \end{lstlisting}
  \lstset{style=track2freeze}
  \begin{lstlisting}[language=Python,
     title={\scriptsize\textit{frozen -- do not modify}}]
# Task interface, domain agents,
# planning prompt and base model

def build_planning_prompt(scenario):
    return default_prompt(scenario)
  \end{lstlisting}
\end{minipage}
\caption{%
  Editable \texttt{TODO} regions per track, mirroring the released
  starter templates}
  \vspace{-0.1in}
\label{fig:code_schematic}
\end{figure}

The two tracks create complementary controlled experiments within
this shared environment. \textbf{Track~1} holds the executor fixed
and asks the participants to improve the plan. Edits are restricted to
the region of prompt-construction and agent-formatting of 
\code{track1\_planning.py}. The core hypothesis is that better
prompts produce higher-quality Directed Acyclic Graphs (DAGs) over domain agents, and
higher-quality DAGs improve downstream execution regardless of
individual agent capability. \textbf{Track~2} holds the plan fixed
and asks participants to improve the executor. Edits are restricted
to the workflow-execution logic of \code{track2\_execution.py},
where the baseline \code{SequentialWorkflow} can be replaced with a
\code{DynamicWorkflow} supporting parallel execution paths,
multi-agent collaboration per task, cross-task context aggregation,
and fault-tolerant fallback. The domain agents, the base model, and
the planning prompt remain frozen. Figure~\ref{fig:code_schematic} shows
the exact editable and frozen regions for each track.

By design, the editable surface in Track~1 concentrates
the variation controlled by the participant in the prompt and planning code, while
Track~2 concentrates on workflow execution and context handling.
This separation is a central methodological feature of the competition,
although residual variation can still arise from packaging choices, submission practices, and scorer details.

All submissions use a fixed LLaMA-3-70B model and pass through three
evaluation stages: an optional local warm-up on 2--3 scenarios for
pipeline validation, Phase~1 (Development) on 11 scenarios drawn from the public
pool of 141, and a Phase~2 (Evaluation) generalization test on 11 novel
scenarios from unseen asset classes. The Per-track ($C_t$) scores combine a
public component, a hidden component, and a semantic \tmatch{}
signal:
\begin{equation}
C_t = 0.6\,S^{\text{pub}}_t + 0.3\,S^{\text{priv}}_t + 0.1\,\tau_t,\quad
F = 0.4\,C_{\text{plan}} + 0.6\,C_{\text{exec}},\quad
\Delta = S_{\text{priv}} - S_{\text{pub}}
\label{eq:final_formula}
\end{equation}
where $t \in \{\text{plan},\text{exec}\}$, the semantic t-match score ($\tau_t$), $F$ is the final ranking score, and $\Delta$ denotes the difference between the public and hidden scores. Execution ($C_{\text{exec}}$) receives greater weight (60\%) than planning ($C_{\text{plan}}$)
(40\%), reflecting the organizers' view that robust execution under
real tool-use conditions is the harder and more deployment-relevant
challenge. We return to both design choices in Section~\ref{sec:scores}, where the sensitivity of the scoring formula to the \tmatch{} scale becomes relevant. The organizers selected each team's best-scoring public leaderboard submission per track as the canonical entry for hold-out evaluation. Full details of the evaluation
scenarios are given in the Appendix~\ref{sec:app:scenarios}.

% ── Figure 2: Code schematic ─────────────────────────────────

% ── 2.2 Artifacts, Evidence Levels, and Data Cleaning ────────
\vspace{-0.1in}
\subsection{Artifacts, Analysis Data, and Counting Conventions}
\label{sec:artifacts}
The competition produced six interlocking artifact classes that 
together reconstruct every stage. These include registration, submissions, 
scores, team selection, and solution code. The \textbf{participation 
data} covers 149 teams, with the registration forms 
recording the composition of the team and the institutional affiliation. 
\textbf{Submission records} comprise a 300-row server log with 
timestamps, status, public scores, agent trajectories, and 
usernames linked to team identities. The \textbf{scoring data} 
includes best-submission exports and the official ranking 
workbook, which exposes hidden scores, \tmatch{} values, and the 
aggregation formula. \textbf{Qualitative evidence} from the 
organizers' award report contextualizes top-team selection.
The \textbf{platform documentation}, covering the public challenge 
pages and starter-kit guidelines, anchors all claims about 
allowed edits to the instructions the participants received. 
\textbf{Code-level evidence} provides ground truth on what the top 
teams implemented. The spreadsheets require light cleaning. We normalize case  variants (e.g., \code{Infinity}/\code{infinity}), drop blank rows, 
retain the latest registration form per team, and resolve one 
repeated planning entry from same team by keeping the higher 
public score. 

\begin{comment}
Because these artifacts were produced under competition conditions 
rather than curated for research, their reliability varies. We 
therefore distinguish three evidence levels throughout this paper. 
\textit{Verified code} means we directly inspected the released 
source tree. \textit{Organizer summary} means the method 
description comes from organizer notes without a matching 
accessible source tree. \textit{Incomplete archive} means a source 
tree exists but does not appear to be the exact evaluated 
submission. Whenever sources disagree, the released final 
spreadsheets are treated as authoritative for all quantitative 
ranking analyzes.
\end{comment}

\paragraph{Counting conventions.}
We keep distinct denominators separate throughout the paper.
\emph{Declared member slots} are person-level entries on the
registration forms; \emph{registered teams} are the 149
team records after keeping the latest form per team; and
\emph{platform participants} are public Codabench-level metadata used
only for cross-competition scale comparisons. A \emph{submission attempt} is
one row in the 300-row server log, while a \emph{Finished submission}
is one of the 234 attempts that completed platform evaluation. The
leaderboard and hold-out analyzes use selected best team-track
submissions, not all attempts. The method taxonomy uses 331 accessible
source-level artifacts, which are code artifacts available for
strategy analysis and are not treated as additional server-log
attempts. Cost and failure analyzes operate on trajectory files, where
one file records a per-scenario execution trace.

\section{Competition Results and Retrospective Analysis}

\vspace{-0.1in}
\subsection{Participation Funnel, Platform Realities, and Final Leaderboard.}

\begin{wraptable}{r}{0.44\textwidth}
\centering
\vspace{-11pt}
\caption{Participation funnel statistics.}
\label{tab:participation}
\small
\setlength{\tabcolsep}{4pt}
\renewcommand{\arraystretch}{1.2}
\begin{tabular}{@{} l r @{}}
\toprule
\textsc{Statistic} & \textsc{Value} \\
\midrule
Registered teams          & 149 \\
Member slots declared     & 349 \\
Undergraduates            & 45.8\% \\
Industry professionals    & 27.8\% \\
Master's / PhD            & 22.3\% \\
Other                     & 4.0\% \\
Multi-username teams      & 78/149 (52.3\%) \\
\midrule
Logged attempts           & 300 \\
\ding{52}~Finished        & 234 (78.0\%) \\
\ding{55}~Failed          & 53 (17.7\%) \\
\ding{109}~Cancelled      & 9 (3.0\%) \\
\ding{110}~In progress    & 4 (1.3\%) \\
\midrule
Non-zero public-score teams & 24 + 1 anon \\
Fully ranked teams        & 11 \\
Cross-track accounts      & 4/11 \\
\bottomrule
\end{tabular}
\vspace{-25pt}
\end{wraptable}

Competition registration and ranking required clearing three independent thresholds, submitting a registration form, producing a valid scored submission, and populating both tracks. As
shown in Table~\ref{tab:participation}, of 149 registered teams, 24
cleared the second threshold and 11 cleared the third. This pattern of attrition
 is not incidental; it is a direct empirical measure of the
cost of platform-mediated agent evaluation in practice.

\paragraph{Funnel and friction.}
The registration required two steps, a Google Form for team metadata
and individual Codabench enrollment per member. Over seven and a
half weeks (2025-09-21 to 2025-11-13), 300 submission attempts were
recorded across the two tracks. Of these, \textbf{234 (78.0\%)
finished}, \textbf{53 (17.7\%) failed to pass conformance checks},
\textbf{9 (3.0\%) were cancelled}, and 4 (1.3\%) remained in
progress at the close of the competition. The 17.7\% failure rate is a
direct measurement of \textbf{platform-conformance cost}. The submissions rejected for packaging or workflow-format violations
consume attempts from the 50-trial per-team budget without
producing evaluable agent output. This is a fixed infrastructure
overhead distinct from the agent capability and establishes a
lower bound on the submission budget required for a team to
produce any scored agent at all. To redistribute this cost from
the evaluation time to the preparation time, the organizers provided two
local dry-run scenarios executed in identical log formats,
whose effect is visible in the low cancelation rate (3.0\%).

\begin{figure*}[h]
\centering
\begin{subfigure}[t]{0.48\textwidth}
    \includegraphics[width=\textwidth]{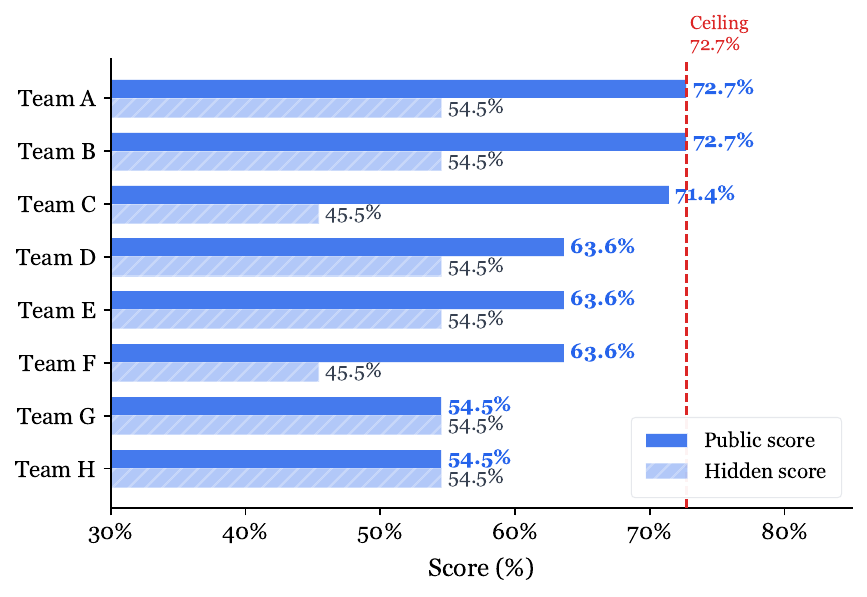}
    \caption{Track~1 (Planning) public leaderboard. Scores saturate
    at $72.73\%$ across top teams.}
    \label{fig:leaderboard_t1}
\end{subfigure}
\hfill
\begin{subfigure}[t]{0.48\textwidth}
    \includegraphics[width=\textwidth]{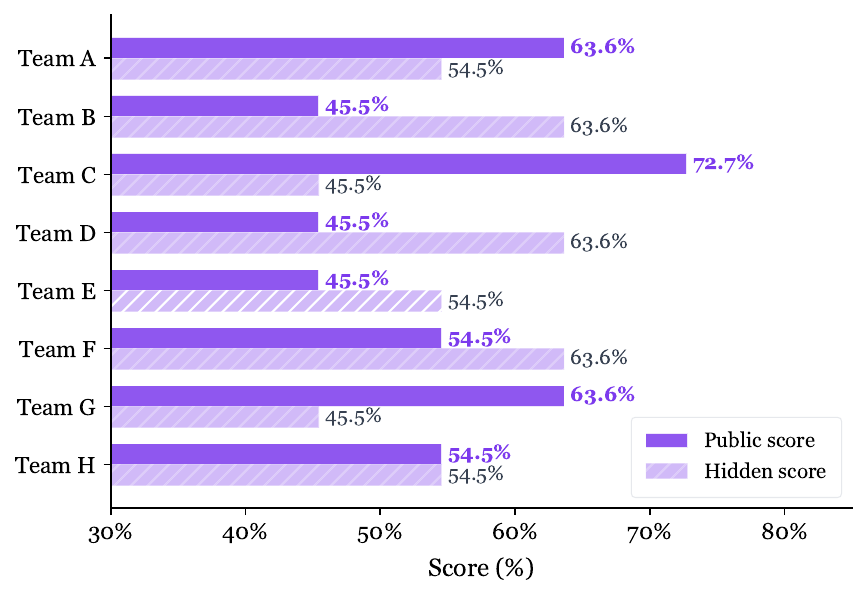}
    \caption{Track~2 (Execution) public leaderboard. Low public--hidden
    agreement.}
    \label{fig:leaderboard_t2}
\end{subfigure}
\caption{\textbf{CODS 2025 \assetops{} leaderboards.} Full rankings
in Table~\ref{tab:fullranking}.}
\label{fig:leaderboards}
\end{figure*}

\paragraph{Account identity as an evaluation variable.}
More than half of registered teams (78/149, 52.3\%) list multiple
Codabench usernames, with a mean of 2.21 accounts per team, and
four of the 11 ranked teams use distinct accounts across tracks.
The competition is team-based at the level of strategy; the
platform records at the level of individual accounts. Manual team
mapping in the released spreadsheets resolves this, but the
reconciliation is invisible to any analysis that consumes the
submission export without cross-referencing the registration
artifact. Submission identity, specifically \textit{who} submits
under \textit{which} account and \textit{when}, is a fairness and
reproducibility variable. Future competitions should either
enforce shared team accounts at the platform level or release
per-member submission attribution as an independently citable
artifact. Table~\ref{tab:fullranking} shows the finalized
ranking used for the official award decision, and Figure \ref{fig:leaderboards} shows the public leaderboard and private ranking for both tracks.

\begin{table*}[h]
\centering
\caption{Released final team ranking. Gold, silver, and bronze
rows indicate the top-three finishers.}
\small
\renewcommand{\arraystretch}{1.20}
\begin{tabular}{r l l l l r r r}
\toprule
\textbf{Rank} & \textbf{Label} & \textbf{Team} & \textbf{Planning owner}
  & \textbf{Execution owner}
  & $C_{\text{plan}}$ & $C_{\text{exec}}$ & \textbf{Final} $F$ \\
\midrule
\rowcolor{gold!60}
1  & C & Smart M. Crew & \texttt{vamsikv28}
   & \texttt{shashank\_1904} & 56.528 & 57.318 & 57.002 \\
\rowcolor{silver!50}
2  & A & WaterLevel         & \texttt{kanishk\_007}
   & \texttt{harshvardhan1}  & 60.049 & 54.593 & 56.775 \\
\rowcolor{bronze!40}
3  & F & LostSouls          & \texttt{h1t35h}
   & \texttt{h1t35h}         & 51.857 & 51.869 & 51.864 \\
4  & B & BlueCube           & \texttt{rohith\_arumu.~}
   & \texttt{samah}          & 60.049 & 46.406 & 51.863 \\
5  & G & Scalar\_nitk       & \texttt{scalar\_anjali}
   & \texttt{scalar\_anjali} & 49.141 & 51.863 & 50.774 \\
6  & D & Entropians         & \texttt{supminal}
   & \texttt{supminal}       & 54.580 & 46.408 & 49.677 \\
7  & H & Infinity           & \texttt{abhinf104}
   & \texttt{abhinf104}      & 49.137 & 49.139 & 49.138 \\
8  & E & aviation\_agent    & \texttt{shoeb}
   & \texttt{shoeb}          & 54.598 & 43.682 & 48.048 \\
9  & J & horizon            & \texttt{horizon22}
   & \texttt{horizon22}      & 32.768 & 57.323 & 47.501 \\
10 & I & kinatic            & \texttt{vinaykarman}
   & \texttt{subhadeep}      & 43.680 & 40.956 & 42.046 \\
11 & K & EXL Health         & \texttt{uthrasuresh}
   & \texttt{uthrasuresh}    & 32.767 & 43.680 & 39.315 \\
\bottomrule
\end{tabular}
\label{tab:fullranking}
\vspace{-0.1in}
\end{table*}

\paragraph{Participant heterogeneity as a validity consideration.}
The participant population is highly diverse, comprising 45.8\%
undergraduates, 27.8\% industry professionals, 22.3\% master's or
PhD students, and 4.0\% other, spanning 84 universities from the
host country (India), 8 international institutions, and 91 industry
organizations. The cohort distribution among the 11 ranked teams differs 
sharply from this overall composition. Industry-professional 
teams are over-represented at $54.5\%$ (6 of 11 ranked), 
nearly twice their share of registered teams; undergraduate 
teams are present but under-represented at $36.4\%$ (4 of 
11) relative to their $45.8\%$ pool share; and master's/PhD 
teams, despite forming $22.3\%$ of the registered pool, are 
\emph{absent} from the top 11 entirely. The top three 
finishers comprise two industry teams (Smart Maintenance 
Crew, LostSouls) and one undergraduate team (WaterLevel). This concentration is consistent with the strategy 
attribution in Section~\ref{sec:solutions}. Execution-track 
top performers favour deployment-style guardrail 
engineering. A leaderboard ranking across cohorts 
therefore measures not only agent capability but also the 
engineering practices each cohort brings to the competition. 
Future agentic competitions should report cohort 
stratification alongside leaderboard positions, and may need 
deliberate scenario design to surface academic-style 
contributions (e.g., novel architectures) that the current 
guardrail-rewarding scoring structure underweights.

\subsection{Public saturation, hidden-phase reordering, and
score-composition sensitivity}
\label{sec:scores}

The public leaderboard is substantially coarser than the methods
it ranks. Planning produces only eight distinct positive public
scores across 20 teams and saturates at $72.73\%$; execution
produces five distinct values across 13 teams (see
Figure~\ref{fig:leaderboards}). For the hold-out evaluation, we
selected each team's best public submission per track, yielding
20 planning and 13 execution hold-out evaluations.

\begin{wraptable}{r}{0.48\textwidth}
\centering
\vspace{-12pt}
\caption{Score dynamics. $N$: Number of teams, $\rho$ denotes Spearman rank correlation coefficient.}
\small
\setlength{\tabcolsep}{5pt}
\renewcommand{\arraystretch}{1.3}
\begin{tabular}{@{} l r r @{}}
\toprule
\textsc{Metric} & \textsc{Planning} & \textsc{Execution} \\
\midrule
\rowcolor{rowblue!20}
$N$                        & 20        & 13        \\
\rowcolor{rowblue!20}
Unique public scores       & 8         & 5         \\
\rowcolor{rowblue!20}
Mean public                & 54.03     & 54.55     \\
\rowcolor{rowblue!20}
Mean private               & 42.73     & 53.15     \\
\rowcolor{rowblue!20}
Mean $\Delta$              & $-11.30$  & $-1.40$   \\
\midrule
\rowcolor{rowgreen!20}
$\rho$(pub, priv)            & $0.69$    & $-0.13$   \\
\rowcolor{rowgreen!20}
$\rho$(pub, \tmatch{})       & $-0.01$   & $-0.22$   \\
\midrule
\rowcolor{rowamber!20}
Top public ties            & 4         & 1         \\
\rowcolor{rowamber!20}
Max rank shift             & 8         & 9         \\
\bottomrule
\end{tabular}
\label{tab:hidden_stats}
\vspace{-8pt}
\end{wraptable}

As shown in Table~\ref{tab:hidden_stats}, the two tracks exhibit
structurally different failure modes. In planning, the mean
private score falls 11.30 points below the mean public score
($\rho=0.69$), indicating moderate signal but systematic optimism.
In execution, the average drop is negligible ($-1.40$ points), 
yet the public--hidden correlation is statistically 
indistinguishable from zero ($\rho = -0.13$, $n=13$, $p=0.71$). Clearly,  public execution scores carry no signal about hidden 
performance. The rank-reversal pattern is consistent with this 
absence of signal. The public leader 
\texttt{Team~C} falls $72.73 \searrow 45.45$, while 
\texttt{Team~B} and \texttt{Team~D} rise $45.45 \nearrow 63.64$. 
Public iteration and hidden evaluation reward different 
behaviors. Appendix~\ref{sec:app:behaviour} expands this dimension along 
four axes along cross-track score distributions and specialization, temporal score 
progression, team activity 
patterns, and submission-level learning dynamics and reliability.

\begin{wraptable}{r}{0.55\textwidth}
\centering
\vspace{-10pt}
\caption{Ranking sensitivity to score-composition choices.}
\small
\setlength{\tabcolsep}{6pt}
\renewcommand{\arraystretch}{1.3}
\begin{tabular}{@{} l c c r @{}}
\toprule
\textsc{Setting} & \textsc{Top-1} & \textsc{Top-2} & \textsc{Margin} \\
\midrule
\rowcolor{rowblue!20}
Official release                 & Team C & Team A & 0.227 \\
No \tmatch{} term                & Team C & Team A & 0.230 \\
\rowcolor{rowamber!20}
Equal track weights              & Team A & Team C & 0.398 \\
\rowcolor{rowamber!20}
Rescaled \tmatch{} $(\times100)$ & Team A & Team C & 0.126 \\
\bottomrule
\end{tabular}
\label{tab:sensitivity}
\vspace{-4pt}
\end{wraptable}

The \tmatch{} term exposes a score-composition error. Public 
and private scores are on a $[0,100]$ scale, but \tmatch{} 
remains on $[0,1]$, making its effective contribution at most 
$0.05$ composite points per track, two orders of magnitude 
below the other terms. The nominal $10\%$ semantic weight is 
numerically inert. Rescaling \tmatch{} to percentage units 
would swap the top two teams and exchange third and fourth 
place (Table~\ref{tab:sensitivity}). Track weighting compounds this effect, as under equal weights \texttt{Team~A} would finish first. To assess the joint effect of both choices, we sweep the 
execution weight $\alpha$ and the \tmatch{} rescaling factor 
$s$ across 80 $(\alpha, s)$ configurations and record the 
top-ranked team for each (Appendix~\ref{sec:app:stability}, 
Figure~\ref{fig:winner_stability}). The official top-1 holds 
in only $44\%$ of configurations; the remaining $56\%$ crown 
a different team. Mean Kendall's $\tau$ between the official 
ranking and the alternatives is $\bar{\tau} = 0.61$ ($\sigma = 
0.19$), indicating moderate, not high, concordance. The 
official ranking therefore reflects two simultaneous 
methodological choices, each of which independently changes 
the top-two ordering, and a majority of plausible 
alternatives within the same scoring family produce a 
different winner.

\subsection{Agent-level Cost Fingerprint}

\begin{figure*}[t]
  \centering
  \includegraphics[width=\textwidth]{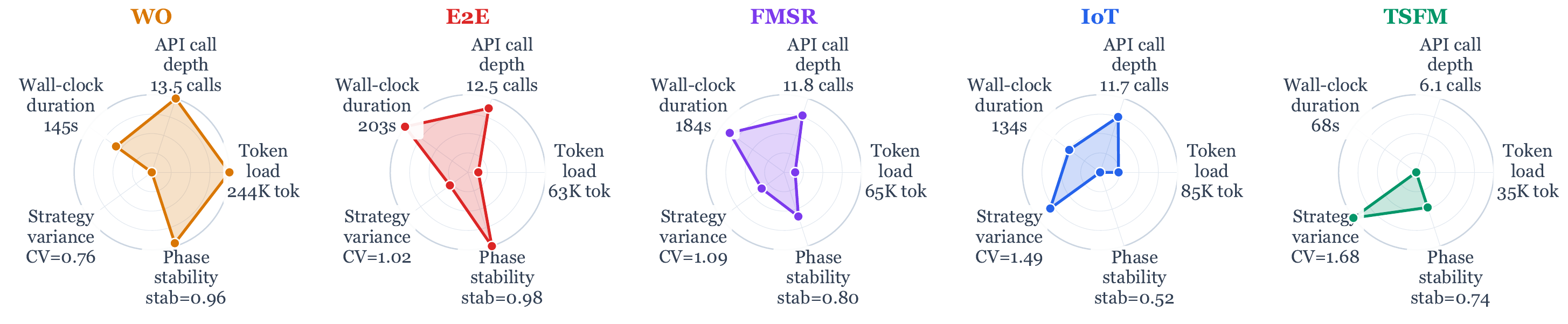}
  \vspace{-0.15in}
  \caption{%
    \textbf{Benchmark fingerprint.}
    Normalised computational cost profile per agent domain
    across five axes. Raw values and provenance is provided in Table~\ref{tab:fingerprint_provenance}.%
  }
  \label{fig:benchmark_fingerprint}
  \vspace{-0.2in}
\end{figure*}

Each submission executes against four domain-specific agents
(IoT, FMSR, TSFM, WO) plus an end-to-end multi-agent class
(E2E). As every execution produces a trajectory log, we construct a five-axis fingerprint per domain from four automatically logged quantities, namely \textit{tokens sent, API calls, wall-clock duration, and phase label.} Figure~\ref{fig:benchmark_fingerprint}) and Appendix Table~\ref{tab:fingerprint_provenance} defines each axis. 

Three findings are readable from the shape contrast.
\textbf{(i)}~WO is expensive but fair: the lowest strategy variance
($\mathrm{CV}=0.76$) and the highest phase stability ($0.96$) among four domains confirm that its cost is task-intrinsic and not gameable. \textbf{(ii)}~
TSFM is cheap but gameable: lowest token load yet highest
variance ($\mathrm{CV}=1.68$), so a leaderboard weighted toward
TSFM \textbf{measures prompt sensitivity} more than agent
capability. \textbf{(iii)}~E2E isolates orchestration as a fixed
latency cost independent of the prompt strategy (stability $=0.98$),
invisible to token-count metrics alone. The key inversion is that the cheapest domain (TSFM, 35K tokens) is the most prompt-sensitive, while the most expensive (WO, 244K tokens) is the most robust. This inversion has a direct implication in the design of the leaderboard. A
scoring formula that weights domain agents by token cost, a
common efficiency-aware choice, would assign disproportionate
weight to TSFM and, therefore, measure prompt sensitivity rather
than agent capability. Conversely, the uniform weighting used in
this competition underweights the domain (WO) whose scores most
reliably reflect capability rather than prompt-tuning artifacts.

\subsection{Scenario Complexity Analysis.} The development and evaluation phases use disjoint sets of 11 scenarios, and decomposing leaderboard scores along the six qualitative metrics at the per-scenario level surfaces two patterns. First, hallucination is a coupled rather than independent failure mode. In all 22 scenarios, the Pearson correlation between hallucination rate and overall task quality is strongly negative ($r \approx -0.93$), indicating that it serves as a leading indicator of greater failure rather than a standalone metric. Second, the work-order category forms the consistent capability ceiling. The three hardest scenarios in both phases are work-order tasks (Q424, Q405, Q400 in development, and Q411, Q410, Q403 in evaluation), each with hallucination rates above $0.73$. This convergence holds despite WO exhibiting the highest cross-phase cost stability ($0.96$) and lowest strategy variance ($\mathrm{CV}=0.76$) in our cost fingerprint, arguing that long-horizon reasoning over historical maintenance records is a genuine capability gap rather than a benchmark artifact.

\vspace{-0.05in}
\subsection{Top Submissions and Strategy Patterns}
\label{sec:solutions}

Each successful execution produces six score per-scenario, namely \textit{task completion, retrieval accuracy, result verification, action sequencing, clarity, and hallucination avoidance}~\citep{patel2025assetopsbench}. This gives a 
fine-grained feedback from the participant to guide their next submission.
Track~1 isolates the planning fidelity, exposing action
sequencing and task completion as the main skill axes.
Track~2 keeps the planner fixed and varies the workflow
architecture and context handling, exposing instead result
verification and avoidance of hallucinations. Since this is a
\textbf{code submission competition}, we study the implementation
patterns that appear in accessible source-level artifacts. We distinguish three levels of abstraction. A \textit{strategy} refers to a high-level approach adopted by the participants (e.g., fallback handling or prompt refinement). A \textit{method} denotes the concrete implementation of a strategy in code. An \textit{archetype} refers to a cluster of similar strategies identified through a clustering based on embedding of submissions. We cluster all 331 accessible source artifacts (210 Track~1, 121 Track~2) in a
sentence-transformer embedding space
(\S\ref{sec:archetype-taxonomy};
Appendix~\ref{app:clustering}) and use the resulting taxonomy below. 

\begin{figure}[h]
\centering
\includegraphics[width=0.85\columnwidth]{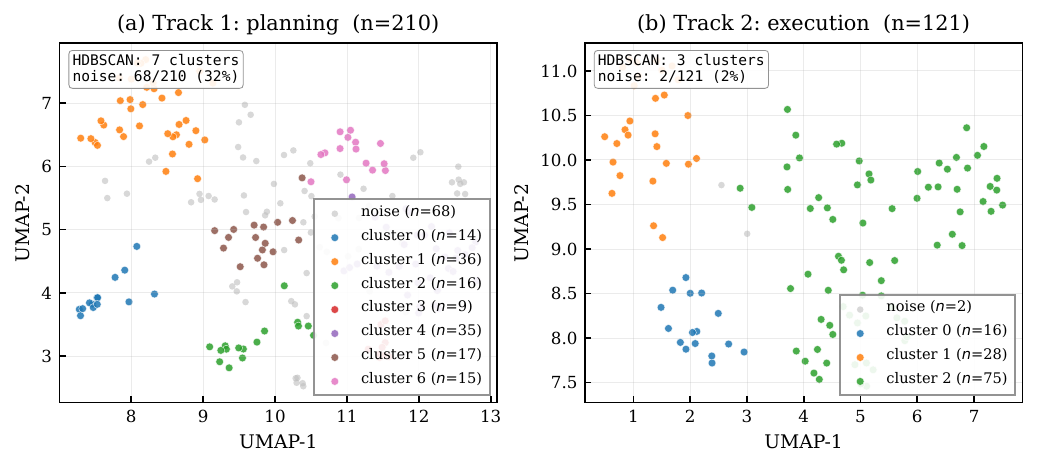}
\vspace{-0.1in}
\caption{Track~1 (left) scatters with 32\% noise; Track~2 (right)
forms three clusters with $<$2\% noise.}
\label{fig:umap}
\vspace{-0.1in}
\end{figure}

\paragraph{Archetype taxonomy.}
\label{sec:archetype-taxonomy}

Clustering reveals a sharp asymmetry between tracks
(Figure~\ref{fig:umap}). Planning submissions are spread
out diffusely. We use HDBSCAN, a density-based clustering algorithm that discovers clusters and labels outliers as noise without requiring a predefined number of clusters. HDBSCAN returns 7 clusters with 32\% noise and
a silhouette below 0.12 across all $K\in[2,10]$. Execution
submissions are concentrated into 3 tight clusters with only 1.6\% noise. Planning is \emph{commoditized}; execution is \emph{structured}. The complete methodology is in Appendix~\ref{app:clustering}.

At $K{=}5$ (see Figure~\ref{fig:archetype_bars}), the 
planning surface is dominated by P1. P1 covers $32.4\%$ 
of the Track~1 submissions. It consists of knowledge-base-grounded 
prompts that invoke agents directly from a structured 
catalog. The execution surface is dominated by E1. E1 
accounts for $28.9\%$ of the Track ~2 submissions. It is a 
soft-validation fallback pattern that catches 
\texttt{ValueError} exceptions and proceeds with partial 
results rather than failing. The three dominant planning 
archetypes (P1--P3) cover $80\%$ of Track~1. The four 
output-control execution archetypes (E1--E4) cover $88.4\%$ 
of Track~2.

\begin{figure*}[t]
  \centering
  \includegraphics[width=\textwidth]{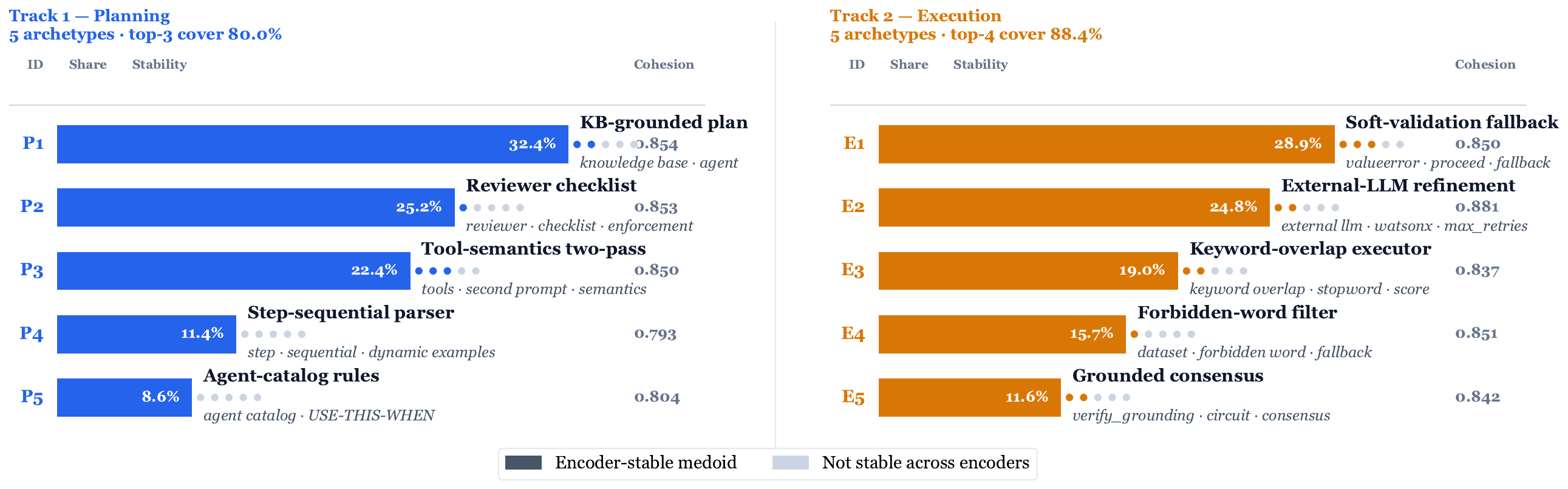}
  \caption{%
    \textbf{Archetype taxonomy at $K{=}5$.}
    Bar length: cluster share (\%). Filled dots: medoids
    stable across both encoders. Italic terms:
    class-TF--IDF top tokens.%
  }
  \label{fig:archetype_bars}
  \vspace{-0.1in}
\end{figure*}
% =====================================================================
\paragraph{Track~1 Top Submissions.} Track~1 is the cleanest scientific subset. The main pattern is clear from released code (Table~\ref{tab:track1_static}):
the prompt length expands $12\times$ and code size more than
doubles without improving the public score, confirming
that the public surface is saturated. Hidden scores refine
rather than overturn this. Among the three submissions tied
publicly at $72.73\%$, hidden scores span an 18-point range:
\code{jainrishi601} (P1, knowledge-base-grounded) leads at
$63.64\%$, \code{kanishk\_007} (P3, tool-semantics) sits at
$54.55\%$, and \code{vamsikv28} (P5, agent-catalog) drops to
$45.45\%$, consistent with high-cohesion, encoder-stable
archetypes generalizing better than low-stability ones.

\begin{table}[h!]
\centering
\vspace{-0.1in}
\caption{Verified Track~1 static metrics. \textit{LoC}: number of lines in code submission. \textit{Arch.}: Archetype inferred from textual description~(\S\ref{sec:archetype-taxonomy}). 
\textit{T.Sim}:Template similarity against the starter kit}
\scriptsize
\setlength{\tabcolsep}{5pt}
\renewcommand{\arraystretch}{1.35}
\begin{tabularx}{\columnwidth}{@{} l r r r r r X r @{}}
\toprule
\textsc{Submission} &
\textsc{Pub.(\%)} &
\textsc{Hid.(\%)} &
\textsc{LoC} &
\textsc{Prompt} &
\textsc{T.sim} &
\textsc{Main motif} &
\textsc{Arch.} \\
\midrule
\rowcolor{rowblue!12}
\code{rohith} (proxy) & 72.73 & 54.55 & 200 & 745  & 1.000 & Template-equivalent baseline          & base \\
\rowcolor{rowblue!12}
\code{radhesham}      & 72.73 & ---   & 194 & 757  & 0.982 & Near-template, cosmetic edits         & base \\
\rowcolor{rowgreen!12}
\code{jainrishi601}   & 72.73 & \textbf{63.64} & 428 & 8{,}849 & 0.330 & Typed requests, worked examples & P1 \\
\rowcolor{rowamber!12}
\code{kanishk\_007}   & 72.73 & 54.55 & 372 & 3{,}440 & 0.254 & Worked examples, fuzzy repair       & P3 \\
\rowcolor{rowred!12}
\code{vamsikv28}      & 71.43 & \textbf{45.45} & 331 & 7{,}033 & 0.396 & Anti-hallucination, tool guidance   & P5 \\
\bottomrule
\end{tabularx}
\label{tab:track1_static}
\vspace{-0.1in}
\end{table}

\paragraph{Track~2 accessible top artifacts as guardrail engineering.}

Track~2 is methodologically looser than Track~1 but 
scientifically more revealing. The four dominant archetypes E1--E4 are structural variants of a single strategy that catches bad intermediate outputs and redirects them. Among accessible Track~2 artifacts, the observed improvements are therefore better explained by guardrails around a fixed ecosystem than by new agent architectures. The encoder-stability markers in Figure~\ref{fig:archetype_bars} support this reading, with every execution archetype retaining at least one stable medoid across encoders, while two of the five planning archetypes do not. The execution structure is real; two planning 
archetypes are encoder artifacts. 

\begin{figure}[t]
  \centering
  \includegraphics[width=0.9\linewidth]{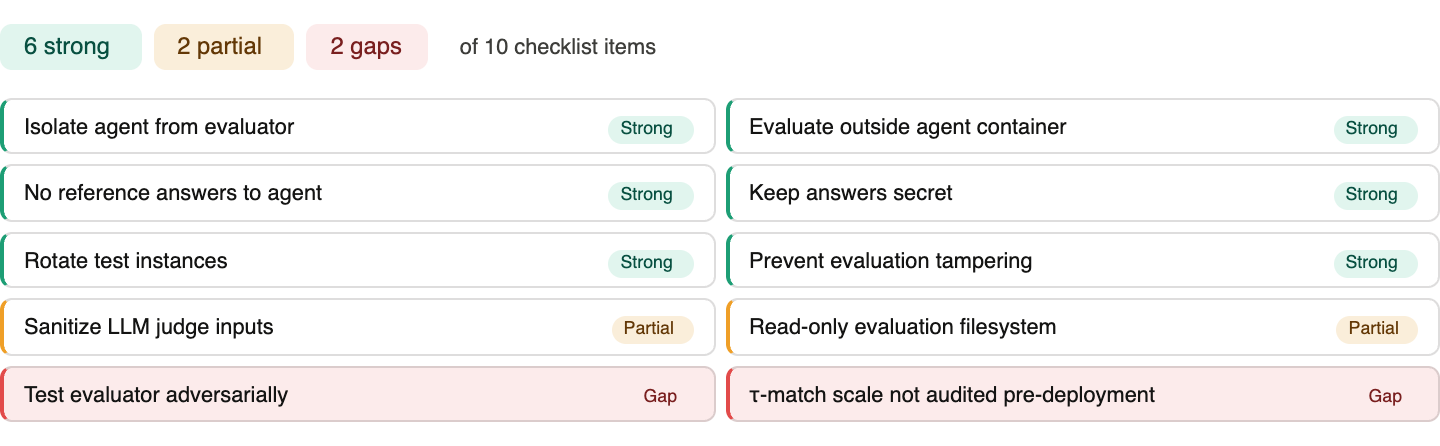}
  \vspace{-0.1in}
  \caption{Alignment with the Agent-Eval
Checklist. 
Green =
satisfied, amber = partial, red = gap.}
   \vspace{-0.15in}
  \label{fig:checklist}
\end{figure}

\begin{comment}
The winner report is consistent~\citep{anonymous_supp}.
\code{Smart Maintenance Crew} wins through balanced
cross-track performance, \code{LostSouls} leads on private
execution, and \code{WaterLevel} on planning quality.
Combined with the public--private reversals in
Table~\ref{tab:hidden_stats} and the 88\% concentration into
output-control archetypes, execution robustness is a distinct
capability slice that the public leaderboard only weakly
reflects. A reproducibility caveat applies: several
submission trees have unimplemented helper logic, and the
public archive is not a perfect record of the executed
system~\citep{anonymous_supp}.
\end{comment}

\vspace{-0.1in}
\section{Discussion}

\paragraph{Alignment with the Agent-Eval Checklist.}
\label{sec:checklist}

\begin{wrapfigure}{r}{0.44\textwidth}
  \centering
  \vspace{-10pt}
  \includegraphics[width=\linewidth]{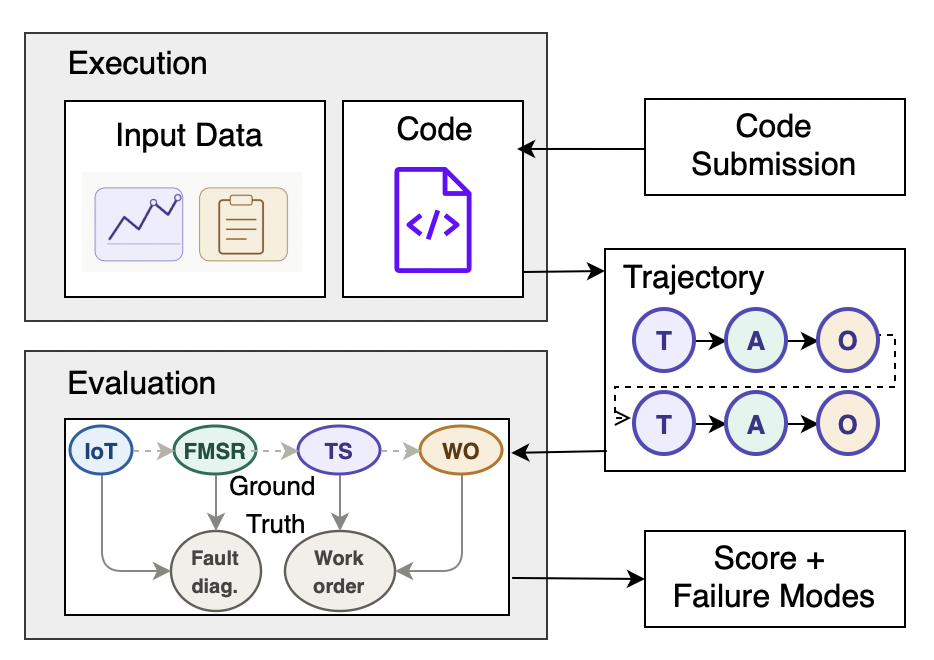}
  \caption{%
    Execution environment. Agent and evaluator run in isolated
    containers.%
  }
  \label{fig:exe_env}
  \vspace{-6pt}
\end{wrapfigure}

A recent practitioner report proposes a minimum-bar checklist for
trustworthy agent benchmarks~\citep{wang2026trustworthybenchmarks},
published after our competition ended. Figure~\ref{fig:checklist}
maps our infrastructure to all ten requirements. Six are fully
satisfied by design, covering agent-evaluator isolation (See Figure \ref{fig:exe_env}), answer secrecy, test-instance rotation, and tamper prevention. Two are
partially satisfied: LLM judge input sanitization used structured
dimensions but was not adversarially tested, and read-only filesystem
enforcement was not explicitly documented. The two genuine gaps both
concern the robustness of the evaluation. The \tmatch{} scale sensitivity identified in Section~\ref{sec:scores} is precisely the scoring robustness issue flagged by the checklist, and a pre-deployment scorer audit would have caught it before the results were published. Interestingly, Trajectory-level analysis confirms the evaluation agent is 
not biased toward verbosity or over-engineered behavior such as 
success correlates negatively with token usage, execution 
length, and tool entropy 
(Appendix~\ref{app:eval-bias}). We offer this as an empirical validation of the
checklist's value and encourage future organizers to treat it as a
pre-publication requirement rather than a post-hoc audit.

%\paragraph{Limitations and Broader Impacts.}
%Our analysis is bounded by the released artifacts.  Hidden-phase conclusions rest on best-submission   spreadsheets that required manual normalization. Beyond the leaderboard, the competition trains \textbf{a community of practitioners and students} to build agentic systems for industrial domains, with infrastructure that lowers the barrier to entry for this research. Finally, coarse leaderboards can produce misplaced confidence if interpreted without semantic validation. Thus, evaluation transparency is priority over leaderboard-first reporting. An open question is whether micro-benchmarking~\citep{yauney2026microbenchmark} can be adapted to agentic competitions; our results suggest that reliable subset-based evaluation for long-horizon systems remains challenging.

\paragraph{Limitations and Broader Impacts.}
Our analysis is bounded by the released artifacts. 
Hidden-phase conclusions rest on best-submission 
spreadsheets that required manual normalization. Beyond the  leaderboard, the competition trains \textbf{a community of practitioners and students} to build agentic systems for industrial domains. Finally, coarse leaderboards can produce misplaced confidence if interpreted without semantic validation, so evaluation transparency should take precedence over leaderboard-first reporting. An open question is whether micro-benchmarking~\citep{yauney2026microbenchmark} can be adapted to agentic competitions. Our results suggest that reliable subset-based evaluation remains challenging.

\section{Conclusion}
The CODS~2025 \assetops{} challenge demonstrates why 
agent competitions should be analyzed as evaluation 
instruments, not only ranking mechanisms. Its hidden 
execution phase, track isolation, scoring traces, and 
submission artifacts reveal leaderboard saturation, 
public--hidden mismatch, metric-scale sensitivity, and 
guardrail-centric top-submission strategies. These 
observations suggest four prescriptions for future 
competitions: isolate capability dimensions across tracks, 
surface skill-aware subscores, ensure scoring components 
are numerically commensurate, and treat team metadata, 
container digests, and versioned code as core deliverables. 
Adopting these practices lets competitions function as 
rigorous scientific instruments that publish the evidence 
needed to understand what was actually measured.

\begin{comment}
\section{Related Work}
What is special about out agentic challenge. 
1. Domain 
2. Multi-Agent
3. Two tracks
4. Multiple Eval metric
5. Quetion wise guidance
6. Failure Mode analysis
7. long hotizon running tasks
8. multi-step reasoning including tool calling
9. Code based/final outcome
\end{comment}

\bibliographystyle{plainnat}
\bibliography{main}

\clearpage
\appendix

\appendix

\section{Technical appendices and supplementary material}

This appendix presents a structured post-competition 
analysis of \assetops{} organized along nine complementary 
analytical dimensions. The base layer is the competition 
itself: its design, participants, and evaluation protocol. 
Nine analytical dimensions are built on top of this 
foundation, each asking a distinct scientific question that 
aggregate leaderboard results alone cannot answer.

Appendix \ref{sec:materials} includes the competition related materials. 

\textit{Dimension 1} (Relevant Competitions, Section ~\ref{sec:related}) situates \assetops{} within the 
recent landscape of competition-track benchmarks and 
identifies the methodological gap it addresses. 

\textit{Dimension 2} (Participation and Setup, Section ~\ref{sec:app:setup}) characterizes who competed, 
under what evaluation conditions, and what diagnostic 
information was provided.

\textit{Dimension 3} (Submission Behaviour, Section ~\ref{sec:app:behaviour}) examines how teams engaged 
with the benchmark over time.

\textit{Dimension 4} (Ranking Robustness, Section ~\ref{sec:app:robustness}) evaluates whether the 
official leaderboard faithfully reflects true agent quality.

\textit{Dimension 5} (Computational Footprint, Section ~\ref{sec:app:cost}) analyzes the compute cost of 
running the benchmark and its relationship to task 
difficulty.

\textit{Dimension 6} (Strategy Attribution, Section ~\ref{sec:app:strategy}) examines what participants 
actually implemented and which design choices were 
associated with score differences.

\textit{Dimension 7} (Evaluation Agent Robustness, Section ~\ref{app:eval-bias}) tests whether the LLM-as-judge evaluation agent exhibits bias toward superficial 
trajectory characteristics such as verbosity or execution  complexity. 

\textit{Dimension 8} (Clustering Methodology, Section ~\ref{app:clustering}) details the methodology for clustering submission strategies, including embedding, dimensionality reduction, clustering algorithms, and validation metrics used to derive archetype-level groupings. 

\textit{Dimension 9} (Failure Mode Distribution and Taxonomy Analysis of Submissions, Section ~\ref{sec:appendix_failure_modes}) presents a comprehensive analysis of failure-mode distributions in submissions, including frequency patterns, clustering, and a hierarchical taxonomy that consolidates semantically related failure behaviors.

Our goal is not to reframe \assetops{} as a methods 
contribution, but to document the competition with 
sufficient rigour that its design choices, outcomes, and 
failure modes can be verified, compared against, and built 
upon by future benchmark organizers. We propose this 
nine-dimensional retrospective as a reusable analytical 
template for future agentic benchmark competitions, where 
structured post-competition analysis of this depth is 
currently absent from the literature.

\section{Competition Related Materials}
\label{sec:materials}
%Figures~\ref{fig:coda}, \ref{fig:web1}, \ref{fig:web2} and \ref{fig:web3} illustrate the Codabench interface, the official challenge website, Agentic AI advertisement webpage at CODS 2025, and the registration portal for the AssetOpsBench Agentic AI Challenge.

\subsection{Official Challenge Website}
\label{sec:website}

\begin{figure}[!ht]
    \centering
    \includegraphics[width=0.9\linewidth]{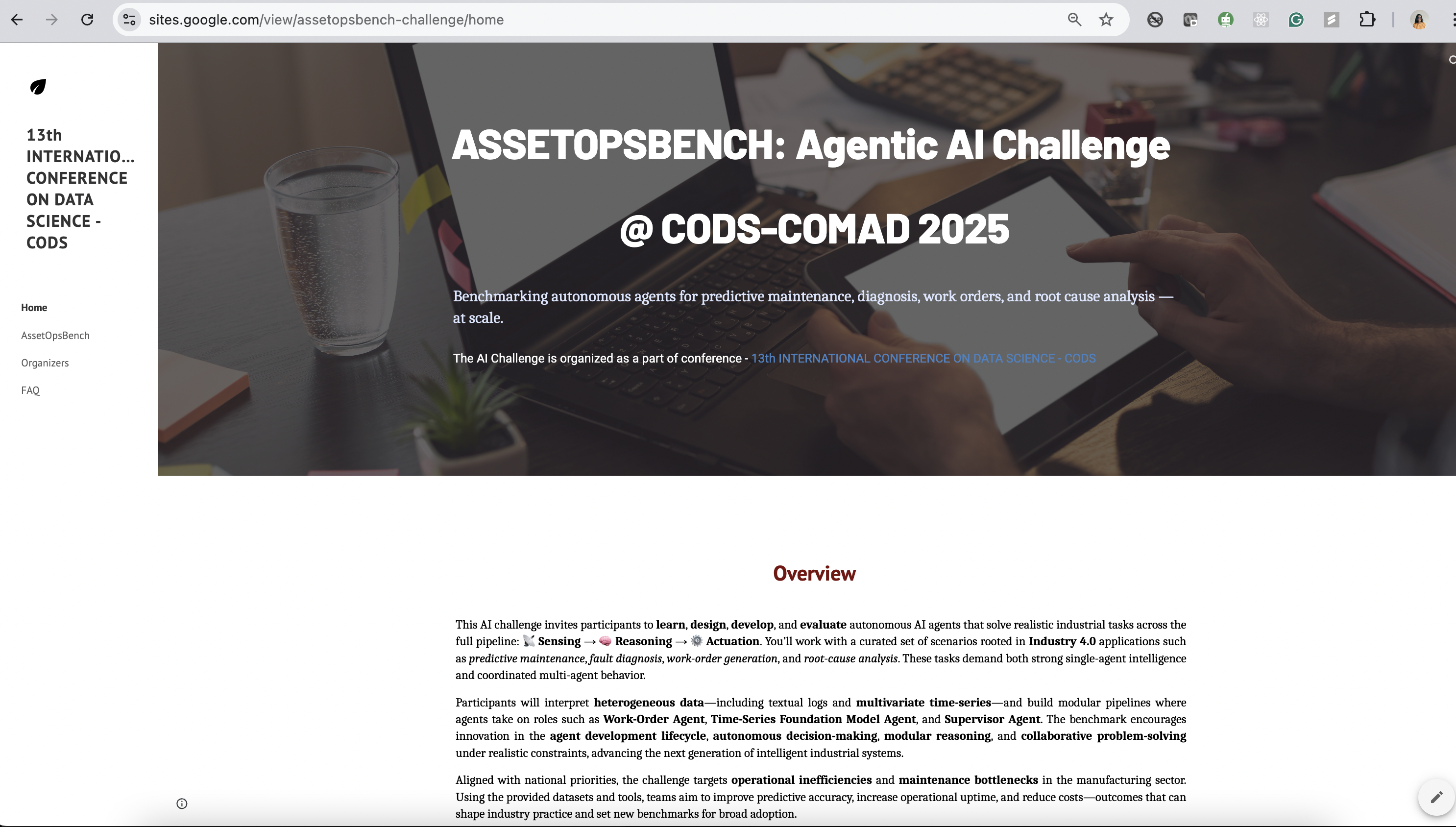}
    \caption{Official Agentic-AI competition website. Link: \url{https://sites.google.com/view/assetopsbench-challenge/home?authuser=0}}
    \label{fig:web1}
\end{figure}

The official website (\ref{fig:web1}) provides an overview of the challenge, including task descriptions, objectives, and participation guidelines.

\subsection{Challenge Advertisement Page}
\label{sec:advertisement}

\begin{figure}[!ht]
    \centering
    \includegraphics[width=0.9\linewidth]{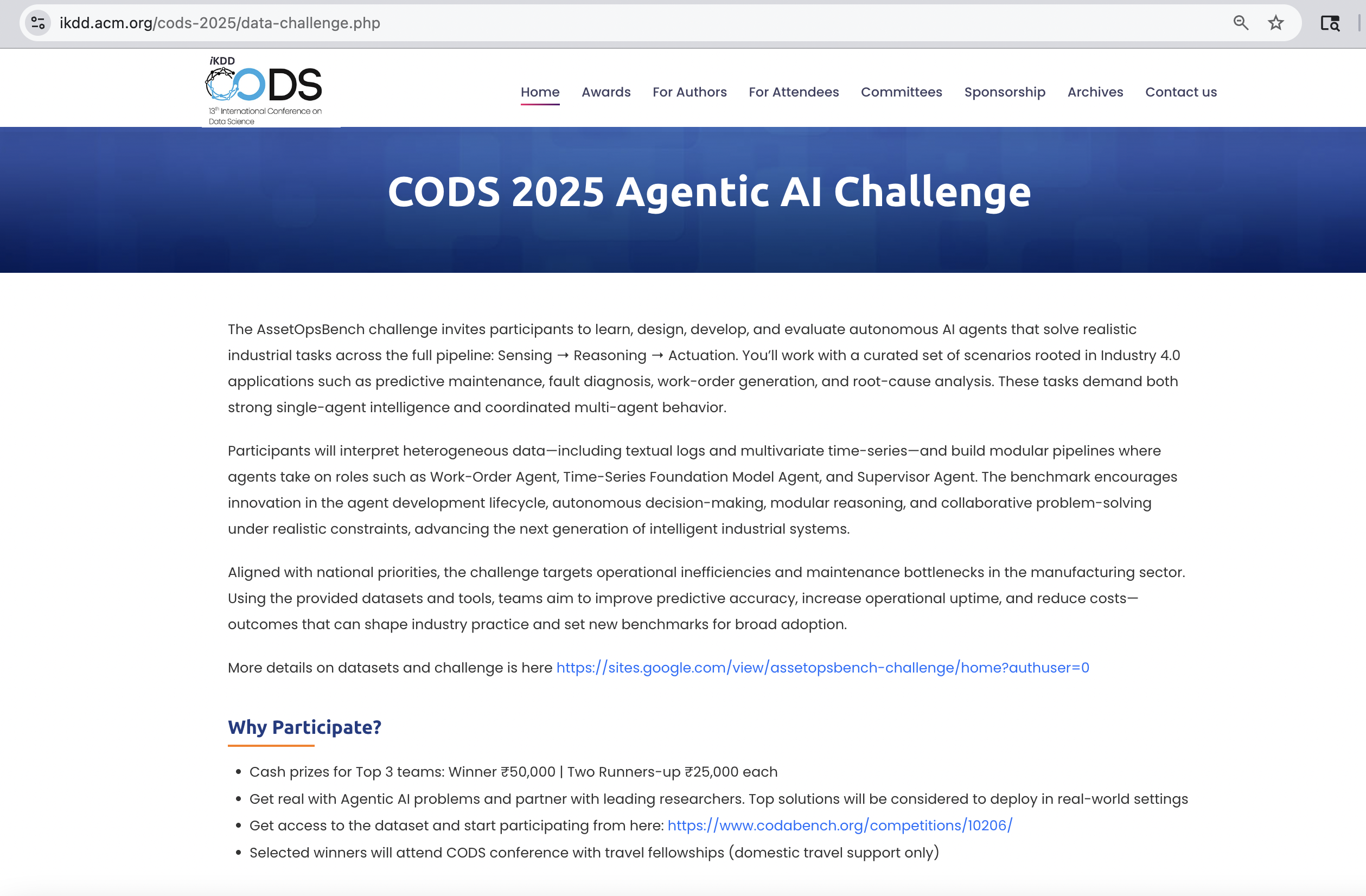}
    \caption{Agentic AI Challenge Advertisement Webpage at CODS 2025. Link: \url{https://ikdd.acm.org/cods-2025/data-challenge.php}}
    \label{fig:web2}
\end{figure}

The CODS 2025 challenge page (Figure \ref{fig:web2}) serves as an announcement and dissemination platform, outlining the motivation and scope of the benchmark.

\subsection{Registration Interface}
\label{sec:registration}

\begin{figure}[!ht]
    \centering
    \includegraphics[width=0.9\linewidth]{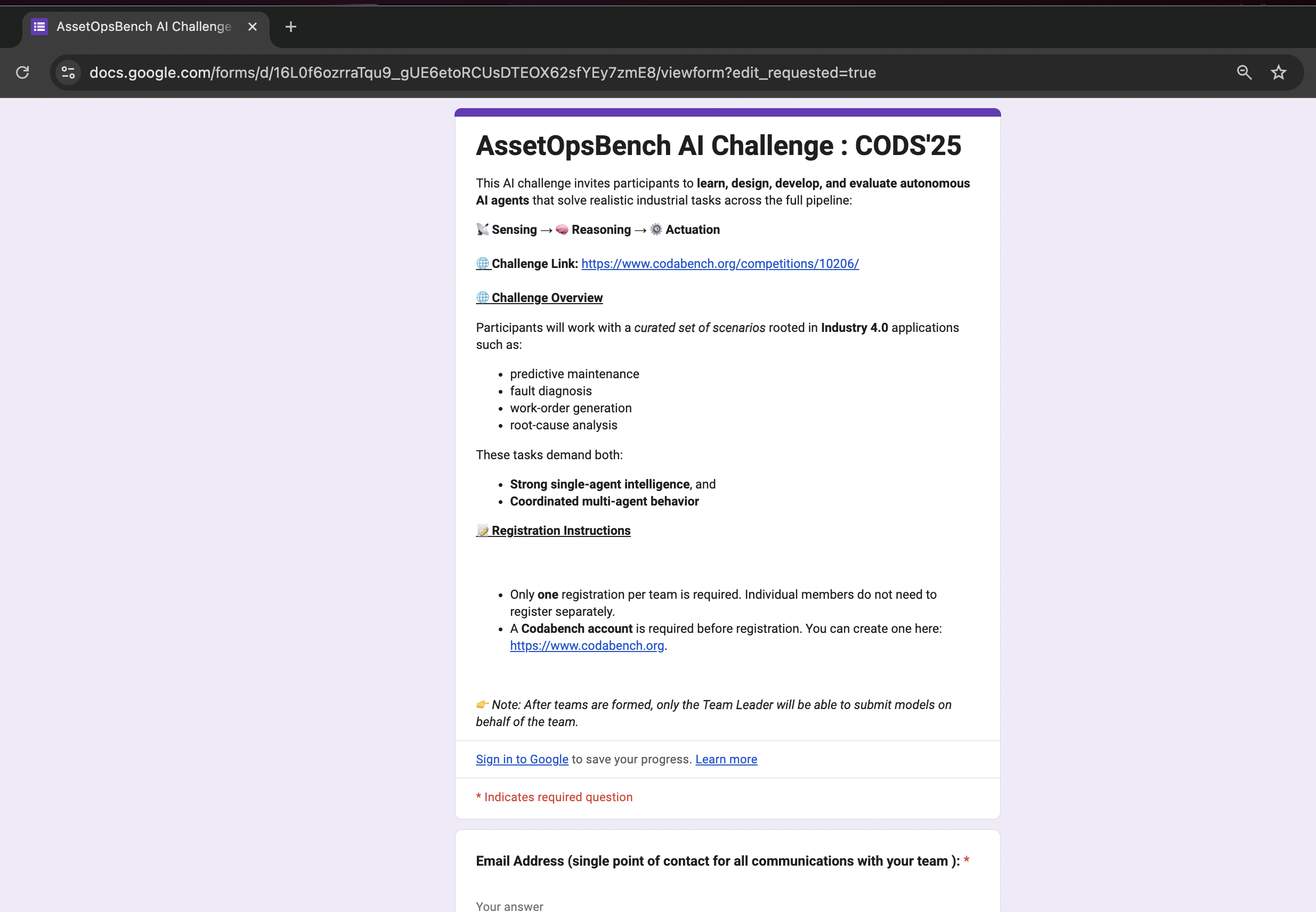}
    \caption{Registration for for Agentic AI Challenge. Link: \url{https://docs.google.com/forms/d/16L0f6ozrraTqu9_gUE6etoRCUsDTEOX62sfYEy7zmE8/viewform?edit_requested=true}}
    \label{fig:web3}
\end{figure}

Participants register for the challenge through the online form (Figure \ref{fig:web3}) that collects team and contact information.

\clearpage

\section{Dimension 1: Relevant Competitions}
\label{sec:related}

\textsc{AssetOpsBench} sits in a lineage of competition-track
benchmarks (Table~\ref{tab:competition_comparison}) that has shifted
noticeably over the last two years along two axes: (i)~from
\emph{static prediction tasks} to \emph{agentic, multi-step
evaluation}, and (ii)~from \emph{consumer or scientific} domains to
\emph{regulated or privacy-constrained} ones. We situate our
contribution along both axes.

\begin{table*}[h!]
\centering
\caption{%
  Positioning of \textsc{AssetOpsBench} among recent
  competition-track benchmarks at top venues. Participant and
  submission columns report public platform-level metadata when
  available, not the internal registration and server-log denominators
  used elsewhere in this retrospective. $^*$Submission including post competition deadline.
  }
\small
\setlength{\tabcolsep}{5pt}
\renewcommand{\arraystretch}{1.35}
\begin{tabular}{c L{2.6cm} L{1.6cm} r r C{1.1cm} L{1.3cm} L{2.2cm}}
\toprule
\textbf{Year}
  & \textbf{Competition}
  & \textbf{Venue}
  & \textbf{Part.}
  & \textbf{Subs.}
  & \textbf{Agentic}
  & \textbf{Platform}
  & \textbf{Domain} \\
\midrule
% ── 2024 ─────────────────────────────────────────────────────────────────────
\multirow{3}{*}{\rotatebox[origin=c]{90}{\textbf{2024}}}
  & Capture-the-Flag  & SaTML   & 163          & 137K
  & ---        & Custom      & LLM Security \\
  & ML4CFD            & NeurIPS & ${\sim}$240         & 650
  & ---        & CodaBench & Fluid Dynamics \\
  & HiggsML           & NeurIPS & 1{,}785      & ---
  & ---        & Codabench & Energy Physics \\
\midrule
% ── 2025 ─────────────────────────────────────────────────────────────────────
\multirow{6}{*}{\rotatebox[origin=c]{90}{\textbf{2025}}}
  & CRAG-MM           & KDD     & ${\sim}1000$ & ${\sim}5000$
  & \checkmark & AIcrowd & Multi-modal \\
  & CureBench         & NeurIPS & 390   & 2,773
  & \checkmark & Kaggle & Healthcare AI \\
  & MathArena         & NeurIPS & ---   & 162
  & ---        & Custom        & Math Reasoning \\
  & Erasing Invisible & NeurIPS & 298   & 2{,}722
  & ---        & CodaBench & Content Security \\
  & Deepfake Detect & IJCAI & --- & --- & --- & Tianchi & Multi-Model \\
  & \cellcolor{rowhl}\textbf{\textsc{AssetOpsBench}}
  & \cellcolor{rowhl}\textbf{CODS}
  & \cellcolor{rowhl}\textbf{349}
  & \cellcolor{rowhl}\textbf{507$^*$}
  & \cellcolor{rowhl}\checkmark
  & \cellcolor{rowhl}CodaBench
  & \cellcolor{rowhl}\textbf{Physical Asset} \\
\bottomrule
\end{tabular}
\label{tab:competition_comparison}
\end{table*}

\paragraph{Scientific and security benchmarks as the prior norm.}
The 2024 competition cycle was dominated by static, supervised-style
evaluations. \textsc{HiggsML}~\citep{bhimji2025fair} revived the
Higgs-boson classification benchmark with $1{,}785$ registered
participants on Codabench, and \textsc{ML4CFD}~\citep{yagoubiml4cfd}
evaluated surrogate models for fluid dynamics, also on Codabench.
The SaTML Capture-the-Flag event~\citep{debenedetti2024dataset} adapted a
classic security-competition format to LLM prompt-injection attacks
on a custom platform. In each case the system under test produces a
fixed output (label, regression vector, exploit string) given a
fixed input, so success is adjudicated by a single numerical metric
and the platform's role is largely submission ingestion. The 2025
benchmarks \textsc{MathArena}~\citep{balunovicmatharena},
\textsc{Erasing Invisible}~\citep{dingtechnical}, and the
IJCAI deepfake-detection challenge~\citep{deepfake_ijcai2025, tianchi_deepfake2025}
continued this pattern across math reasoning, watermark removal,
and multi-modal forgery respectively, with submission counts in the
hundreds-to-low-thousands range.

\paragraph{The agentic shift.}
The most recent cycle (2025) began to evaluate \emph{systems} rather
than \emph{predictions}. \textsc{CRAG-MM}~\citep{cragmm2025, NEURIPS2024_1435d2d0} at KDD
scored roughly $5{,}000$ submissions from $\sim\!1{,}000$ participants
on multi-modal retrieval-augmented generation, with agents invoking
retrievers and tools over structured multi-turn inputs.
\textsc{CureBench}~\citep{curebench2025, cofala2025medai} at NeurIPS assessed $390$
participants over $2{,}773$ submissions on agentic healthcare
reasoning with patient-simulator interaction. Both competitions
surface the central evaluation difficulty of agentic tasks:
trajectory-level scoring, multi-step tool use, hidden test-set
leakage risk, but they operate in consumer-adjacent domains
(open multi-modal web, open biomedical reasoning) where the
evaluation data can plausibly be released, replicated, or
crowd-sourced. Accordingly, they run on \textsc{AIcrowd} and
\textsc{Kaggle} respectively, both of which assume submission-time
data access.

\paragraph{Gap: agentic evaluation in industrial, privacy-constrained
domains.}
\textsc{AssetOpsBench} is, to our knowledge within the competitions
surveyed in Table~\ref{tab:competition_comparison} and under the
released public descriptions, the only competition-track benchmark
that jointly combines three properties: an \emph{agentic} task
(multi-agent plan-and-execute workflows, not single-shot prediction),
a \emph{physical-asset industrial} domain (condition monitoring,
work-order drafting, and inspection agents grounded in real asset
telemetry), and a \emph{privacy-constrained deployment} (proprietary
agent telemetry, asset-identifier confidentiality, and sponsor
compliance requirements that ruled out Kaggle and Hugging Face).
Satisfying all three forced the infrastructure choices we
document in section \ref{sec:design}: the Codabench container architecture,
the freeze/edit block design that isolates participant code from
evaluation data, and the hidden-phase protocol. Of the ten
benchmarks in Table~\ref{tab:competition_comparison}, only three
(\textsc{CRAG-MM}, \textsc{CureBench}, \textsc{AssetOpsBench}) are
agentic; only \textsc{AssetOpsBench} targets industrial physical
assets; and only the Codabench-hosted entries support the
container-based isolation that privacy-constrained sponsors require.

%\paragraph{Methodological companions.}
%Our analysis also draws on the emerging literature on trustworthy agent evaluation. The Agent-Eval Checklist~\citep{wang2026trustworthybenchmarks} formalises requirements (hidden-phase evaluation, scoring auditability, trajectory-level failure analysis) that our infrastructure independently satisfied; we use it in \S\ref{sec:checklist} as an external validation target. Trajectory-level failure taxonomies from \citet{cemri2025multi} and \citet{zhang2025agent} inform our reading of execution-track outcomes. Finally, our embedding-based archetype analysis (\S\ref{sec:archetype-taxonomy}) extends BERTopic-style class-TF--IDF cluster interpretation~\citep{grootendorst2022bertopic} from document topic modelling to the retrospective characterization of participant \emph{strategies} in an agentic competition --- a use, to our knowledge, novel to this work.

\paragraph{Methodological companions.}
Our analysis also draws on emerging work that treats agent evaluation as a first-class research problem. Recent work on general-agent evaluation~\citep{bandelgeneral} argues that existing benchmarks are fundamentally limited by domain-specific assumptions, bespoke interfaces, and implicit knowledge of task structure, which obscure true generalization. Complementing this perspective, the Agent-Eval Checklist~\citep{wang2026trustworthybenchmarks} formalises requirements (hidden-phase evaluation, scoring auditability, trajectory-level failure analysis) that our infrastructure independently satisfied; we use it in section \ref{sec:checklist} as an external validation target. Trajectory-level failure taxonomies from \citet{cemri2025multi} and \citet{zhang2025agent} inform our reading of execution-track outcomes. Finally, our embedding-based archetype analysis (section \ref{sec:archetype-taxonomy}) extends BERTopic-style class-TF--IDF cluster interpretation~\citep{grootendorst2022bertopic} from document topic modelling to the retrospective characterization of participant \emph{strategies} in an agentic competition --- a use, to our knowledge, novel to this work.
\paragraph{Scale.}
Using the platform-level metadata reported for cross-competition
comparison ($349$ participants and $507$ submissions in
Table~\ref{tab:competition_comparison}), \textsc{AssetOpsBench} sits
in the middle of the distribution: larger than the
$162$-submission \textsc{MathArena} reasoning benchmark, comparable
to \textsc{Erasing Invisible} ($298/2{,}722$) and
\textsc{CureBench} ($390/2{,}773$) in submission throughput per
participant, and smaller than the most mass-participation
benchmarks (\textsc{HiggsML}, \textsc{CRAG-MM}). These public
platform counts are not the same denominator as the retrospective
analysis data: our internal analyzes use 149 registration-form teams,
349 declared member slots, a 300-attempt server log, 234
\texttt{Finished} attempts, and 331 accessible source-level artifacts
for strategy clustering. The scale is sufficient to support the
cluster-level analyzes in section \ref{sec:archetype-taxonomy} while
remaining tractable for the container-based privacy model.

% =============================================================================
\section{Dimension 2: Participation and Setup}
\label{sec:app:setup}
% =============================================================================

This dimension characterises the competition from the perspective of its
design: who participated, what tasks they were evaluated on, and what
information was made available to them at each stage.
A recurring principle throughout the design is that evaluation integrity was
preserved by withholding scenario utterances from participants entirely,
while providing rich per-category diagnostic feedback to support iterative
improvement without leaking information about the evaluation questions.

% ─────────────────────────────────────────────────────────────────────────────
\subsection{Participating Teams}
\label{sec:app:teams}
% ─────────────────────────────────────────────────────────────────────────────

Out of 149 registered teams that participated in \assetops{},
this paper focuses on the top 11 teams by final combined score.
To ensure readability, each team is referred to by an encoded identifier (Team~A through
Team~K) rather than its registered competition name.
Identifiers are assigned alphabetically in order of final ranking position.
The mapping between identifiers and registered names is provided in
Table~\ref{tab:team_mapping}; gold, silver, and bronze colouring identifies
the top-three finishers.

\begin{table}[h]
\centering
\caption{Compact label used throughout the paper for tables and figures. Identifiers are assigned alphabetically by final rank, providing a uniform short label suitable for figure axes and dense tables; team names are not anonymised and appear in Table~\ref{tab:fullranking} (final ranking) and the discussion. 
Row colouring: \colorbox{gold!60}{\strut gold} = rank~1,
\colorbox{silver!50}{\strut silver} = rank~2,
\colorbox{bronze!40}{\strut bronze} = rank~3.}
\small
\renewcommand{\arraystretch}{1.20}
\begin{tabular}{ll}
\toprule
\textbf{Label} & \textbf{Registered Team Name} \\
\midrule
\rowcolor{gold!60}   Team A & WaterLevel \\
\rowcolor{silver!50} Team B & BlueCube \\
\rowcolor{bronze!40} Team C & Smart Maintenance Crew \\
                     Team D & Entropians \\
                     Team E & aviation\_agent \\
                     Team F & LostSouls \\
                     Team G & Scalar\_nitk \\
                     Team H & Infinity \\
                     Team I & kinatic \\
                     Team J & horizon \\
                     Team K & EXL Health AI Lab \\
\bottomrule
\end{tabular}
\label{tab:team_mapping}
\end{table}

% ─────────────────────────────────────────────────────────────────────────────
\subsection{Evaluation Scenarios}
\label{sec:app:scenarios}
% ─────────────────────────────────────────────────────────────────────────────

We disclose the complete set of 11 scenario pairs used across both phases of
\textsc{AssetOpsBench}.
Each scenario consists of a \emph{development utterance} used during the open
competition window (Phase~1) and a semantically related \emph{evaluation
utterance} used in the hidden assessment phase (Phase~2).
The pairing design tests whether agents generalize to new phrasings of the
same underlying task rather than memorising specific query surface forms.
Crucially, neither set of utterances was disclosed to participants at any
point during the competition; teams received only their aggregated execution
scores from the public leaderboard, with no visibility into the underlying
questions being evaluated.

Scenarios are organized into five functional categories that reflect the core
task domains of industrial asset operations.
\emph{IoT} scenarios test basic data retrieval from sensor streams and device
registries, requiring agents to resolve equipment identifiers, sensor
channels, and time ranges.
\emph{Failure Mode and Sensor Relation} (FMSR) scenarios require agents to
reason over fault signatures and match them to detectable sensor modalities,
combining domain knowledge with structured lookup.
\emph{Time Series Forecasting Model} (TSFM) scenarios probe capability
awareness, testing whether the agent accurately reports which forecasting
algorithms are available in the system.
\emph{Work Order} (WO) scenarios span a range of complexity from simple
record retrieval to multi-constraint maintenance scheduling.
\emph{End-to-End} (E2E) scenarios chain multiple sub-tasks including sensor
enumeration, failure mode lookup, and constraint-based filtering into a
single query, serving as holistic tests of multi-step reasoning capability.
This categorical spread was designed to cover distinct layers of the asset
operations stack, intentionally varying both task complexity and required
domain knowledge.

Tables~\ref{tab:dev_utterances} and~\ref{tab:eval_utterances} list all 11
development and evaluation utterances respectively, grouped by functional
category with \texttt{midrule} separators.

\begin{table*}[t]
\centering
\caption{Phase~1 (development) utterances grouped by functional category.
\textbf{Q} denotes the original question identifier from the
\textsc{AssetOpsBench} question bank.
These utterances were hidden from all participants throughout the
competition window.}
\small
\renewcommand{\arraystretch}{1.30}
\setlength{\tabcolsep}{8pt}
\begin{tabular}{c l p{10.2cm}}
\toprule
\textbf{Q} & \textbf{Category} & \textbf{Development Utterance (Phase~1)} \\
\midrule
Q5   & IoT  & Retrieve metadata for Chiller~6 located at the MAIN site. \\
Q8   & IoT  & Download sensor data for Chiller~6's Tonnage from the last
               week of 2020 at the MAIN site. \\
\midrule
Q114 & FMSR & What are the failure modes of Chiller~6 that can be identified
               by analysing the data from the available sensors? \\
Q106 & FMSR & List all failure modes of Chiller~6 that can be detected by
               Chiller~6 Supply Temperature. \\
\midrule
Q203 & TSFM & Are any time series forecasting models supported? \\
Q204 & TSFM & Is TTM (Tiny Time Mixture), a time series model supported? \\
\midrule
Q400 & WO   & Get the work order of equipment CWC04013 for year 2017. \\
Q405 & WO   & Get all the events of equipment CWC04009 for June 2020 and
               provide a summary by event group (work order, alert, and
               anomaly). \\
Q424 & WO   & Can you provide guidance on bundling corrective work orders
               for Chiller~9 (CWC04009) covering 2017--2019, where a bundle
               must occur within two weeks and contain at least two work
               orders? \\
\midrule
Q604 & E2E  & List all failure modes of Chiller~6 at the MAIN site that
               can be detected by Chiller~6 Chiller Efficiency. \\
Q607 & E2E  & Get the failure modes for Chiller~6 at the MAIN site and
               include only those that can be monitored using the available
               sensors. \\
\bottomrule
\end{tabular}
\label{tab:dev_utterances}
\end{table*}

\begin{table*}[t]
\centering
\caption{Phase~2 (evaluation) utterances grouped by functional category.
\textbf{Q} denotes the original question identifier from the
\textsc{AssetOpsBench} question bank.
These utterances were held out entirely and used exclusively for final
scoring; they were not shared with participants before or during the
competition.}
\small
\renewcommand{\arraystretch}{1.30}
\setlength{\tabcolsep}{8pt}
\begin{tabular}{c l p{10.2cm}}
\toprule
\textbf{Q} & \textbf{Category} & \textbf{Evaluation Utterance (Phase~2)} \\
\midrule
Q7   & IoT  & Download the metadata for Chiller~3 at the MAIN facility. \\
Q11  & IoT  & Download all sensor data for Chiller~6 from the last week of
               April~'20 at the MAIN site. \\
\midrule
Q107 & FMSR & List all failure modes of Chiller~6 that can be detected by
               temperature sensors. \\
Q108 & FMSR & List all failure modes of Chiller~6 that can be detected by
               temperature sensors and power input sensors. \\
\midrule
Q201 & TSFM & What types of time series analysis are supported? \\
Q205 & TSFM & Is LSTM model supported in TSFM? \\
\midrule
Q403 & WO   & Retrieve the corrective work order details for equipment
               CWC04013 for the year 2017. \\
Q410 & WO   & Get all the events of equipment CWC04009 for the first week
               of June 2020 and provide a summary by event group
               (work order, alert, anomaly). \\
Q411 & WO   & Which corrective work orders for CWC04009 in 2017 can be
               bundled in the next maintenance window (within two weeks)? \\
\midrule
Q605 & E2E  & List all failure modes of Chiller~6 at MAIN site that can be
               detected by temperature sensors. \\
Q606 & E2E  & List all failure modes of Chiller~6 at MAIN site that can be
               detected by temperature sensors and power input sensors. \\
\bottomrule
\end{tabular}
\label{tab:eval_utterances}
\end{table*}

Table~\ref{tab:scenario_similarity} quantifies the semantic distance between
each development--evaluation pair using cosine similarity of Sentence-BERT
embeddings.
Cells are colour-coded into three bands for readability:
\colorbox{simhigh}{\strut high similarity} ($\geq\!0.90$),
\colorbox{simmed}{\strut medium similarity} ($0.75$--$0.90$), and
\colorbox{simlow}{\strut low similarity} ($<\!0.75$).
The mean inter-phase similarity across all 11 pairs is $0.830$
($\sigma = 0.138$), reflecting the deliberate design choice to vary surface
form while preserving underlying task intent.

The similarity range is wide by design.
TSFM pairs ($0.555$ and $0.617$) represent the most demanding generalization
test: the shift from ``are models supported?'' to ``what types of analysis
are supported?'', and from one specific model name (TTM) to another (LSTM),
requires agents to understand system capabilities generically rather than
pattern-match on model names.
Work Order pairs span the widest within-category range ($0.764$--$0.984$):
Q405/Q410 are near-paraphrases differing only in temporal granularity (full
month vs.\ first week of June), while Q424/Q411 differ substantially in
phrasing and constraint specification.
FMSR pairs are the highest-similarity category ($0.914$ and $0.959$),
reflecting that the task structure is largely preserved and only the sensor
specification scope varies (specific named sensor to broader sensor class).
E2E pairs ($0.880$ and $0.943$) introduce variation through attribute change
(Chiller Efficiency vs.\ temperature sensors) while keeping the multi-step
reasoning chain intact.
IoT pairs ($0.797$ and $0.868$) vary equipment identifier and time
specification while preserving retrieval intent.

\begin{table}[h]
\centering
\caption{Cosine similarity (Sentence-BERT) between Phase~1 (development)
and Phase~2 (evaluation) utterances for each scenario pair.
\textbf{Dev.\ Q} and \textbf{Eval.\ Q} are the original question identifiers
from the question bank.
Colour bands: \colorbox{simhigh}{\strut$\;\geq\!0.90\;$} high,
\colorbox{simmed}{\strut$\;0.75\text{--}0.90\;$} medium,
\colorbox{simlow}{\strut$\;<\!0.75\;$} low.
Mean and standard deviation are reported at the foot.}
\small
\renewcommand{\arraystretch}{1.25}
\begin{tabular}{l c c c}
\toprule
\textbf{Category} & \textbf{Dev.\ Q} & \textbf{Eval.\ Q}
  & \textbf{Similarity} \\
\midrule
IoT  & Q5   & Q7   & \cellcolor{simmed}  0.797 \\
IoT  & Q8   & Q11  & \cellcolor{simmed}  0.868 \\
FMSR & Q114 & Q107 & \cellcolor{simhigh} 0.914 \\
FMSR & Q106 & Q108 & \cellcolor{simhigh} 0.959 \\
TSFM & Q203 & Q201 & \cellcolor{simlow}  0.617 \\
TSFM & Q204 & Q205 & \cellcolor{simlow}  0.555 \\
WO   & Q400 & Q403 & \cellcolor{simmed}  0.855 \\
WO   & Q405 & Q410 & \cellcolor{simhigh} 0.984 \\
WO   & Q424 & Q411 & \cellcolor{simmed}  0.764 \\
E2E  & Q604 & Q605 & \cellcolor{simhigh} 0.943 \\
E2E  & Q607 & Q606 & \cellcolor{simmed}  0.880 \\
\midrule
\multicolumn{3}{l}{\textit{Mean}}         & 0.830 \\
\multicolumn{3}{l}{\textit{Std.\ dev.}}   & 0.138 \\
\bottomrule
\end{tabular}
\label{tab:scenario_similarity}
\end{table}

% ─────────────────────────────────────────────────────────────────────────────
\subsection{Submission Feedback Design}
\label{sec:app:feedback}
% ─────────────────────────────────────────────────────────────────────────────

Upon each submission during Phase~1, participants received two levels of
feedback calibrated to support iterative agent improvement without compromising
evaluation integrity.

\subsubsection{Quantitative Score}

Each submission was assigned a single aggregate \emph{Task Completion} score
measuring the proportion of scenarios in which the agent successfully
completed the required task.
This score formed the basis of the public leaderboard ranking and provided
participants with an immediate signal of overall agent performance relative
to other teams.
The score alone, however, cannot indicate which aspect of agent behaviour
drove improvements or regressions between runs, which motivates the second
feedback component.

\subsubsection{Qualitative Diagnostic Feedback}

In addition to the aggregate score, participants received a structured
diagnostic report comprising two components: a set of skill-level metrics
and a failure mode distribution.

The first component reported six qualitative metrics across predefined
\emph{skill dimensions}, each evaluated per scenario and reported as a binary
pass or fail.
The six dimensions were selected to assess complementary and orthogonal
aspects of agent behaviour.
\emph{Task Completion} measures end-to-end success at the scenario level.
\emph{Data Retrieval Accuracy} isolates factual grounding, specifically
whether the agent retrieved correct values from the underlying data sources.
\emph{Generalized Result Verification} tests whether outputs were validated
against known constraints before being returned.
\emph{Agent Sequence Correct} captures procedural correctness of tool
invocation order, penalising agents that reached a correct answer via an
incorrect execution path.
\emph{Clarity and Justification} assesses whether the agent's output was
coherent and appropriately motivated.
\emph{Hallucinations} flags content fabricated by the agent that was not
grounded in the available data.
Together these six dimensions provide a diagnostic profile that allows
participants to distinguish, for example, an agent that completes tasks
but hallucinates from one that is factually grounded but sequentially
incorrect, enabling targeted rather than undirected improvement between
submissions.

Table~\ref{tab:skill_metrics} shows an example aggregated summary returned
after a single illustrative submission.
The agent in this example performs strongest on \emph{Clarity and
Justification} (pass rate 90.9\%), indicating that its outputs are
well-formed and coherently motivated, and on \emph{Agent Sequence Correct}
(81.8\%), indicating that when tasks are completed the tool invocation order
is largely correct.
The weakest dimension is \emph{Data Retrieval Accuracy} (63.6\%), pointing
to systematic factual grounding failures, and the agent produces
hallucinations in two of the eleven scenarios (18.2\%).
A participant reading this summary can immediately prioritise data retrieval
grounding as the highest-leverage improvement target.

\begin{table}[h]
\centering
\caption{Example skill metric summary returned to participants after a
single submission across all 11 development scenarios (Phase~1).
\textbf{True} indicates the criterion was satisfied; \textbf{False}
indicates it was not.
For \emph{Hallucinations}, True indicates a hallucination was detected,
which is undesirable; the pass rate for this row therefore reflects the
proportion of scenarios free of hallucination. The lower the pass rate, the better. Results shown are for an illustrative submission only.}
\small
\renewcommand{\arraystretch}{1.30}
\begin{tabular}{l c c c}
\toprule
\textbf{Criterion} & \textbf{True} & \textbf{False} & \textbf{Pass Rate} \\
\midrule
Task Completion                 & \cellcolor{truegreen}  8
                                & \cellcolor{falsered}   3 & 72.7\% \\
Data Retrieval Accuracy         & \cellcolor{truegreen}  7
                                & \cellcolor{falsered}   4 & 63.6\% \\
Generalized Result Verification & \cellcolor{truegreen}  8
                                & \cellcolor{falsered}   3 & 72.7\% \\
Agent Sequence Correct          & \cellcolor{truegreen}  9
                                & \cellcolor{falsered}   2 & 81.8\% \\
Clarity and Justification       & \cellcolor{truegreen} 10
                                & \cellcolor{falsered}   1 & 90.9\% \\
Hallucinations                  & \cellcolor{falsered}   2
                                & \cellcolor{truegreen}  9 & 18.2\% \\
\midrule
\textit{Total scenarios} & \multicolumn{3}{c}{11} \\
\bottomrule
\end{tabular}
\label{tab:skill_metrics}
\end{table}

To support deeper investigation, participants could further inspect results
at the individual question level.
Table~\ref{tab:question_level_metrics} presents the same submission broken
down per scenario, revealing the specific questions driving each aggregate
statistic.
Three questions account for the majority of failures: question~5 fails on
all criteria except \emph{Clarity and Justification}, suggesting the agent
produced a well-formed but fundamentally incorrect response; question~10
fails across all six dimensions, indicating complete task breakdown; and
question~11 fails on task completion and retrieval but partially recovers
on agent sequence and clarity.
This granularity allows teams to directly pinpoint which scenario categories
are driving score drops and to target changes accordingly, rather than
relying on aggregate statistics alone.
Question indices are uniformly assigned across runs to enable cross-run
comparison; the underlying question identifiers from the question bank
were not disclosed to participants.

\begin{table*}[t]
\centering
\caption{Question-level skill metric results for the same illustrative
submission as Table~\ref{tab:skill_metrics}, across all 11 development
scenarios (Phase~1).
Each cell indicates whether the criterion was satisfied:
$1$ = satisfied, shaded \colorbox{cgreen}{\strut green};
$0$ = not satisfied, shaded \colorbox{cpink}{\strut pink}.
For \emph{Hallucinations}, the colour logic is inverted: $1$ indicates a
hallucination was detected (undesirable) and $0$ indicates none.
The \textbf{Sum} row aggregates column totals and matches the
\textbf{True} column of Table~\ref{tab:skill_metrics}.}
\small
\renewcommand{\arraystretch}{1.25}
\setlength{\tabcolsep}{5pt}
\begin{tabular}{c c c c c c c}
\toprule
\textbf{Q}
  & \thead{Task\\Completion}
  & \thead{Data Retrieval\\Accuracy}
  & \thead{Gen. Result\\Verification}
  & \thead{Agent Sequence\\Correct}
  & \thead{Clarity \&\\Justification}
  & \thead{Halluc-\\inations} \\
\midrule
1  & \cyes & \cyes & \cyes & \cyes & \cyes & \cnoh \\
2  & \cyes & \cyes & \cyes & \cyes & \cyes & \cnoh \\
3  & \cyes & \cno  & \cyes & \cyes & \cyes & \cnoh \\
4  & \cyes & \cyes & \cyes & \cyes & \cyes & \cnoh \\
5  & \cno  & \cno  & \cno  & \cno  & \cyes & \chal \\
6  & \cyes & \cyes & \cyes & \cyes & \cyes & \cnoh \\
7  & \cyes & \cyes & \cyes & \cyes & \cyes & \cnoh \\
8  & \cyes & \cyes & \cyes & \cyes & \cyes & \cnoh \\
9  & \cyes & \cyes & \cyes & \cyes & \cyes & \cnoh \\
10 & \cno  & \cno  & \cno  & \cno  & \cno  & \cnoh \\
11 & \cno  & \cno  & \cno  & \cyes & \cyes & \chal \\
\midrule
\textbf{Sum}
  & \textbf{8} & \textbf{7} & \textbf{8}
  & \textbf{9} & \textbf{10} & \textbf{2} \\
\bottomrule
\end{tabular}
\label{tab:question_level_metrics}
\end{table*}

\begin{figure*}[t]
  \centering
  \includegraphics[width=\textwidth]{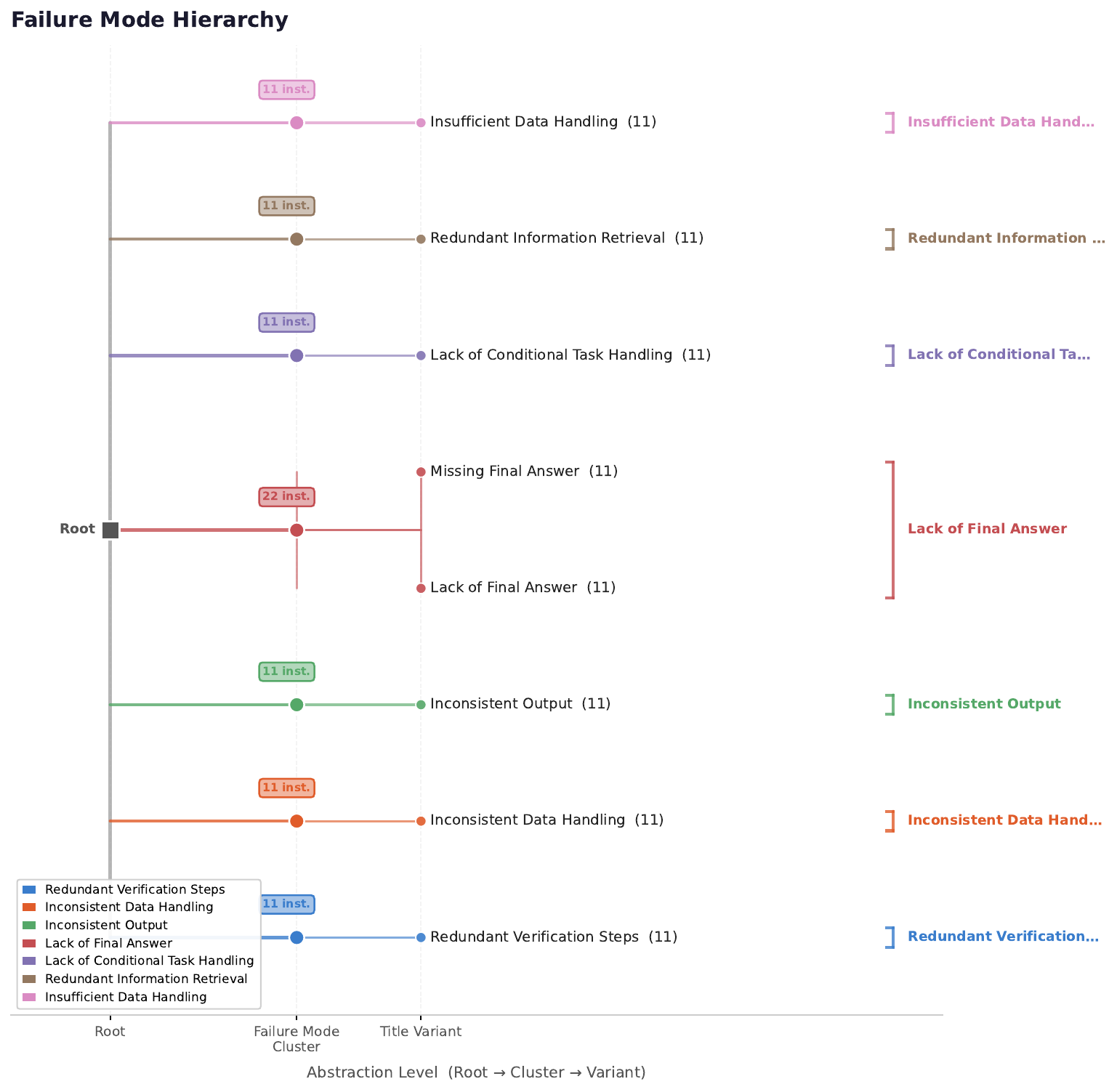}
  \caption{%
    \textbf{Hierarchical cluster structure of observed failure modes.}
    The tree is organized across three abstraction levels: a single root
    node (left), seven failure mode clusters at the middle level (one
    colour per cluster), and eight title variants at the leaf level (right).
    Count badges above each cluster node show the total number of instances
    belonging to that cluster.
    \emph{Lack of Final Answer} (cluster~3, centre) is the only cluster
    with two distinct title variants, reflecting two observationally
    distinct manifestations of the same underlying incompletion failure.%
  }
  \label{fig:failure_hierarchy}
\end{figure*}

The second feedback component was a \emph{failure mode distribution} reported
per functional category, illustrated in Figure~\ref{fig:failure_hierarchy}. Figure~\ref{fig:failure_hierarchy} renders the same clustering structure
as a hierarchical tree dendrogram, making the three-level abstraction
explicit: a single root node, seven failure mode clusters at the middle
level, and eight title variants at the leaf level.
The count badge above each cluster node immediately communicates the cluster
weight.
The design rationale for providing per-category failure mode distributions
rather than aggregate error counts is that qualitatively different scenario
categories expose qualitatively different agent weaknesses.
An agent failing on TSFM scenarios due to capability awareness errors
requires a fundamentally different fix than one failing on WO scenarios due
to multi-step data retrieval errors.
Reporting the distribution per category allows participants to direct their
debugging effort precisely rather than attempting unfocused improvements.

% =============================================================================
\section{Dimension 3: Submission Behaviour}
\label{sec:app:behaviour}
% =============================================================================

This dimension characterises how teams engaged with the benchmark over the
competition window, drawing exclusively on submission logs without access to
participant code or strategy information (which is addressed in
Dimension~5).
A total of 300 submission attempts were recorded across 25 unique team
identities, of which 234 reached \texttt{Finished} status and are used
throughout this analysis.
Scores are normalised to $[0,100]$ following the official competition rubric.

% ─────────────────────────────────────────────────────────────────────────────
\subsection{Score Distributions and Track Specialization}
\label{sec:app:score_dist}
% ─────────────────────────────────────────────────────────────────────────────

\begin{figure*}[t]
  \centering
  \includegraphics[width=\textwidth]{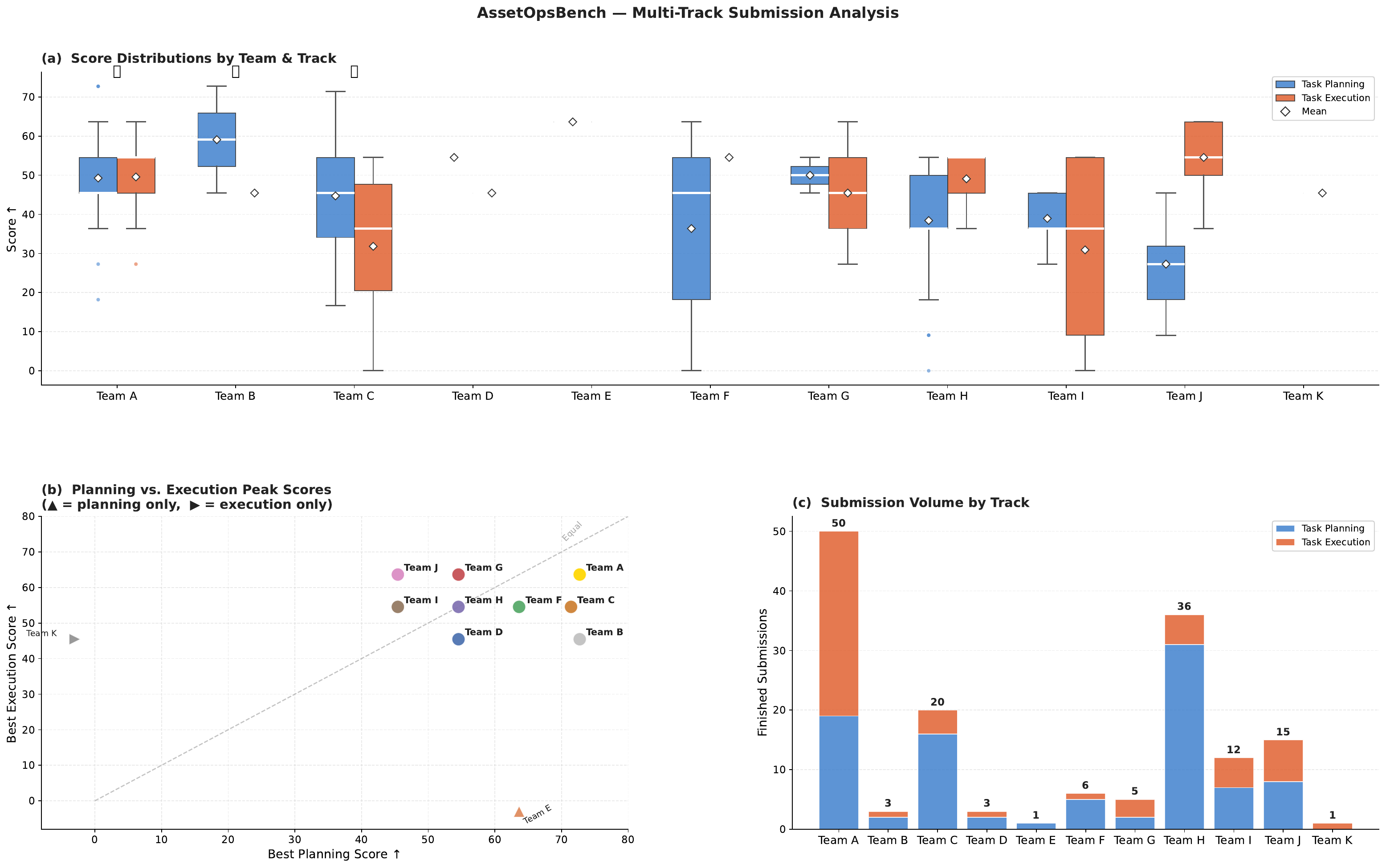}
  \caption{%
    \textbf{Competition overview.}
    \textbf{(a)} Score distributions per team split by track (blue =
    Task Planning, orange = Task Execution).
    White diamonds denote the per-team mean; medal icons identify the
    top-three finishers by final combined score.
    Task Planning exhibits wider variance (IQR up to 27 points for
    high-submission teams) than Task Execution.
    \textbf{(b)} Best planning score vs.\ best execution score for each team.
    Circles = both-track participants; upward triangles ($\triangle$) =
    planning-only; rightward triangles ($\triangleright$) = execution-only.
    The dashed diagonal marks equal performance across tracks.
    Spearman $\rho = 0.41$ ($p = 0.21$) indicates a weak non-significant
    correlation between planning and execution best scores.
    \textbf{(c)} Total finished submission counts per team, stacked by track.
    Team identifiers follow Table~\ref{tab:team_mapping}.%
  }
  \label{fig:overview}
\end{figure*}

Figure~\ref{fig:overview} provides a three-panel overview of competition
outcomes.
Panel~(a) presents side-by-side box plots for each ranked team, decomposed
by track.
The Task Planning track (blue) exhibits substantially wider score variance
than the Task Execution track (orange), with interquartile ranges spanning
up to 27 points for high-submission teams such as Team~A and Team~C,
compared to ranges below 10 points for most execution-track submissions.
This asymmetry in variance is consistent with the planning track being the
main arena of iterative strategy refinement: teams submitted more planning
runs (as confirmed by panel~c) and their score distributions reflect the
range of strategies explored.
The white diamond markers indicate per-team mean scores; in several cases the
mean lies substantially below the upper quartile, reflecting early
low-scoring submissions before teams converged on effective approaches.

Panel~(b) plots each team's best Planning score against its best Execution
score, with the dashed diagonal indicating equal performance across tracks.
Teams cluster near or below the diagonal, indicating a general tendency for
planning scores to meet or exceed execution scores.
Spearman $\rho = 0.41$ ($p = 0.21$) between best planning and best execution
scores across all teams that participated in both tracks indicates a weak and
statistically non-significant correlation.
This confirms that planning proficiency does not reliably predict execution
proficiency, suggesting the two tracks test qualitatively distinct agent
capabilities.
Teams that participated in only one track are shown as marginal markers:
upward triangles ($\triangle$) on the $x$-axis for planning-only
participants, and rightward triangles ($\triangleright$) on the $y$-axis for
execution-only participants.

Panel~(c) shows total submission volume per team disaggregated by track.
Team~A leads with 31 finished execution submissions, reflecting a sustained
iterative optimization strategy across both tracks.
Team~H concentrated almost entirely on the planning track (31 planning,
5 execution), consistent with a deliberate track specialization strategy
that produced a higher planning component score but a weaker combined score
relative to more balanced teams.

% ─────────────────────────────────────────────────────────────────────────────
\subsection{Score Progression over the Competition Window}
\label{sec:app:trajectory}
% ─────────────────────────────────────────────────────────────────────────────

\begin{figure*}[t]
  \centering
  \includegraphics[width=\textwidth]{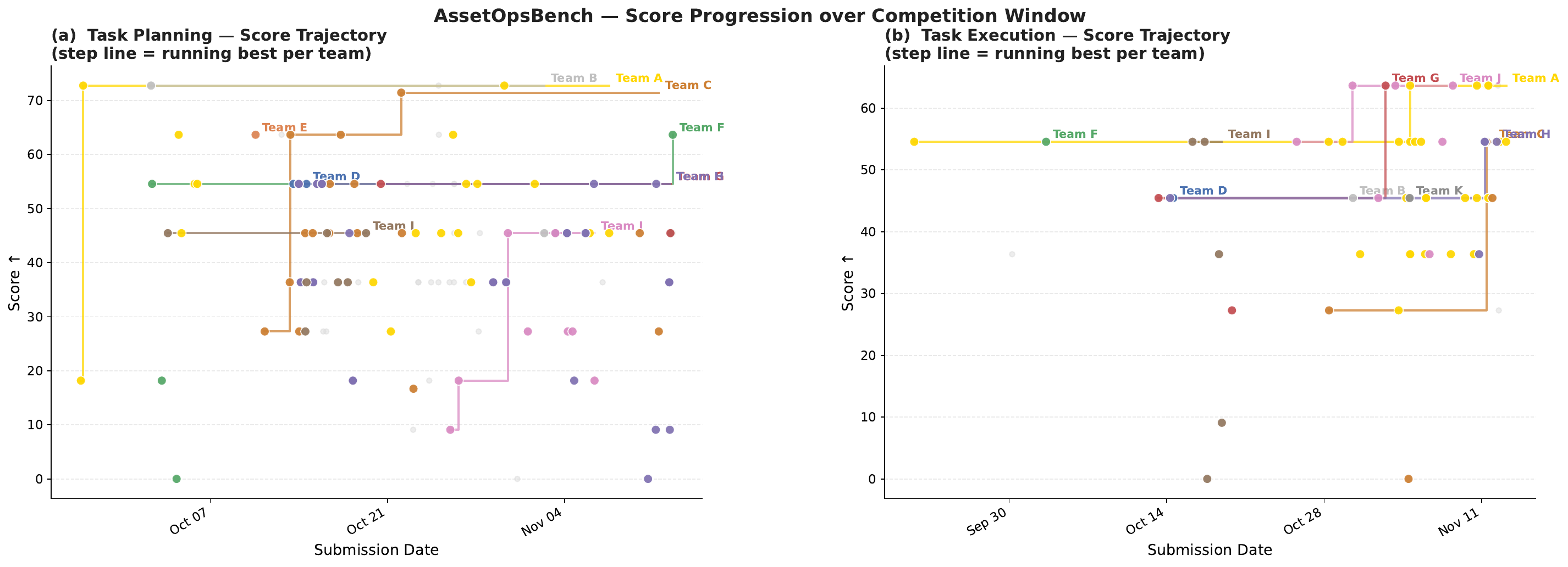}
  \caption{%
    \textbf{Score trajectories over the competition window.}
    Each dot is one finished submission; the step line shows the running
    best score for each team.
    Light grey dots in the background represent all submissions aggregated
    across teams, providing a reference for the overall scoring landscape.
    \textbf{(a)} Task Planning track: most teams converge within the first
    two weeks; Team~A is a notable exception with a sustained ascending
    trajectory reaching 72.73 in the final days.
    \textbf{(b)} Task Execution track: trajectories are flatter and more
    compressed, consistent with lower submission volumes.
    Team identifiers follow Table~\ref{tab:team_mapping}.%
  }
  \label{fig:trajectory}
\end{figure*}

Figure~\ref{fig:trajectory} shows how scores evolved over time for each
track, with each dot representing one finished submission and the step line
tracing each team's running best score.
The background grey cloud of all submissions provides a visual reference for
the overall scoring landscape.

In the Task Planning track (panel~a), the most striking feature is the rapid
convergence exhibited by most teams: the majority reached within five points
of their final best score within the first two weeks of the competition
window, with subsequent submissions producing only marginal improvements.
This pattern is consistent with a strategy space dominated by prompt
engineering variations on a fixed scaffold, where initial configurations
already capture most of the available score and later submissions refine
without fundamentally changing the approach.
Team~A is a clear exception to this pattern, maintaining a steadily
ascending trajectory throughout the competition window and reaching its
best score of 72.73 only in the final days.
This sustained improvement trajectory suggests that Team~A's later
submissions reflected broader implementation iteration rather than
incremental prompt tuning alone, a hypothesis supported by the verified
code analysis in Dimension~5.

In the Task Execution track (panel~b), score trajectories are generally
flatter and more compressed, consistent with the lower submission volumes
observed in this track.
Several teams recorded their highest execution scores late in the competition,
suggesting that insights gained during planning-track iterations transferred
to the execution setting in the later stages of the window.

% ─────────────────────────────────────────────────────────────────────────────
\subsection{Team Activity Patterns}
\label{sec:app:activity}
% ─────────────────────────────────────────────────────────────────────────────

\begin{figure*}[t]
  \centering
  \includegraphics[width=\textwidth]{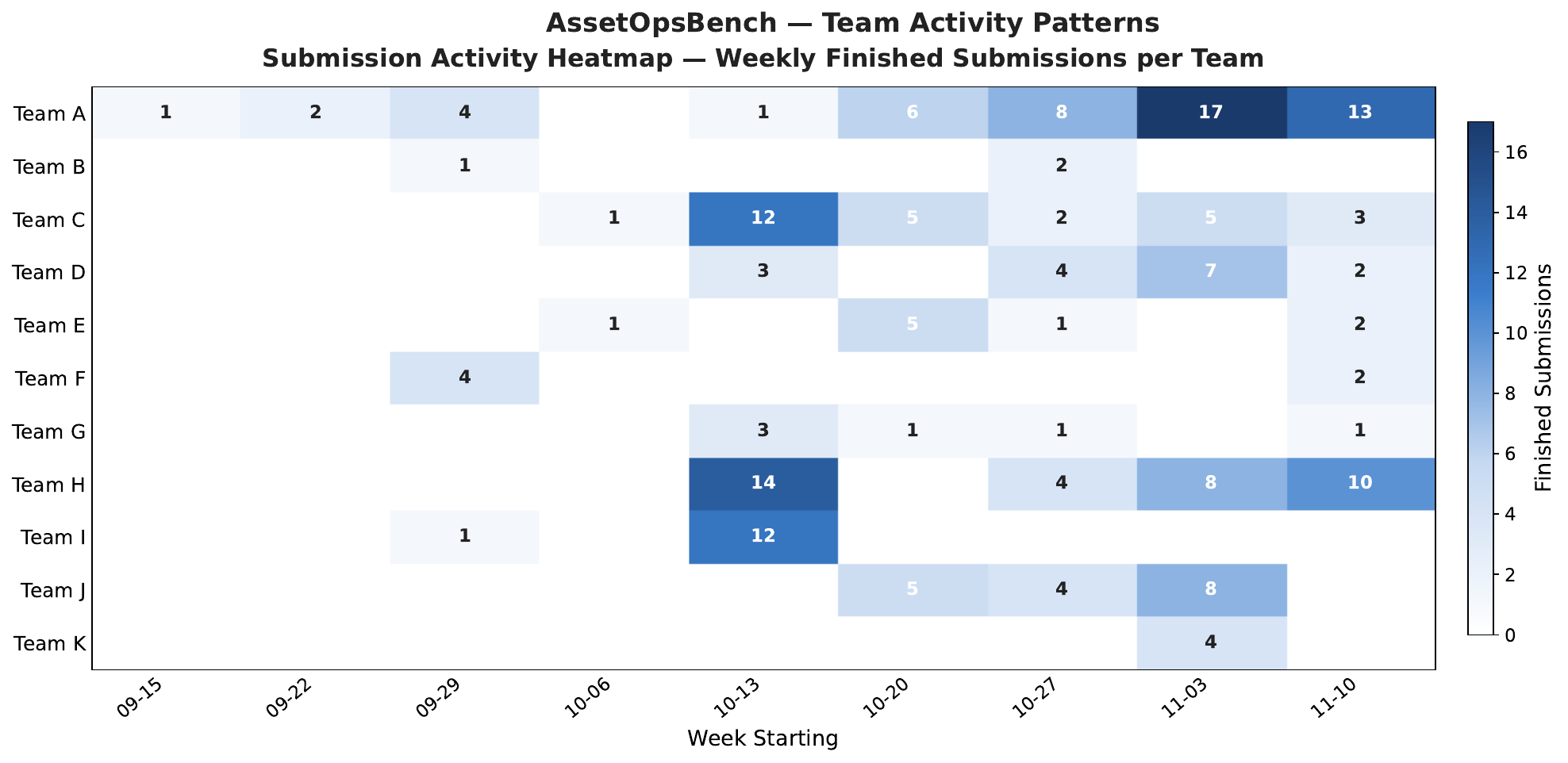}
  \caption{%
    \textbf{Weekly submission activity heatmap.}
    Each cell records the number of \texttt{Finished} submissions by a given
    team (rows) during a given calendar week (columns).
    Colour intensity encodes volume: white = zero submissions,
    dark blue = highest activity.
    Cell values are annotated directly; values ${\geq}5$ are printed in
    white for contrast.
    Two distinct engagement patterns are visible: sustained high-activity
    teams (Team~A, Team~C) versus late-concentrated teams (Teams~K, E).
    A competition-wide deadline surge is visible in the final two weeks.
    Team identifiers follow Table~\ref{tab:team_mapping}.%
  }
  \label{fig:heatmap_weekly}
\end{figure*}

Figure~\ref{fig:heatmap_weekly} presents a weekly submission heatmap in which
each cell records the number of finished submissions made by a given team
during a given calendar week.
Colour intensity encodes volume, ranging from white (zero submissions) to
dark blue (high activity), with count values annotated directly inside each
cell.

The heatmap reveals two clearly distinct engagement patterns across the
11 teams.
The first pattern is sustained high-activity engagement: Team~A exhibits the
most consistent activity, peaking at 11 finished submissions in a single
week and maintaining non-zero activity across nearly every week of the
competition window; Team~C similarly sustains activity across multiple weeks
with a pronounced surge in the closing days.
The second pattern is concentrated late-stage engagement: Teams~K and~E
show sparse or absent activity for most of the competition and concentrate
their submissions in the final two weeks, consistent with a late-entry
strategy that may have limited the number of improvement iterations available
before the deadline.

A competition-wide surge in submission volume is visible in the final two
weeks across nearly all teams, reflecting the typical deadline-driven
acceleration as teams exhaust remaining submission budget.
This late surge was particularly pronounced for Teams~A and~C, both of which
submitted multiple refinements in the closing days, suggesting continued
active development rather than merely consuming remaining attempts.

% ─────────────────────────────────────────────────────────────────────────────
\subsection{Learning Dynamics and Submission Reliability}
\label{sec:app:learning}
% ─────────────────────────────────────────────────────────────────────────────

\begin{figure*}[t]
  \centering
  \includegraphics[width=\textwidth]{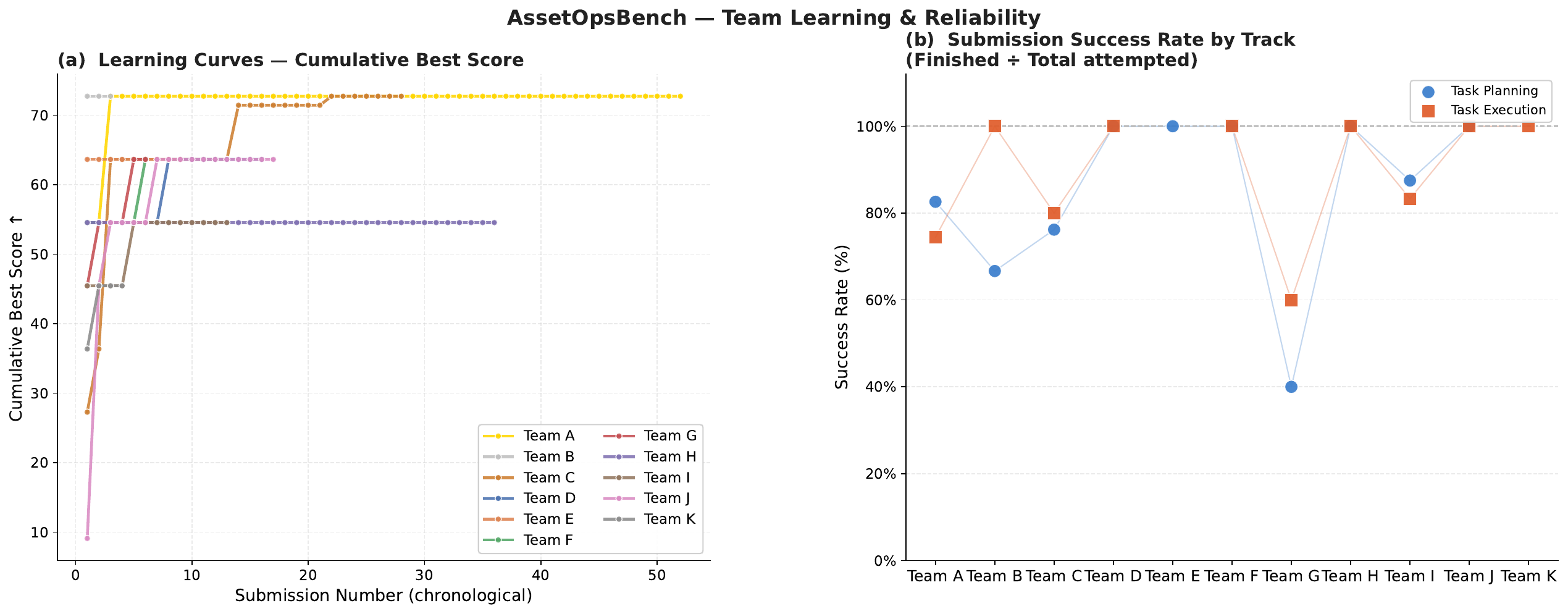}
  \caption{%
    \textbf{Learning dynamics and submission reliability.}
    \textbf{(a)} Cumulative best score as a function of submission number,
    ordered chronologically within each team.
    Each marker is one finished submission; lines connect consecutive
    submissions per team.
    Most teams plateau within five submissions; Team~A is the notable
    exception with continued improvement past submission~20.
    Team~H plateaus early despite 31 total submissions, illustrating a
    diminishing-returns failure mode.
    \textbf{(b)} Submission success rate (Finished $\div$ Total~attempted)
    per team per track.
    Circles~($\circ$) = Task Planning; squares~($\square$) = Task Execution.
    Task Planning achieves ${\geq}80\%$ success rates for most teams;
    Task Execution shows substantially higher variability.
    Team identifiers follow Table~\ref{tab:team_mapping}.%
  }
  \label{fig:learning}
\end{figure*}

Figure~\ref{fig:learning} examines two complementary aspects of team
behaviour: how quickly teams improved over their submission sequence, and
how reliably their submissions completed without errors.

Panel~(a) plots cumulative best score against submission number for each
team.
The most prominent feature is the rapid initial improvement followed by
plateau that characterises most teams: the majority achieved their best score
within the first five finished submissions, consistent with an initial
strategy that already captured most of the available score and subsequent
submissions producing diminishing returns.
This rapid saturation is a meaningful benchmark design observation: it
suggests that in a competition with 11 binary-graded scenarios, the
achievable score space is discrete and limited, and early submissions
quickly exhaust the distinguishable performance levels.
Team~A is a clear exception, continuing to improve beyond submission~20 and
reaching its highest score only in the closing days; as noted above, this
sustained improvement is consistent with broader implementation iteration.
Team~H presents the starkest contrasting case: its cumulative best score
plateaus at submission~5 despite accumulating the largest total number of
finished submissions in the planning track (31 submissions), indicating a
failure mode of repeated marginal refinements that do not address the
underlying causes of remaining failures.

Panel~(b) reports the submission success rate, defined as the proportion of
non-cancelled attempts that reached \texttt{Finished} status, disaggregated
by track.
Task Planning submissions achieved consistently high success rates across all
teams, with most exceeding 80\%, reflecting the well-defined output format
requirements of planning tasks.
Task Execution submissions show substantially greater variability, with
several teams experiencing failure rates exceeding 30\%, attributable to
environment-interaction errors that are more frequent when agents must
execute actions in a live system rather than produce structured plans.
Teams~B and~I recorded the lowest execution success rates among all
participants, which partly explains their conservative submission volumes
in the execution track and their lower execution component scores in the
final leaderboard.

\paragraph{Summary.}
Three findings emerge consistently from Dimension~2.
First, planning and execution performance are weakly and non-significantly
correlated (Spearman $\rho = 0.41$, $p = 0.21$), confirming that the two
tracks test qualitatively distinct agent capabilities and that cross-track
generalization is a meaningful challenge.
Second, most teams saturate within five submissions, suggesting that the
score space is quickly exhausted under incremental prompt refinement;
sustained improvement appears to require broader implementation iteration,
as illustrated by Team~A's trajectory.
Third, execution-track submission reliability is substantially lower than
planning-track reliability, confirming that environment interaction introduces
engineering overhead invisible in planning-only evaluations and should be
accounted for in competition design and agent development time estimates.
These findings motivate the next dimension's examination of whether the
resulting ranking is a reliable reflection of the quality differences these
behavioural patterns produced.

% =============================================================================
\section{Dimension 4: Ranking Robustness}
\label{sec:app:robustness}
% =============================================================================

This dimension evaluates whether the official leaderboard faithfully reflects
hidden-scenario robustness under the released scoring rule by examining three
distinct threats to ranking validity: sensitivity of the top-ranked team to
metric specification, resolution of the public evaluation signal, and
alignment between public and hidden scores.
These threats are particularly important for industrial agent benchmarks,
where composite scoring functions with multiple weighted components can
produce rankings sensitive to implementation choices that participants have
no visibility into.

% ─────────────────────────────────────────────────────────────────────────────
\subsection{Final Competition Rankings}
\label{sec:app:final_rank}
% ─────────────────────────────────────────────────────────────────────────────

Table~\ref{tab:final_rank_appendix} reproduces the 11 valid rows of the
released final ranking sheet after dropping blank lines, making the
cross-track aggregation and per-track score components explicit.
The final score $F$ is a weighted combination of planning component score
$C_{\text{plan}}$ and execution component score $C_{\text{exec}}$, each of
which incorporates the semantic \textit{t-match} term under fixed weights.

A notable structural feature of the ranking is the near-tie at ranks~2
and~4: BlueCube achieves the higher planning component score (60.049, equal
to WaterLevel) yet ranks two positions lower due to its substantially weaker
execution component (46.406 vs.\ 54.593).
This illustrates a property of the aggregation formula that penalises track
specialization and rewards balanced cross-track performance, even when
planning-track scores are equal.
The separation between ranks~3 and~4 is remarkably small (LostSouls:
51.864, BlueCube: 51.863), a gap of 0.001 points that would be reversed by
any minor perturbation of the scoring function and motivates the stability
analysis that follows.

\begin{table*}[h]
\centering
\caption{Released final team ranking after dropping blank rows.
Planning and execution owners identify the accounts from which the selected
best submissions were drawn.
The final score $F$ is a weighted combination of planning and execution
component scores; both components incorporate the semantic \textit{t-match}
term under fixed weights.
Gold, silver, and bronze rows indicate the top-three finishers.}
\small
\renewcommand{\arraystretch}{1.20}
\begin{tabular}{r l l l r r r}
\toprule
\textbf{Rank} & \textbf{Team} & \textbf{Planning owner}
  & \textbf{Execution owner}
  & $C_{\text{plan}}$ & $C_{\text{exec}}$ & \textbf{Final} $F$ \\
\midrule
\rowcolor{gold!60}
1  & Smart Maint.\ Crew & \texttt{vamsikv28}
   & \texttt{shashank\_1904} & 56.528 & 57.318 & 57.002 \\
\rowcolor{silver!50}
2  & WaterLevel         & \texttt{kanishk\_007}
   & \texttt{harshvardhan1}  & 60.049 & 54.593 & 56.775 \\
\rowcolor{bronze!40}
3  & LostSouls          & \texttt{h1t35h}
   & \texttt{h1t35h}         & 51.857 & 51.869 & 51.864 \\
4  & BlueCube           & \texttt{rohith\_arumugam}
   & \texttt{samah}          & 60.049 & 46.406 & 51.863 \\
5  & Scalar\_nitk       & \texttt{scalar\_anjali}
   & \texttt{scalar\_anjali} & 49.141 & 51.863 & 50.774 \\
6  & Entropians         & \texttt{supminal}
   & \texttt{supminal}       & 54.580 & 46.408 & 49.677 \\
7  & Infinity           & \texttt{abhinf104}
   & \texttt{abhinf104}      & 49.137 & 49.139 & 49.138 \\
8  & aviation\_agent    & \texttt{shoeb}
   & \texttt{shoeb}          & 54.598 & 43.682 & 48.048 \\
9  & horizon            & \texttt{horizon22}
   & \texttt{horizon22}      & 32.768 & 57.323 & 47.501 \\
10 & kinatic            & \texttt{vinaykarman}
   & \texttt{subhadeep}      & 43.680 & 40.956 & 42.046 \\
11 & EXL Health AI Lab  & \texttt{uthrasuresh}
   & \texttt{uthrasuresh}    & 32.767 & 43.680 & 39.315 \\
\bottomrule
\end{tabular}
\label{tab:final_rank_appendix}
\end{table*}

% ─────────────────────────────────────────────────────────────────────────────
\subsection{Top-Rank Stability Under Metric Reparameterization}
\label{sec:app:stability}
% ─────────────────────────────────────────────────────────────────────────────

To evaluate the robustness of the published ranking, we perform a post-hoc
sensitivity analysis of the scoring function.
We systematically vary two key factors: (i) the relative weight between
planning and execution scores, sweeping execution weight $\alpha \in [0,1]$
with planning weight $= 1 - \alpha$; and (ii) the numerical scale of the
\textit{t-match} component, rescaling $\tau$ by a factor $s \in [1,100]$.
For each $(\alpha, s)$ configuration we recompute the final score for all
11 teams, record the top-ranked team, and compute Kendall's $\tau$ between
the resulting ranking and the official ranking.
This produces both a top-rank identity map and a continuous robustness surface
over the parameter space.

\begin{figure}[!ht]
\centering
\includegraphics[width=\linewidth]{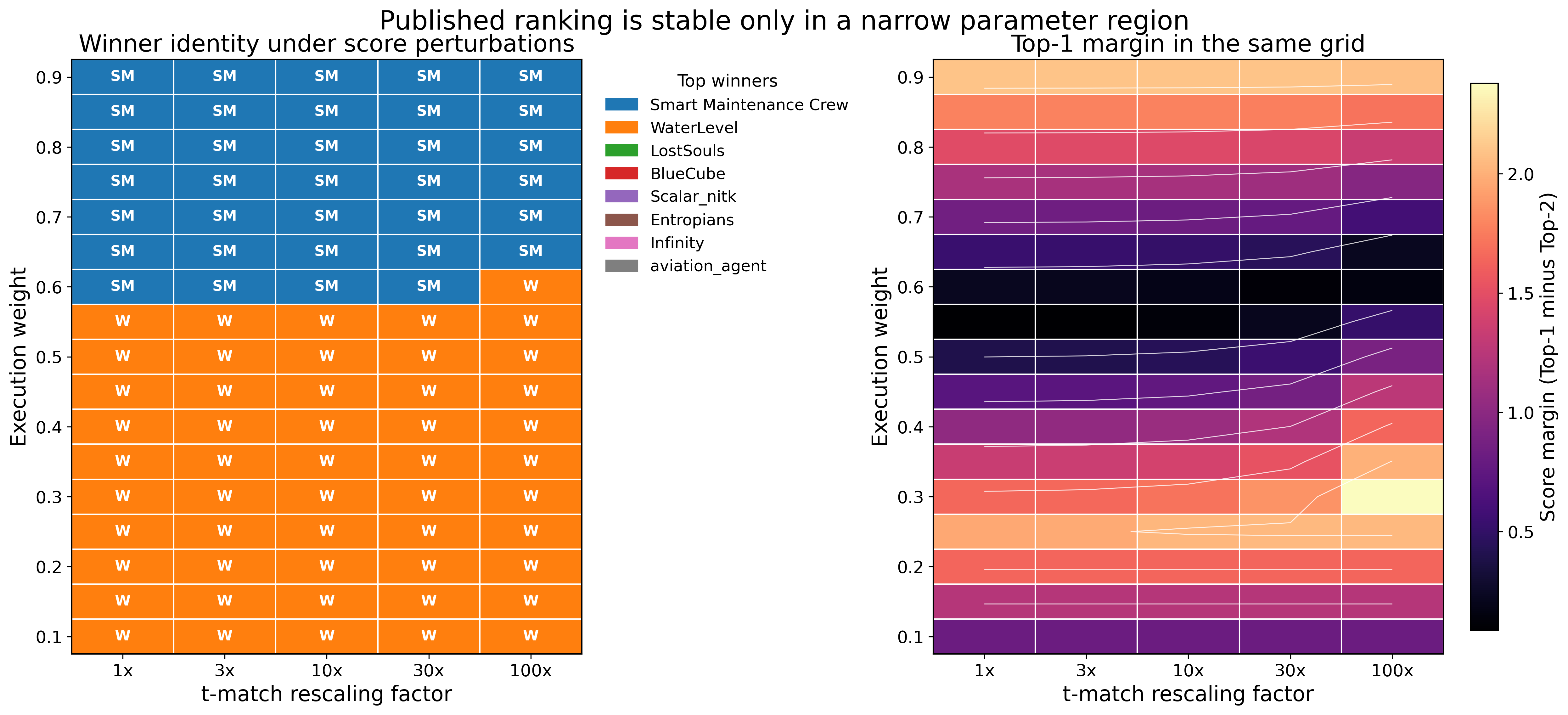}
\caption{%
  \textbf{Top-rank stability under score reparameterization.}
  Each cell shows the identity of the top-ranked team for a given
  combination of execution weight $\alpha$ and \textit{t-match} scaling
  factor $s$.
  The officially top-ranked team is stable only within a narrow band around
  the released configuration; the top-ranked team changes across large regions of
  the parameter space.
  Mean Kendall $\bar{\tau} = 0.61$ ($\text{SD} = 0.19$) between the
  official ranking and rankings under each alternative configuration,
  indicating moderate robustness.%
}
\label{fig:winner_stability}
\end{figure}

Figure~\ref{fig:winner_stability} shows the top-rank identity map.
The officially top-ranked team is stable only within a narrow band around
the released scoring configuration; modest changes in either $\alpha$ or $s$
lead to different teams being ranked first across large regions of the
parameter space.
The mean Kendall $\tau$ across the full $(\alpha, s)$ sweep is
$\bar{\tau} = 0.61$ ($\text{SD} = 0.19$), indicating moderate but not high
concordance between the official ranking and rankings under alternative
specifications.
The \textit{t-match} scaling factor $s$ has a larger effect on rank changes
than the planning-execution weight $\alpha$, because $s$ alters the absolute
magnitude of a term that is otherwise numerically dominated by the raw score
components: when $s$ is small, \textit{t-match} is effectively irrelevant;
when $s$ is large, it dominates the final score and reorders teams according
to a different quality dimension.
This sensitivity to $s$ is a design observation that future competitions
should address by normalising all composite score components to a common
scale before aggregation, or by reporting a robustness check over a range
of weight configurations alongside the official ranking.

% ─────────────────────────────────────────────────────────────────────────────
\subsection{Score Saturation of the Public Leaderboard}
\label{sec:app:saturation}
% ─────────────────────────────────────────────────────────────────────────────

\begin{figure}[t]
\centering
\includegraphics[width=\linewidth]{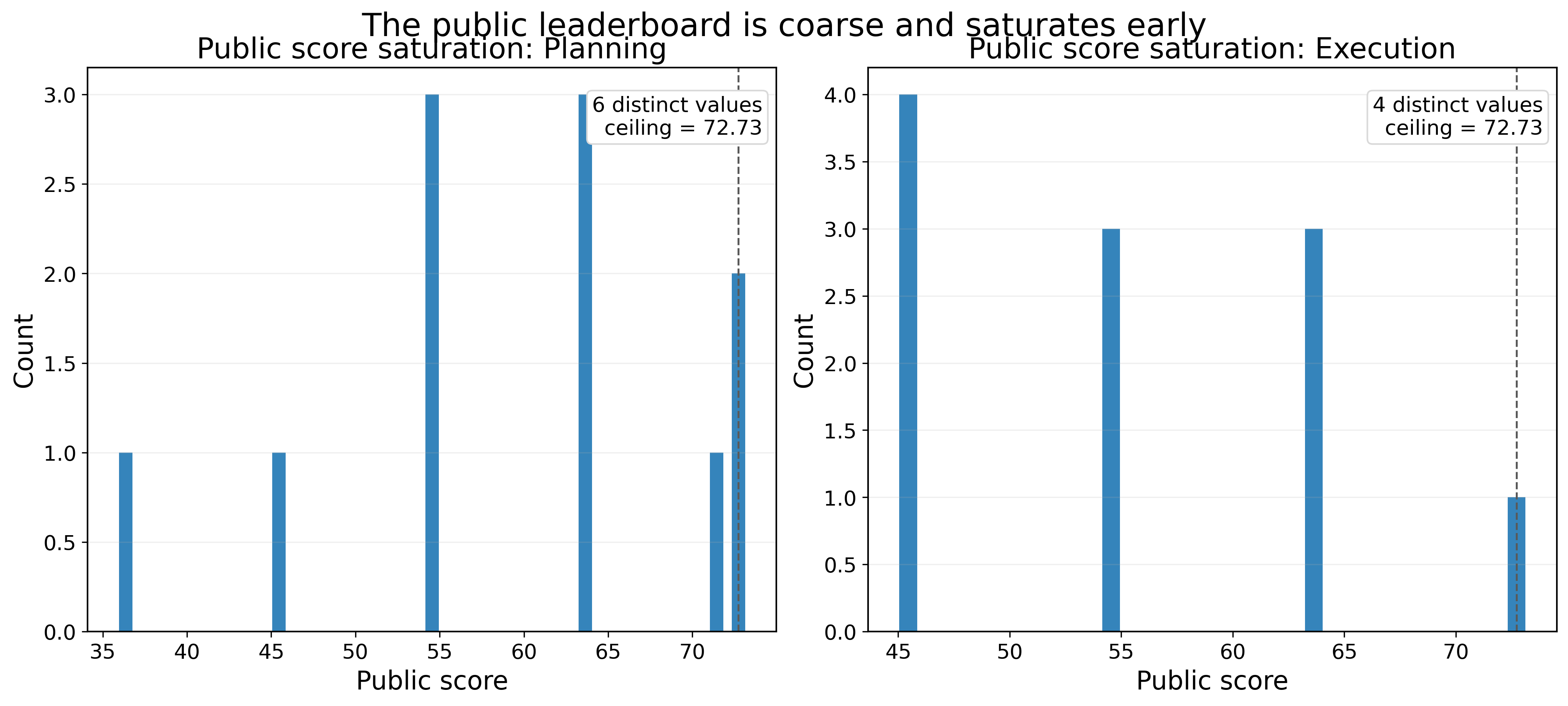}
\caption{%
  \textbf{Distribution of public leaderboard scores.}
  Scores cluster at a small set of discrete values arising from binary
  per-scenario evaluation ($\sim\!9.09$ points per scenario with 11
  scenarios total).
  Four teams share the maximum planning-track score of 72.73, illustrating
  that the public evaluation signal cannot discriminate between these teams
  on the planning track; a second cluster appears at 63.64 points.
  This saturation means the execution component is the decisive
  discriminator for the top-four ranking positions.%
}
\label{fig:saturation}
\end{figure}

Figure~\ref{fig:saturation} shows the distribution of public leaderboard
scores across all teams.
The most significant feature is the saturation at the top of the
planning-track score distribution: four teams share the maximum public
planning score of 72.73, meaning the public leaderboard cannot discriminate
between these teams on the planning track at all.
Their relative ordering in the final combined ranking is determined entirely
by their execution component scores, making the execution track, despite
having lower submission volumes and higher engineering overhead, the decisive
factor for the top-four ranking positions.
Multiple other teams cluster at 63.64 points in the planning track, a second
saturation tier.
This coarse-grained discrete scoring structure arises from the binary
per-scenario evaluation: with 11 scenarios each contributing approximately
9.09 points, the achievable public scores form a fixed discrete grid.
Future benchmark editions should increase the scenario count or introduce
partial credit scoring to improve discriminative resolution, particularly
for separating high-performing systems that are indistinguishable under the
current binary evaluation scheme.

% ─────────────────────────────────────────────────────────────────────────────
\subsection{Public vs.\ Hidden Score Alignment}
\label{sec:app:pub_hidden}
% ─────────────────────────────────────────────────────────────────────────────

To quantify the extent to which hidden evaluation alters leaderboard
outcomes, we compute rank shifts $\Delta r = r_{\text{hidden}} -
r_{\text{public}}$ for each team separately for the planning and execution
tracks, and report Spearman $\rho$ between public and hidden scores within
each track.

\begin{figure}[!ht]
\centering
\includegraphics[width=\linewidth]{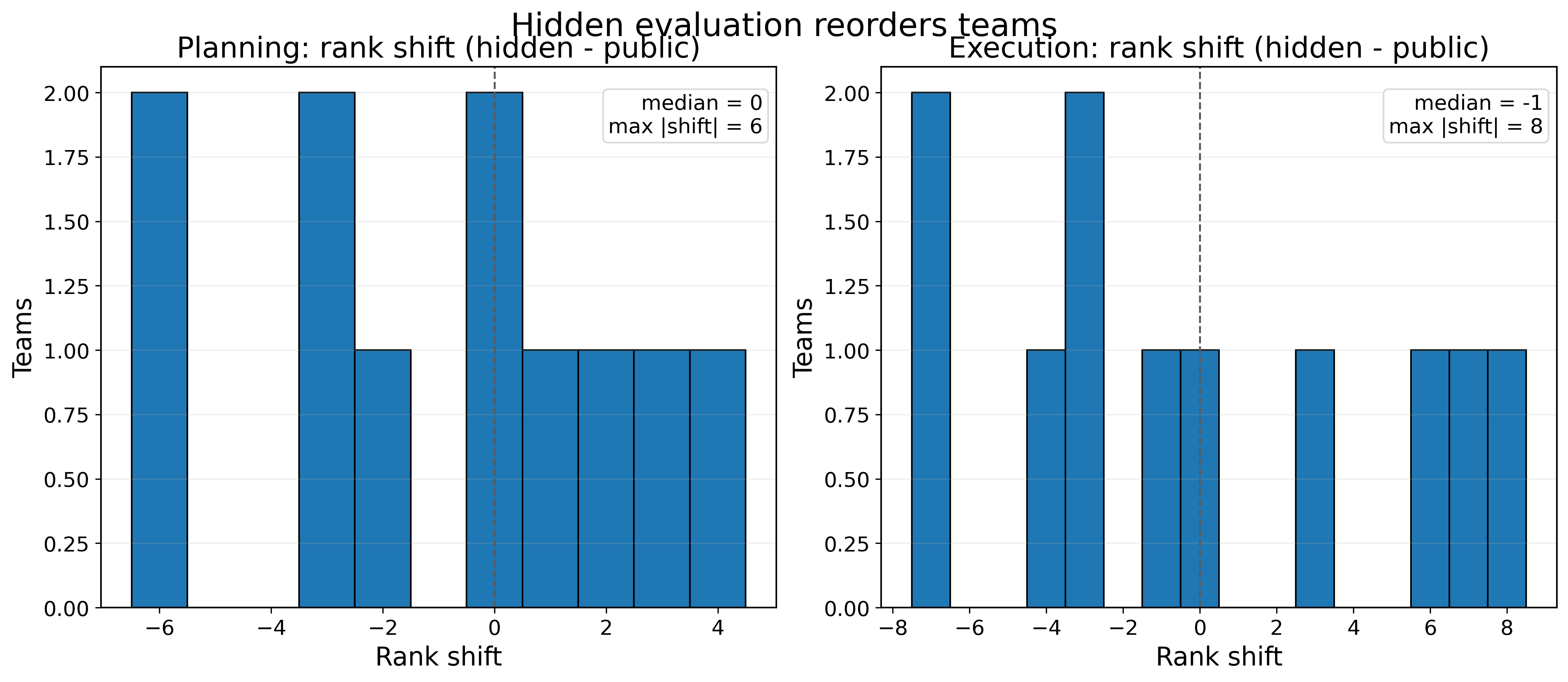}
\caption{%
  \textbf{Rank shifts between public and hidden evaluation.}
  Each histogram shows the distribution of rank changes
  $\Delta r = r_{\text{hidden}} - r_{\text{public}}$ per team.
  Planning track: Spearman $\rho = 0.62$ ($p = 0.04$), moderate
  significant alignment with maximum absolute shift of 6 positions.
  Execution track: Spearman $\rho = -0.13$ ($p = 0.71$), near-zero
  non-significant alignment with maximum absolute shift of 8 positions,
  indicating that public execution scores do not predict hidden performance.%
}
\label{fig:rank_shift}
\end{figure}

Figure~\ref{fig:rank_shift} shows the distribution of rank shifts for both
tracks.
In the planning track, the distribution is approximately centred at zero
with Spearman $\rho_{\text{plan}} = 0.62$ ($p = 0.04$), indicating moderate
and statistically significant public-to-hidden alignment: public planning
scores provide a meaningful but imperfect signal of generalization to hidden
planning scenarios.
Maximum absolute shifts reach 6 positions in the planning track,
demonstrating that even in the better-calibrated track, individual teams can
experience substantial reordering.

\begin{figure}[!ht]
\centering
\includegraphics[width=\linewidth]{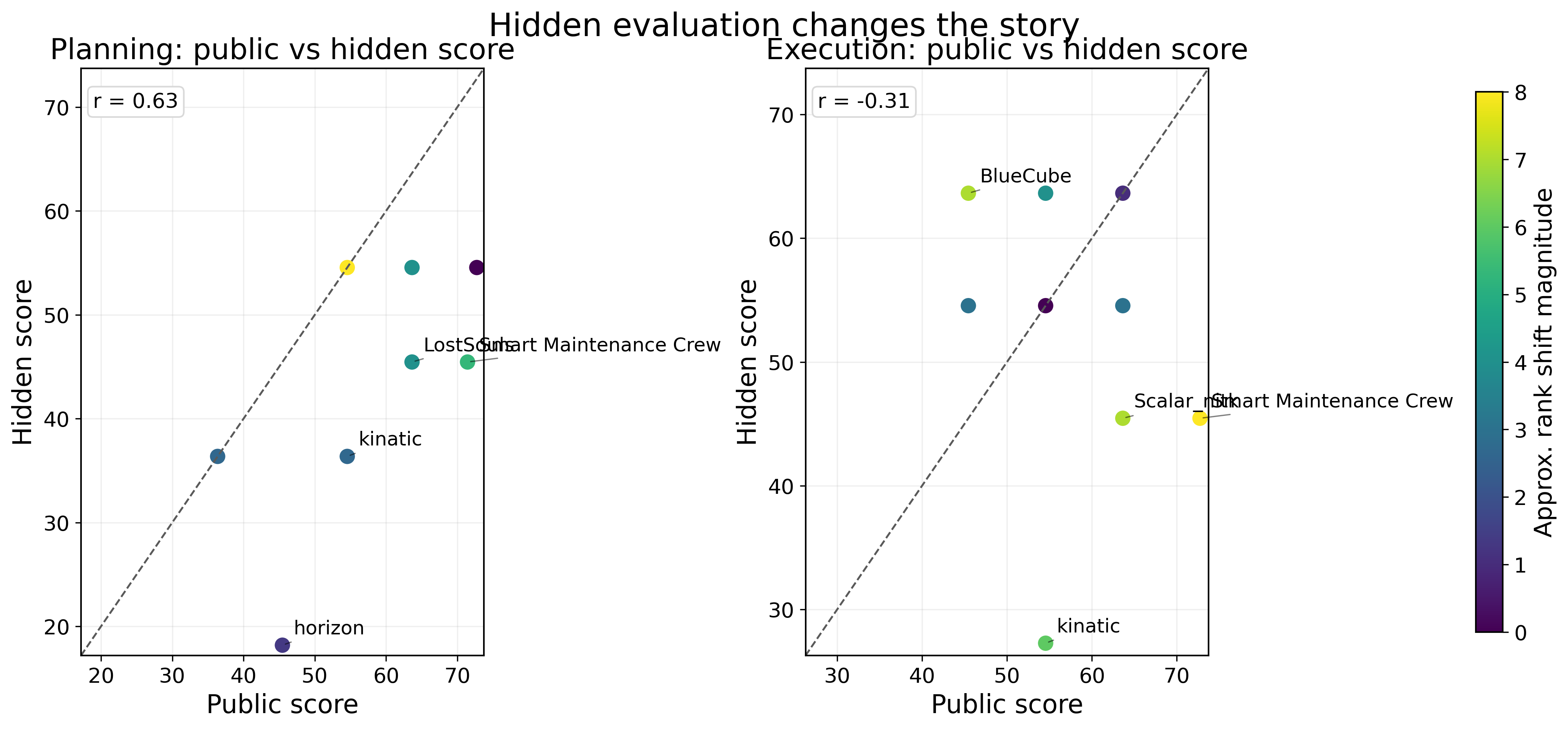}
\caption{%
  \textbf{Public vs.\ hidden score comparison per track.}
  Planning (left panel) shows a recognisable positive relationship between
  public and hidden scores.
  Execution (right panel) shows scatter consistent with near-zero
  correlation, confirming that public execution scores do not provide a
  reliable proxy for hidden performance.
  Shaded regions indicate the 95\% confidence band around the regression
  line where fitted.%
}
\label{fig:pub_priv}
\end{figure}

The execution track tells a sharply different story.
Spearman $\rho_{\text{exec}} = -0.13$ ($p = 0.71$) is near-zero and
non-significant, meaning that public execution scores are essentially
uninformative about hidden execution performance.
Maximum absolute shifts reach 8 positions in the execution track, with the
direction of shifts largely unpredictable from public scores alone.
Figure~\ref{fig:pub_priv} visualises this asymmetry directly through scatter
plots of public versus hidden scores per track: the planning track shows a
recognisable positive relationship, while the execution track scatter is
consistent with random noise.

This finding directly validates the inclusion of a hidden evaluation phase
in \textsc{AssetOpsBench}: without it, the execution-track ranking would
be indistinguishable from a random ordering of agents.
It also motivates a concrete design recommendation for future competitions:
hidden evaluation should be applied to both tracks independently from the
start of the competition rather than as a final-phase reveal, so that the
signal participants optimize against is aligned with the signal that
actually measures generalization.

% =============================================================================
\section{Dimension 5: Computational Footprint}
\label{sec:app:cost}
% =============================================================================

This dimension analyzes the computational cost of running
\textsc{AssetOpsBench}, drawing on 2,196 successful execution traces spanning
22 scenarios (11 development, 11 evaluation) over both phases.
Each trace records tokens sent to the model, tokens received, API calls made,
and wall-clock duration for a single scenario execution.
The cost distribution is heavily right-skewed: the top 10\% of executions
by token consumption account for 42.5\% of all tokens sent, indicating that
a small subset of scenario-strategy combinations drives the majority of
compute expenditure.
A central finding of this dimension is that token expenditure does not
reliably predict task completion score, which has important implications for
how benchmark designers should think about cost-difficulty relationships.

% ─────────────────────────────────────────────────────────────────────────────
\subsection{Overall Cost Distributions}
\label{sec:app:cost_dist}
% ─────────────────────────────────────────────────────────────────────────────

\begin{figure}[h!]
  \centering
  \includegraphics[width=\columnwidth]{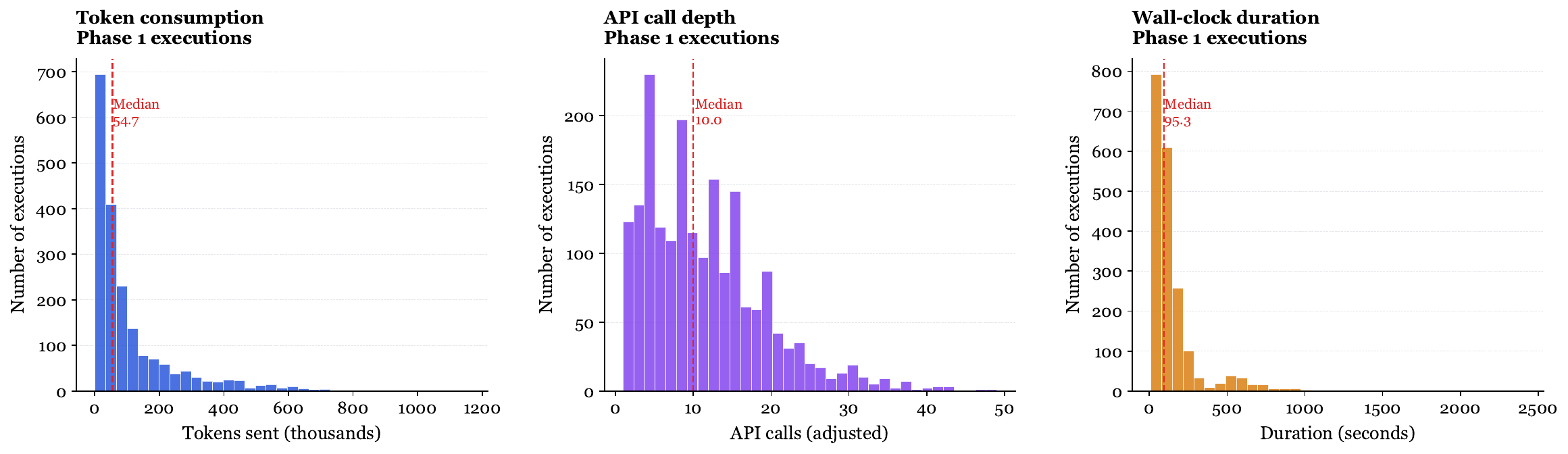}
  \caption{%
    \textbf{Distribution of computational cost metrics across Phase~1
    executions.}
    All three metrics are right-skewed.
    Token consumption: median 54K, mean 110K, CV $= 1.47$, indicating a
    small number of high-token executions dominate aggregate cost.
    API call depth and wall-clock duration are similarly right-skewed but
    with lower dispersion, reflecting structural constraints of the agent
    architecture rather than per-token efficiency differences.%
  }
  \label{fig:cost_distributions}
\end{figure}

Figure~\ref{fig:cost_distributions} shows the full distribution of all three
cost metrics across Phase~1 executions.
All three distributions are right-skewed, sharing the property that the
median is substantially below the mean.
Token consumption has a median of 54K tokens versus a mean of 110K, a ratio
that reflects the presence of a small number of executions with extremely
high token usage.
The coefficient of variation for token consumption is $\text{CV} = 1.47$,
indicating that the standard deviation exceeds the mean, a level of
dispersion that reflects the wide range of computational demands across
the five scenario categories characterised in the domain cost analysis below.
API call depth and wall-clock duration show similar right-skewed patterns
but with lower dispersion, as these metrics are more constrained by
structural properties of the agent architecture than by token efficiency.

% ─────────────────────────────────────────────────────────────────────────────
\subsection{Phase Comparison}
\label{sec:app:phase_comparison}
% ─────────────────────────────────────────────────────────────────────────────

\begin{figure}[h]
  \centering
  \includegraphics[width=\columnwidth]{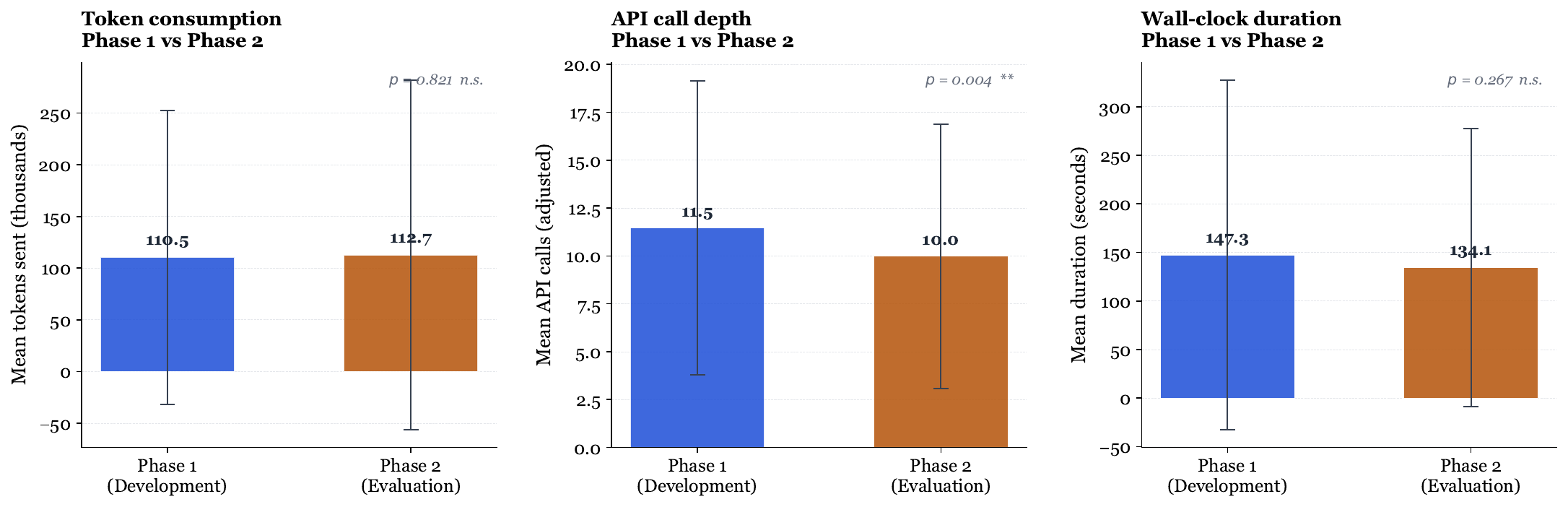}
  \caption{%
    \textbf{Cost comparison between Phase~1 (Development) and Phase~2
    (Evaluation).}
    Error bars denote one standard deviation.
    Token consumption ($p = 0.82$) and wall-clock duration ($p = 0.27$) are
    statistically indistinguishable, confirming comparable computational
    demands across scenario sets.
    API call depth differs significantly ($p = 0.004$): Phase~1 averages
    11.5 calls vs.\ 10.0 in Phase~2, consistent with more exploratory
    multi-step behaviour during development.%
  }
  \label{fig:phase_comparison}
\end{figure}

Figure~\ref{fig:phase_comparison} compares all three cost metrics across the
two evaluation phases.
Token consumption and wall-clock duration are statistically indistinguishable
across phases ($p = 0.82$ and $p = 0.27$ respectively), confirming that the
Phase~1 and Phase~2 scenario sets impose comparable computational demands
despite using different question utterances.
This cross-phase cost consistency is an important benchmark validity property:
it confirms that participants were not systematically advantaged or
disadvantaged by the specific phrasing of the evaluation utterances, and
that the difficulty differences observed in the scenario-level analysis
(Section~\ref{sec:app:difficulty}) reflect intrinsic task structure rather
than utterance-specific properties.
API call depth differs modestly but significantly ($p = 0.004$): Phase~1
executions average 11.5 API calls versus 10.0 in Phase~2.
This asymmetry is consistent with more exploratory multi-step behaviour
during the development phase, where agents are being refined and may attempt
more tool invocations per scenario before converging.

% ─────────────────────────────────────────────────────────────────────────────
\subsection{Agent Domain Cost Profile}
\label{sec:app:domain_cost}
% ─────────────────────────────────────────────────────────────────────────────

\begin{figure}[h]
  \centering
  \includegraphics[width=\columnwidth]{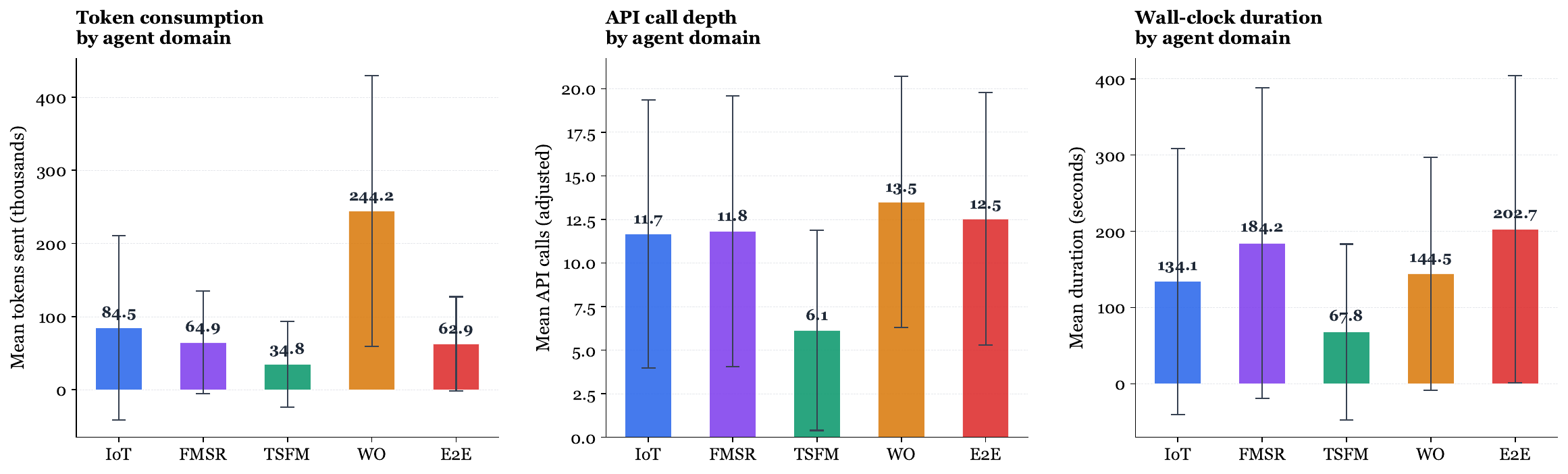}
  \caption{%
    \textbf{Mean execution cost per agent domain across all three metrics.}
    Error bars denote one standard deviation.
    WO is the most expensive domain across token consumption and API calls;
    E2E has the highest wall-clock duration despite moderate token consumption,
    exposing an orchestration latency cost invisible to token-based analysis.
    TSFM is consistently the least expensive domain.%
  }
  \label{fig:agent_domain}
\end{figure}

The five agent domains impose structurally different computational costs,
summarised in Table~\ref{tab:cost_by_domain} and visualised in
Figure~\ref{fig:agent_domain}.
Work Order scenarios are the most expensive by a substantial margin, requiring
a mean of 244K tokens and 13.5 API calls per execution, driven by the
Decision Support query type which demands multi-step retrieval, temporal
aggregation, and constraint reasoning over historical maintenance records.
TSFM scenarios are the least expensive (35K tokens, 6.1 calls), reflecting
the direct lookup nature of capability queries that do not require multi-step
reasoning over operational data.

End-to-end multi-agent (E2E) scenarios occupy a distinctive and analytically
important position in the cost profile.
Despite requiring orchestration across multiple domain agents, their token
consumption (63K) is lower than single-agent WO scenarios.
Yet their wall-clock duration (203 seconds) is the highest of any domain,
substantially exceeding the WO duration (145 seconds).
This dissociation between token count and wall-clock duration exposes an
orchestration latency cost that token counts alone do not capture.
It also means that token expenditure can be an unreliable proxy
for agent complexity in benchmarks that mix single-agent and multi-agent
settings: a naive comparison based on tokens would rank E2E scenarios as
the second-cheapest domain, while a duration-based comparison ranks them
as the most expensive.
Figure~\ref{fig:agent_domain} makes this dissociation visually explicit
across all three cost metrics simultaneously.

\begin{table}[t]
\centering
\caption{Mean execution cost per agent domain across all 2,196 trajectory
files. API calls are adjusted by a factor of $\frac{1}{2}$ to correct for
request--response double-counting in the raw trajectory logs.
E2E denotes end-to-end multi-agent scenarios; all other domains are
single-agent. The \textit{All} row reports unweighted means across all
executions.}
\label{tab:cost_by_domain}
\small
\setlength{\tabcolsep}{6pt}
\renewcommand{\arraystretch}{1.25}
\begin{tabularx}{\columnwidth}{@{} l
    >{\raggedleft\arraybackslash}X
    >{\raggedleft\arraybackslash}X
    >{\raggedleft\arraybackslash}X
    >{\raggedleft\arraybackslash}X @{}}
\toprule
\textsc{Domain}
  & \textsc{Tokens sent (K)}
  & \textsc{API calls}
  & \textsc{Duration (s)}
  & \textsc{Agent class} \\
\midrule
IoT   &  84.5 & 11.7 & 134.1 & Single \\
FMSR  &  64.9 & 11.8 & 184.2 & Single \\
TSFM  &  34.8 &  6.1 &  67.8 & Single \\
WO    & 244.2 & 13.5 & 144.5 & Single \\
E2E   &  62.9 & 12.5 & 202.7 & Multi  \\
\midrule
\textit{All} & \textit{110.7} & \textit{11.3}
             & \textit{145.8} & ---    \\
\bottomrule
\end{tabularx}
\end{table}

% ─────────────────────────────────────────────────────────────────────────────
\subsection{Single-Agent vs.\ Multi-Agent Overhead}
\label{sec:app:single_multi}
% ─────────────────────────────────────────────────────────────────────────────

\begin{figure}[h]
  \centering
  \includegraphics[width=\columnwidth]{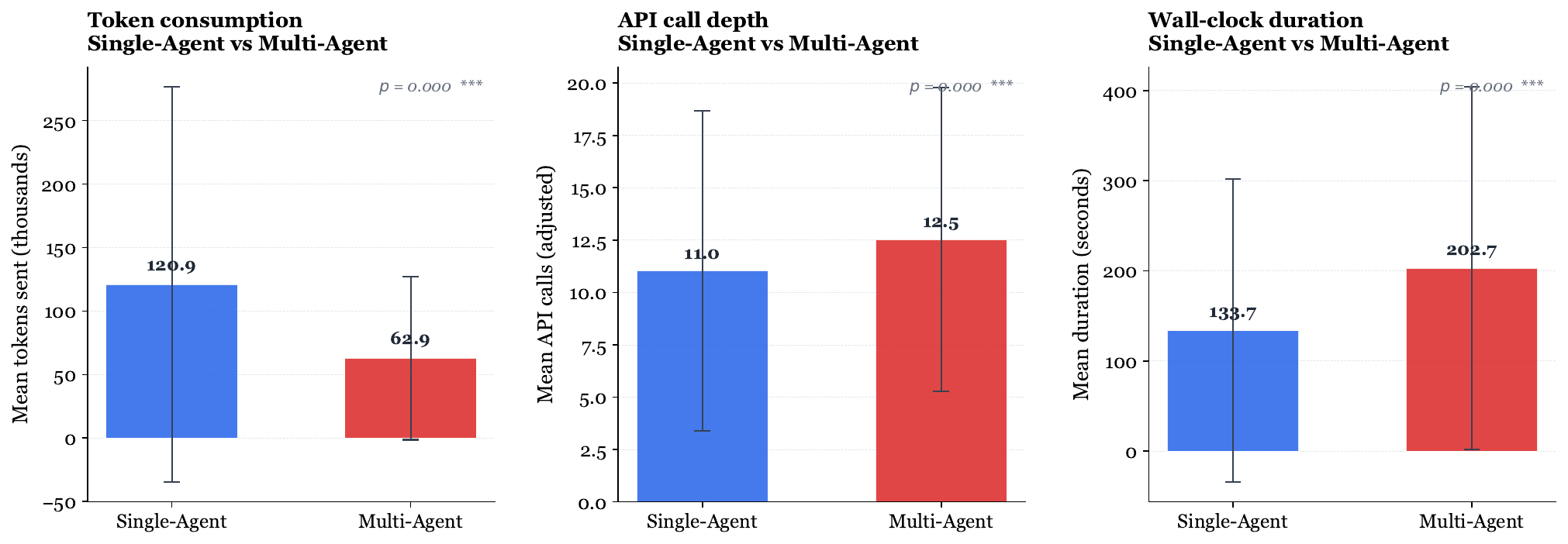}
  \caption{%
    \textbf{Execution cost comparison between single-agent and multi-agent
    (E2E) partitions.}
    Single-agent executions consume nearly twice the tokens of multi-agent
    executions (121K vs.\ 63K, $t = 7.18$, $p < 0.001$), driven by the
    dominance of token-intensive WO Decision Support scenarios in the
    single-agent pool.
    Multi-agent executions have higher wall-clock duration despite lower
    token consumption, reflecting orchestration latency.%
  }
  \label{fig:single_vs_multi}
\end{figure}

Figure~\ref{fig:single_vs_multi} compares execution cost profiles between
the single-agent and multi-agent partitions of the benchmark, revealing a
counterintuitive result.
Single-agent executions consume nearly twice the tokens of multi-agent
executions on average (121K vs.\ 63K, $t = 7.18$, $p < 0.001$).
This inversion is explained by the structural composition of the scenario
pool: the single-agent partition is dominated by WO Decision Support
scenarios, which are the most token-intensive scenario type in the benchmark.
Multi-agent E2E scenarios, by contrast, distribute reasoning across multiple
domain agents, reducing the per-call token load of any individual agent
while increasing total orchestration latency.
The result cautions against a common assumption in benchmark design: that
multi-agent evaluations are inherently more expensive than single-agent
evaluations.
Token count is an unreliable proxy for task complexity in benchmarks that
mix query types across agent boundaries, and duration should be reported
alongside token counts to provide a complete picture of computational
overhead.

% ─────────────────────────────────────────────────────────────────────────────
\subsection{Scenario-Level Difficulty and Cross-Run Variance}
\label{sec:app:difficulty}
% ─────────────────────────────────────────────────────────────────────────────

\begin{figure}[h]
  \centering
  \includegraphics[width=\columnwidth]{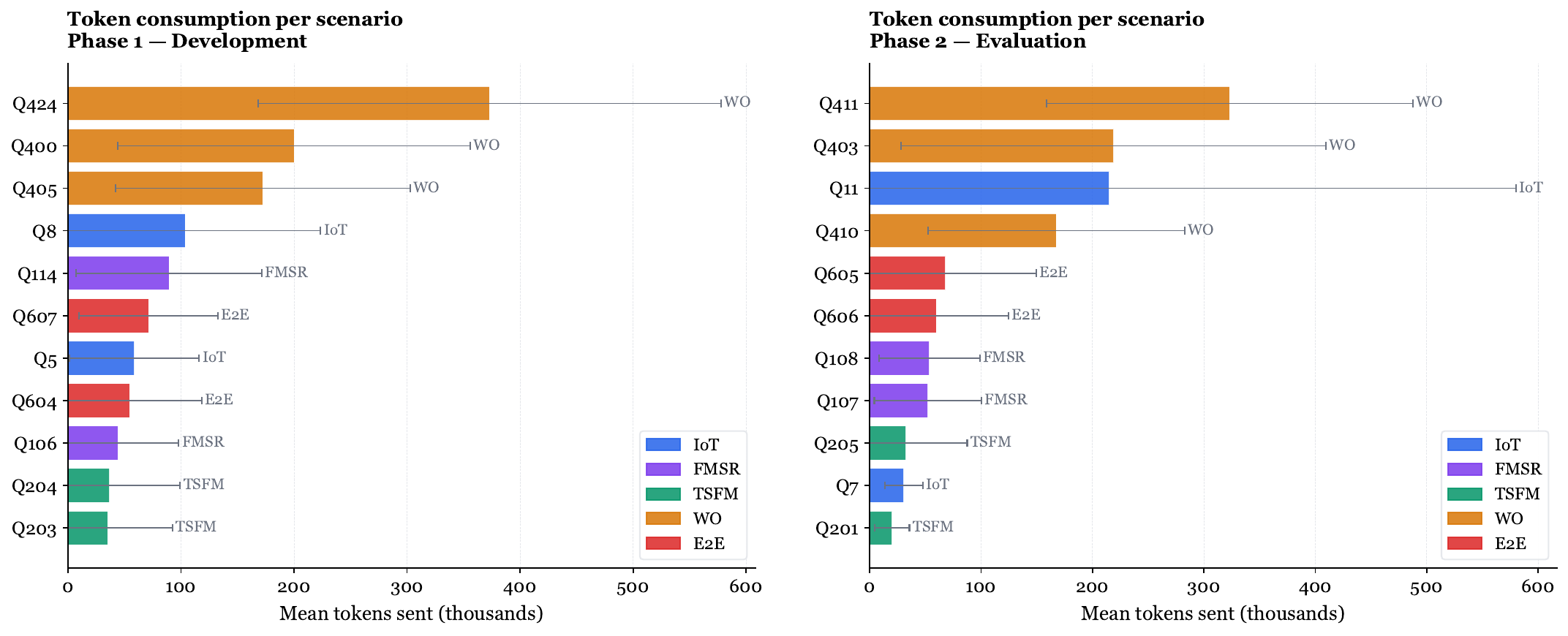}
  \caption{%
    \textbf{Mean token consumption per scenario ordered by difficulty.}
    Left panel: Phase~1 (Development); right panel: Phase~2 (Evaluation).
    Error bars denote one standard deviation across runs.
    Colours indicate agent domain.
    The 18-fold difficulty range between the hardest scenario (Q424, WO,
    373K tokens) and the easiest (Q201, TSFM, 20K tokens) confirms extreme
    difficulty heterogeneity.
    Spearman $\rho = 0.89$ ($p < 0.001$) between phase orderings confirms
    that scenario difficulty is an intrinsic task property, not an artefact
    of participant pool or utterance form.%
  }
  \label{fig:difficulty_cv}
\end{figure}

Figure~\ref{fig:difficulty_cv} shows mean token consumption per scenario
ordered by difficulty within each phase, with error bars indicating
cross-run standard deviation.
The difficulty range across scenarios is extreme: Q424 (WO Decision Support)
requires a mean of 373K tokens per execution, while Q201 (TSFM Knowledge
Query) requires only 20K tokens, an 18-fold difference.
This 18-fold difficulty range has a practical implication for benchmark
design: because all 11 scenarios contribute equally to the task completion
score regardless of their token consumption, the computational cost of
running the benchmark is dominated by the handful of hard scenarios, while
the score signal is distributed equally across all scenarios including the
easy ones.
Future benchmark designs with similar difficulty heterogeneity should
consider weighting the scoring function by scenario difficulty, or reporting
scenario-level pass rates alongside aggregate scores to make the difficulty
structure transparent to participants.

Importantly, the difficulty ordering is highly consistent across Phase~1
and Phase~2 (Spearman $\rho = 0.89$, $p < 0.001$ between the two orderings),
confirming that scenario hardness is an intrinsic property of task structure
rather than an artefact of participant pool composition, submission volume,
or utterance-specific properties of either phase.

\paragraph{Data provenance.}
Table~\ref{tab:fingerprint_provenance} reports the raw execution counts and
cost statistics behind the domain-level cost analysis, providing a complete
data provenance record.
The Phase stability column reports $1 - |1 - \mu_\tau^{P2}/\mu_\tau^{P1}|$
for token consumption, where higher values indicate more consistent costs
across phases.
IoT shows the lowest phase stability (0.52), suggesting that token
consumption in this domain is more sensitive to utterance-level variation
than other domains; this is consistent with the lower inter-phase similarity
of IoT scenario pairs (Table~\ref{tab:scenario_similarity}).
Any reader with access to the released trajectory files can reproduce every
cell in this table by grouping execution records on the domain column and
computing the statistics listed in the column headers.

\begin{table}[h]
\centering
\caption{Data provenance for the computational footprint analysis.
$n$ = total executions in domain; $n_{P1}$ = Phase~1 (development);
$n_{P2}$ = Phase~2 (evaluation).
Tokens (\textit{Tok.}), Calls, and Duration (\textit{Dur.}) are per-execution means.
CV $= \sigma/\mu$ for token consumption (higher = more strategy-sensitive).
Phase stability (\textit{Stab.}) $= 1 - |1 - \mu^{P2}/\mu^{P1}|$ for tokens
(higher = more consistent across phases).}
\label{tab:fingerprint_provenance}
\setlength{\tabcolsep}{6pt}
\renewcommand{\arraystretch}{1.30}
\begin{tabular}{@{} l r r r r r r r r @{}}
\toprule
\textsc{Dom.}
  & \multicolumn{1}{c}{$n$}
  & \multicolumn{1}{c}{$n_{P1}$}
  & \multicolumn{1}{c}{$n_{P2}$}
  & \multicolumn{1}{c}{\textsc{Tok.}}
  & \multicolumn{1}{c}{\textsc{Calls}}
  & \multicolumn{1}{c}{\textsc{Dur.}}
  & \multicolumn{1}{c}{\textsc{CV}}
  & \multicolumn{1}{c}{\textsc{Stab.}} \\
\midrule
WO   & 591   & 522   &  69 & 244K & 13.5 & 145 & 0.76 & 0.96 \\
E2E  & 385   & 339   &  46 &  63K & 12.5 & 203 & 1.02 & 0.98 \\
FMSR & 407   & 362   &  45 &  65K & 11.8 & 184 & 1.09 & 0.80 \\
IoT  & 400   & 356   &  44 &  85K & 11.7 & 134 & 1.49 & 0.52 \\
TSFM & 413   & 368   &  45 &  35K &  6.1 &  68 & 1.68 & 0.74 \\
\midrule
\textit{All}
     & 2{,}196 & 1{,}947 & 249 & 111K & 11.3 & 146 & --- & --- \\
\bottomrule
\end{tabular}
\end{table}

% =============================================================================
% Evaluation results — drop-in LaTeX section
% Requires in preamble:  \usepackage{booktabs}
% =============================================================================

\subsection{Question wise Evaluation Results}
\label{sec:eval-results}

We evaluated the agent system across two phases: a \emph{development}
phase covering 11 curated questions spanning the IoT, FMSR, TSFM, WO, and
E2E categories, and an \emph{evaluation} phase covering 11 held-out
questions from the same categories. Each run is logged in a separate
\texttt{tmp\_*} working directory; results below are aggregated over 244
runs per development question and 35 runs per evaluation question. We
report six metrics per run: \emph{task completion}, \emph{data retrieval
accuracy}, \emph{generalized result verification}, \emph{agent sequence
correctness}, \emph{clarity and justification} (higher is better), and
\emph{hallucinations} (lower is better). To summarize per-question
performance with a single number we compute a \emph{composite score}
defined as the mean of the five higher-is-better metrics minus the
hallucination rate, taking values in $[-1, 1]$.

\subsubsection{Aggregate Performance by Phase}

Table~\ref{tab:phase-summary} reports the mean of each metric within each
phase. Performance on the held-out evaluation questions is comparable to,
and slightly higher than, performance on the development questions across
every higher-is-better metric, while the hallucination rate is meaningfully
lower in the evaluation phase ($0.330$ vs.\ $0.437$). This indicates that
the system generalizes to held-out questions without the degradation that
would suggest overfitting to the development set.

\begin{table}[h]
\centering
\small
\caption{Aggregate metric means by phase. Higher is better for all
metrics except \emph{hallucinations}. \#Runs is the total number of
metric observations (questions $\times$ runs $\times$ metrics).}
\label{tab:phase-summary}
\begin{tabular}{lcccccccc}
\toprule
Phase & Task & Retrieval & Verif. & Seq. & Clarity & Halluc. & \#Runs & \#Qs \\
\midrule
Development & 0.433 & 0.464 & 0.449 & 0.631 & 0.638 & 0.437 & 16{,}050 & 11 \\
Evaluation  & 0.470 & 0.501 & 0.465 & 0.621 & 0.652 & 0.330 & 2{,}310  & 11 \\
\bottomrule
\end{tabular}
\end{table}

\subsubsection{Per-Question Difficulty: Development Phase}

Table~\ref{tab:dev-hardest} reports all 11 development-phase questions
ordered by composite score from hardest to easiest. The three lowest
scores belong to work-order questions: Q424 (corrective work-order
bundling for Chiller~9 over 2017--2019), Q405 (event summary for
CWC04009), and Q400 (work-order retrieval for CWC04013). Each has
near-zero task completion and a hallucination rate above $0.73$. Q8
(IoT bulk download) and Q607 (E2E filtered failure-mode lookup) follow,
while retrieval-style questions Q5, Q203, and Q204, together with the
TSFM model-availability question Q106, achieve composite scores above
$0.48$.

\begin{table}[h]
\centering
\caption{All 11 development-phase questions, ordered by composite score
(mean of five higher-is-better metrics minus hallucination rate). $n$ is
the number of runs per question.}
\label{tab:dev-hardest}
\begin{tabular}{rrrcccccc}
\toprule
QID & $n$ & Comp. & Task & Retr. & Verif. & Seq. & Clar. & Halluc. \\
\midrule
424 & 242 & $-0.764$ & 0.000 & 0.029 & 0.004 & 0.483 & 0.165 & 0.901 \\
405 & 242 & $-0.626$ & 0.000 & 0.017 & 0.041 & 0.091 & 0.376 & 0.731 \\
400 & 243 & $-0.507$ & 0.008 & 0.008 & 0.025 & 0.605 & 0.584 & 0.753 \\
  8 & 244 & $-0.171$ & 0.225 & 0.242 & 0.221 & 0.541 & 0.516 & 0.520 \\
607 & 242 & $\phantom{-}0.050$ & 0.397 & 0.459 & 0.409 & 0.678 & 0.599 & 0.459 \\
114 & 244 & $\phantom{-}0.113$ & 0.447 & 0.488 & 0.545 & 0.639 & 0.639 & 0.439 \\
604 & 242 & $\phantom{-}0.294$ & 0.550 & 0.587 & 0.512 & 0.707 & 0.769 & 0.331 \\
106 & 244 & $\phantom{-}0.481$ & 0.713 & 0.717 & 0.693 & 0.783 & 0.832 & 0.266 \\
203 & 244 & $\phantom{-}0.680$ & 0.791 & 0.885 & 0.820 & 0.770 & 0.832 & 0.139 \\
  5 & 244 & $\phantom{-}0.684$ & 0.811 & 0.824 & 0.820 & 0.836 & 0.848 & 0.143 \\
204 & 244 & $\phantom{-}0.697$ & 0.807 & 0.836 & 0.840 & 0.807 & 0.848 & 0.131 \\
\bottomrule
\end{tabular}
\end{table}

\subsubsection{Per-Question Difficulty: Evaluation Phase}

Table~\ref{tab:eval-hardest} reports the corresponding ranking for the
evaluation phase. Three questions (Q411, Q410, Q403) yield strongly
negative composite scores driven by hallucination rates of $0.80$ or
above and zero performance on the three core correctness metrics.
Q201 represents a distinct failure mode: zero on all higher-is-better
metrics but also zero hallucination, indicating the system declines to
act rather than confabulating an answer (composite score exactly $0.000$).
The remaining seven questions all achieve composite scores above $0.40$,
with Q7 reaching $0.811$.

\begin{table}[h]
\centering
\caption{All 11 evaluation-phase questions, ordered by composite score.}
\label{tab:eval-hardest}
\begin{tabular}{rrrcccccc}
\toprule
QID & $n$ & Comp. & Task & Retr. & Verif. & Seq. & Clar. & Halluc. \\
\midrule
411 & 35 & $-0.691$ & 0.000 & 0.029 & 0.000 & 0.486 & 0.314 & 0.857 \\
410 & 35 & $-0.680$ & 0.000 & 0.000 & 0.000 & 0.200 & 0.400 & 0.800 \\
403 & 35 & $-0.589$ & 0.000 & 0.000 & 0.000 & 0.629 & 0.429 & 0.800 \\
201 & 35 & $\phantom{-}0.000$ & 0.000 & 0.000 & 0.000 & 0.000 & 0.000 & 0.000 \\
107 & 35 & $\phantom{-}0.406$ & 0.657 & 0.657 & 0.543 & 0.771 & 0.829 & 0.286 \\
605 & 35 & $\phantom{-}0.543$ & 0.657 & 0.771 & 0.629 & 0.657 & 0.857 & 0.171 \\
205 & 35 & $\phantom{-}0.571$ & 0.743 & 0.771 & 0.771 & 0.771 & 0.800 & 0.200 \\
108 & 35 & $\phantom{-}0.646$ & 0.771 & 0.800 & 0.800 & 0.743 & 0.829 & 0.143 \\
606 & 35 & $\phantom{-}0.651$ & 0.743 & 0.800 & 0.743 & 0.914 & 0.914 & 0.171 \\
 11 & 35 & $\phantom{-}0.663$ & 0.743 & 0.771 & 0.714 & 0.829 & 0.829 & 0.114 \\
  7 & 35 & $\phantom{-}0.811$ & 0.857 & 0.914 & 0.914 & 0.829 & 0.971 & 0.086 \\
\bottomrule
\end{tabular}
\end{table}

\subsubsection{Discussion}

Three observations stand out across the two phases. First, performance
on the held-out evaluation set tracks development performance closely,
with marginally higher scores on every higher-is-better metric and a
substantially lower hallucination rate (an absolute drop of $0.107$).
We interpret this as evidence that the agent system has not overfit to
the development questions.

Second, \emph{hallucination is concentrated in the lowest-scoring
questions}. In both phases, every question with composite score below
zero exhibits a hallucination rate of at least $0.5$, and every question
with composite score above $0.5$ has a hallucination rate at or below
$0.20$. The Pearson correlation between composite score and
hallucination rate is strongly negative across all 22 questions
($r \approx -0.93$). Hallucination is therefore better understood as a
strong leading indicator of broader task failure than as an independent
failure mode.

Third, \emph{work-order reasoning is the consistent weak spot}. In the
development phase, the three hardest questions (Q424, Q405, Q400) are
all WO tasks involving multi-year event aggregation or bundling. In the
evaluation phase, the three hardest (Q411, Q410, Q403) are also
WO-category questions. This pattern points to long-horizon reasoning
over historical events as the main capability gap; addressing it is
likely to yield the largest single improvement in aggregate performance.

% Optional figures — uncomment if you generate them with a plotting script
% \begin{figure}[h]
%   \centering
%   \includegraphics[width=0.95\linewidth]{phase_means.pdf}
%   \caption{Mean of each metric per phase.}
%   \label{fig:phase-means}
% \end{figure}
%
% \begin{figure}[h]
%   \centering
%   \includegraphics[width=0.95\linewidth]{question_heatmap.pdf}
%   \caption{Per-question metric heatmap across both phases.}
%   \label{fig:question-heatmap}
% \end{figure}

% =============================================================================
\section{Dimension 6: Strategy Attribution}
\label{sec:app:strategy}
% =============================================================================

This dimension analyzes what participants actually implemented and which
design decisions were associated with score differences between submissions.
Unlike the previous dimensions, which derive findings exclusively from
submission logs and trajectory files, this dimension draws on verified
source code for submissions where it was available and on organizer summaries
for the remainder.
Three evidence levels are used throughout: \emph{Verified code} indicates
the source was examined directly; \emph{Organizer summary} indicates the
strategy was reconstructed from the system description provided by the team;
\emph{Template proxy} indicates the submission was assessed against the
public starter template in the absence of accessible source code.

The public-to-private score gap $\delta = \text{score}_{\text{public}} -
\text{score}_{\text{private}}$ serves as a natural quasi-experimental signal
for evaluating strategy quality beyond public-leaderboard optimization.
Strategies that genuinely improve agent generalization should produce small
$\delta$; strategies that overfit to the public evaluation signal should
produce large positive $\delta$.
We are explicit that these attributions are associative rather than causal:
a full ablation study would require re-running submissions with targeted
modifications, which was not feasible at competition scale.
That said, the consistency of the patterns across multiple submissions
provides stronger evidence than any single data point would.

% ─────────────────────────────────────────────────────────────────────────────
\subsection{Method Taxonomy}
\label{sec:app:taxonomy}
% ─────────────────────────────────────────────────────────────────────────────

Table~\ref{tab:appendix_taxonomy} catalogues the main method families
observed across verified submissions, together with their public and private
scores, the $\delta$ gap, evidence level, and main strategy motif.
The table is ordered by public score to make the relationship between public
performance and $\delta$ immediately visible.

\begin{table*}[h]
\centering
\caption{Cross-track method taxonomy with public-private score gap
$\delta = \text{score}_{\text{public}} - \text{score}_{\text{private}}$.
Positive $\delta$ indicates overfitting to the public evaluation signal;
negative $\delta$ indicates a strategy that generalizes better than public
scores suggest.
Entries marked `---' had no private score available.
Ordered by public score descending. Track 1 named 
entries are predominantly verified code (4/6); Track 2 named 
entries are predominantly organizer summary (4/5, with 1 
incomplete archive), reflecting limited source accessibility 
specifically for top Track 2 finishers. The population-level 
Track 2 analysis in Section 3.5 draws on 121 accessible 
source artifacts, of which 119 cluster into four guardrail 
archetypes, independent evidence beyond the named entries 
above.}
\scriptsize
\renewcommand{\arraystretch}{1.25}
\begin{tabularx}{\textwidth}{l r r r c X}
\toprule
\textbf{Submission} & \textbf{Public} & \textbf{Private}
  & $\boldsymbol{\delta}$ & \textbf{Evidence} & \textbf{Main strategy motif} \\
\midrule
\texttt{abc111}
  & 72.73 & 54.55 & $+18.18$
  & Organizer summary
  & Iterative generate--review--rewrite loop with reviewer feedback and
    retry-based refinement at each planning step. \\
\texttt{jainrishi601}
  & 72.73 & 63.64 & $+9.09$
  & Verified code
  & Request typing, richer agent taxonomy, long worked examples in the
    prompt, and parser cleanup to handle malformed outputs. \\
\texttt{kanishk\_007}
  & 72.73 & 54.55 & $+18.18$
  & Verified code
  & Worked examples, fuzzy agent name repair, dependency sanitization,
    plan clamping to the allowed step count, and defensive post-processing. \\
\texttt{radhesham\_95}
  & 72.73 & ---   & ---
  & Verified code
  & Near-template planning prompt with cosmetic edits only; no meaningful
    algorithmic elaboration beyond the public starter. \\
\texttt{rohith\_arumugam}
  & 72.73 & 54.55 & $+18.18$
  & Template proxy
  & Effective baseline-equivalent entry assessed against the public
    starter template; no meaningful algorithmic change identified. \\
\texttt{vamsikv28}
  & 71.43 & 45.45 & $+25.98$
  & Verified code
  & Explicit anti-hallucination instructions and exact tool-usage
    guidance inserted into the planning prompt. \\
\texttt{horizon22}
  & 63.64 & 63.64 & $0.00$
  & Organizer summary
  & Heuristic response scoring with an optional LLM-based cleanup pass
    applied to outputs before submission. \\
\texttt{harshvardhan1}
  & 63.64 & 54.55 & $+9.09$
  & Organizer summary
  & Deterministic keyword-based output ranking and aggressive
    contamination cleanup applied post-generation. \\
\texttt{yasaswinis01}
  & 63.64 & ---   & ---
  & Organizer summary
  & Parallel multi-agent execution with retries and consensus-based
    response selection across parallel outputs. \\
\texttt{samah}
  & 45.45 & 63.64 & $-18.18$
  & Organizer summary
  & Primary-agent execution followed by a fallback-agent path with
    lightweight cross-agent output validation. \\
\texttt{shashank\_1904}
  & 72.73 & 45.45 & $+27.28$
  & Incomplete archive
  & Execution account associated with the officially top-ranked overall team;
    accessible source tree is not treated as the exact evaluated
    implementation given archive incompleteness. \\
\bottomrule
\end{tabularx}
\label{tab:appendix_taxonomy}
\end{table*}

% ─────────────────────────────────────────────────────────────────────────────
\subsection{Planning Prompt Scaffold}
\label{sec:app:scaffold}
% ─────────────────────────────────────────────────────────────────────────────

To make the Track~1 analysis concrete, we include a schematic excerpt
representative of the public starter template and the stable structural
scaffold that all verified planning-track submissions elaborated.
This is not a verbatim reproduction of any single submission; its purpose
is to show the common structural baseline from which teams diverged.

\begin{quote}
\ttfamily
Use only the listed agents. Produce fewer than five plan steps.\\
For each step emit: Task, Agent, Dependency, ExpectedOutput.\\
Do not invent new agents or tools. Keep dependencies explicit.
\end{quote}

Across verified Track~1 submissions, the differences between high-scoring
and lower-scoring strategies lie not in this outer scaffold but in how teams
elaborated it.
The elaborations cluster into five categories: longer agent taxonomies that
reduce agent hallucination risk; worked examples that demonstrate the
expected output format; request typing that categorises input queries before
planning; dependency sanitization that enforces structural validity of the
plan graph; and reviewer-style rewrite loops that iterate over generated
plans before submission.
That many teams achieved the maximum public planning score of 72.73 through
prompt engineering variations alone, without architectural changes, suggests
that the public scaffold was a sufficiently strong baseline to achieve
near-ceiling public performance through surface-level elaboration.
This raises an important question for future benchmark design: if the public
scaffold already achieves near-ceiling scores, what additional challenge
structure would reward substantive implementation changes rather than
surface-level elaboration?

% ─────────────────────────────────────────────────────────────────────────────
\subsection{Strategy-to-Score Attribution}
\label{sec:app:attribution}
% ─────────────────────────────────────────────────────────────────────────────

Examining the $\delta$ column in Table~\ref{tab:appendix_taxonomy} reveals
three patterns that the public leaderboard does not surface and that have
implications for future competition design.

\paragraph{Worked examples show the lowest overfitting among top-scoring strategies.}
The submission most prominently featuring diverse worked examples
(\texttt{jainrishi601}) achieves $\delta = +9.09$, the smallest gap among
submissions that reached the maximum public planning score of 72.73.
This pattern is consistent with the interpretation that worked examples
improve genuine task understanding by demonstrating the expected output
structure rather than suppressing specific error types, making the
improvement robust to utterance-level variation in the hidden scenarios.
In contrast, the three submissions at 72.73 that use other elaboration
strategies (reviewer loops, anti-hallucination rules, or near-template
prompts) all show $\delta = +18.18$, twice the gap of the worked-example
approach.

\paragraph{Reviewer loops and anti-hallucination rules overfit to the public signal.}
Submissions using iterative reviewer--rewrite loops (\texttt{abc111},
$\delta = +18.18$) or explicit anti-hallucination instructions
(\texttt{vamsikv28}, $\delta = +25.98$) achieve high public scores but
experience substantial private score drops.
A plausible mechanism is that these interventions suppress specific error
types that are well-represented in the public scenario set, such as
hallucinated equipment identifiers or incorrect agent names, but introduce
new failure modes on the hidden scenarios where the error distribution
differs.
This pattern illustrates a risk specific to small-scenario-set competitions
with a saturated public evaluation signal: targeted error suppression can
achieve near-ceiling public scores while actively reducing generalization,
because the suppressed error types are not representative of the full error
distribution that the hidden scenarios expose.

\paragraph{Fallback strategies can invert the public ranking.}
The \texttt{samah} submission achieves a public score of 45.45 but a private
score of 63.64, producing $\delta = -18.18$, the only negative gap in the
entire taxonomy.
Under private evaluation, \texttt{samah} would rank above several
submissions that scored 18 points higher on the public leaderboard.
The primary-agent then fallback-agent design sacrifices performance on the
saturated public scenarios, where a rigid single-path primary agent suffices
for most scenarios at the current difficulty level, but benefits from the
greater scenario diversity of the hidden set, where the fallback path
provides a recovery mechanism not available to single-path designs.
This inversion is a direct empirical consequence of the public score
saturation documented in Section~\ref{sec:app:saturation}: when the top of
the public score distribution is saturated, strategies that prioritize
robustness are ranked below strategies that prioritize public-signal
optimization, creating a systematic public--hidden leaderboard misalignment.

\paragraph{Cross-dimensional synthesis.}
The strategy attribution findings connect to findings across all four
preceding dimensions in a coherent picture.
The planning-track score saturation (Dimension~3, Section~\ref{sec:app:saturation})
explains why multiple qualitatively different strategies achieve identical
public scores of 72.73: the binary evaluation cannot resolve the differences
between them.
The near-zero execution-track public-private alignment
(Dimension~3, Section~\ref{sec:app:pub_hidden}) explains why strategies that
overfit to the public planning signal are not penalised publicly but are
severely penalised privately.
The rapid learning curve saturation (Dimension~2, Section~\ref{sec:app:learning})
is explained by the strategy taxonomy: once a team adopts any elaboration
of the public scaffold that avoids the most common failure modes, further
incremental refinements yield diminishing returns on a saturated score scale.
Together, these observations point to a single underlying mechanism: the
public evaluation signal in \textsc{AssetOpsBench}, while informative, was
insufficiently discriminating at the top of the performance distribution to
prevent optimization of non-generalizing strategies.
The hidden evaluation phase was therefore not merely a verification step; in
this competition, it provided the clearest signal of hidden-scenario
robustness and should be designed into the competition from the start rather
than introduced as a final reveal.

% =============================================================================
\section{Dimension 7: Evaluation Agent Robustness}
\label{app:eval-bias}
% =============================================================================
We analyze trajectory-level features to evaluate whether the 
evaluation agent exhibits bias toward specific execution 
patterns such as verbosity, execution length, or tool usage 
complexity.

\paragraph{Setup.}
We use 56 execution trajectories with 32 features derived 
from agent logs, including behavioral features (number of 
steps, tool actions, tool entropy), cost metrics (tokens 
sent, API calls, execution time), and error signals 
(repetition, loops, error rate). Task success is treated as 
the outcome variable.

\paragraph{Method.}
We compare feature distributions across successful and 
failed executions and compute correlations between 
trajectory features and task success. This allows us to 
test whether superficial characteristics of execution 
influence evaluation outcomes.

\begin{figure}[!ht]
\centering
\includegraphics[width=0.99\linewidth]{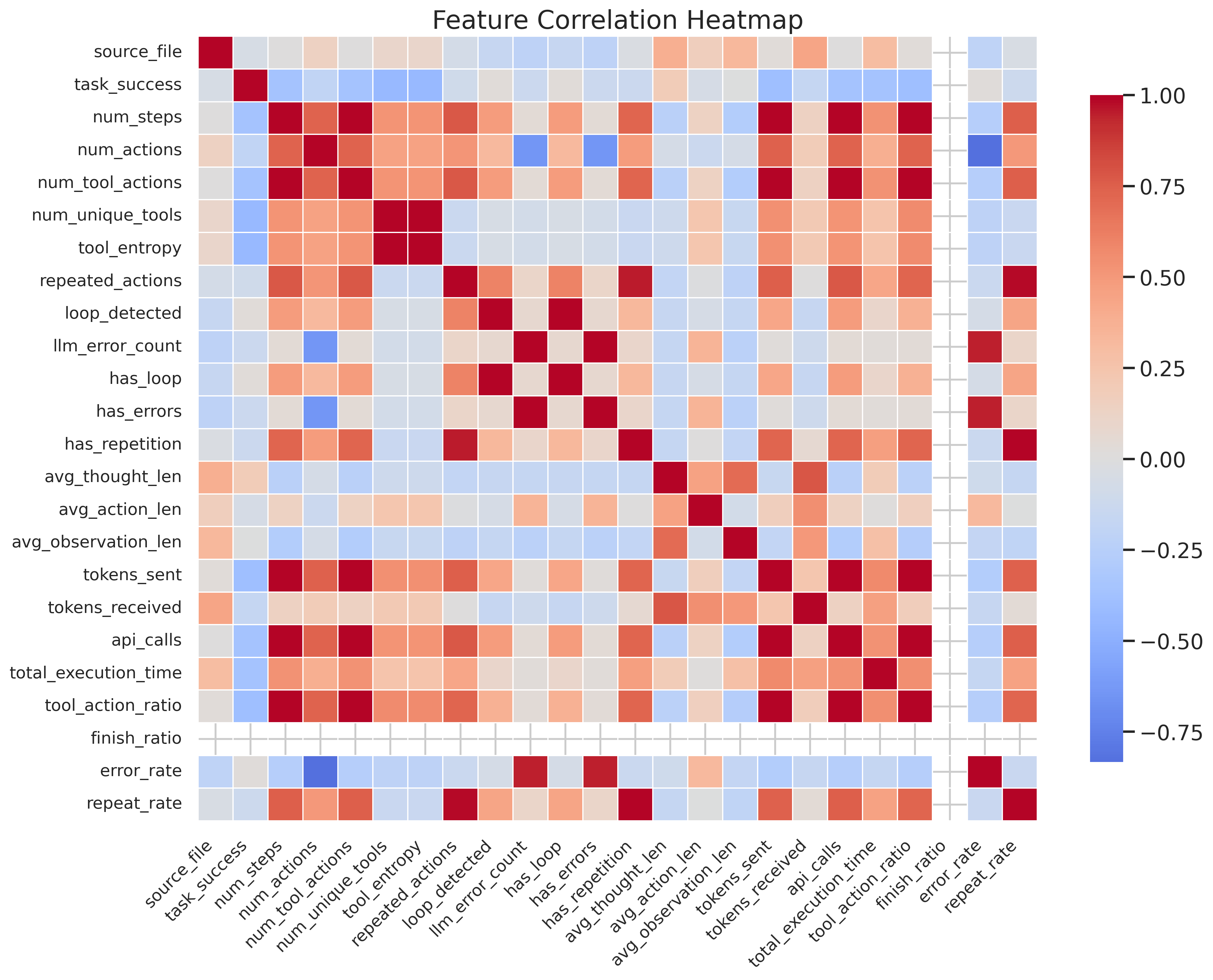}
\caption{Feature correlation heatmap showing relationships 
between trajectory features and task success.}
\label{fig:eval_bias_heatmap}
\end{figure}

\paragraph{Results.}
Figure~\ref{fig:eval_bias_heatmap} shows the correlation 
matrix between trajectory features and task success. Across 
all features, we observe consistent negative correlations 
between success and complexity-related signals. In 
particular, token usage ($r \approx -0.39$), number of 
steps ($r \approx -0.36$), API calls ($r \approx -0.36$), 
and tool entropy ($r \approx -0.43$) are all negatively 
correlated with task success.

\begin{figure}[!ht]
\centering
\includegraphics[width=0.5\linewidth]{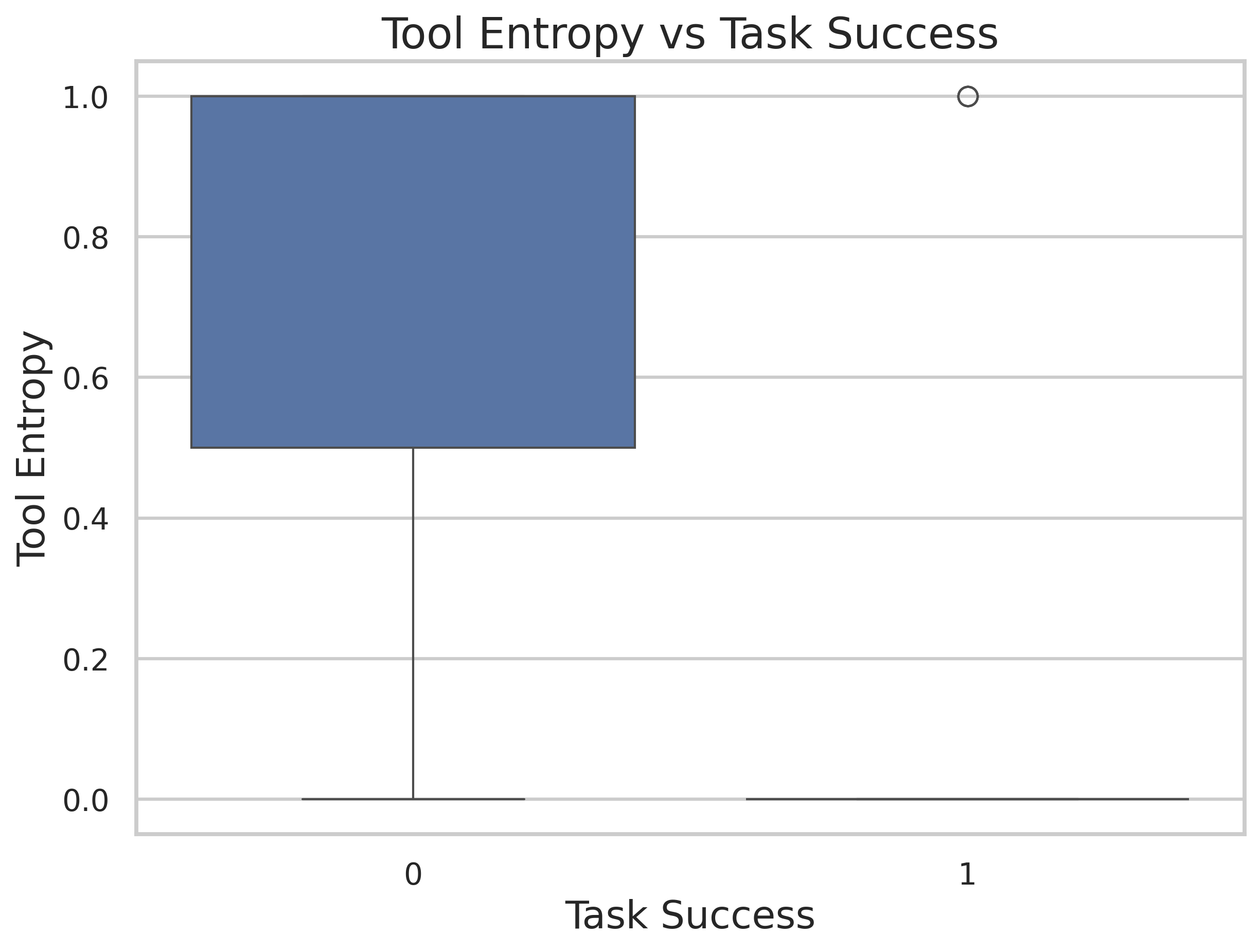}
\caption{Tool entropy vs task success. Successful 
executions exhibit lower entropy.}
\label{fig:entropy_success}
\end{figure}

\begin{figure}[!ht]
\centering
\includegraphics[width=0.5\linewidth]{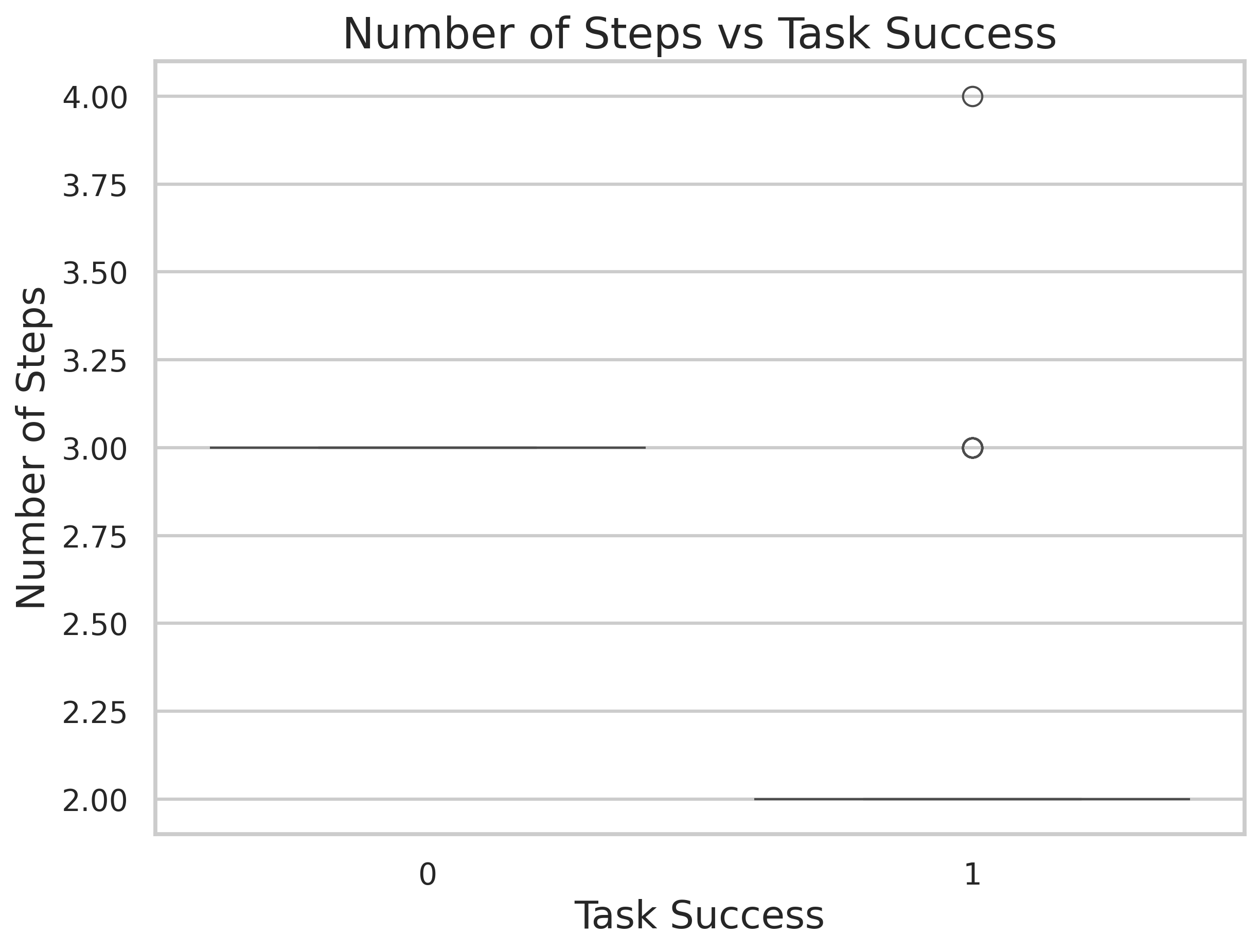}
\caption{Number of steps vs task success. Successful 
executions are shorter.}
\label{fig:steps_success}
\end{figure}

\begin{figure}[!ht]
\centering
\includegraphics[width=0.5\linewidth]{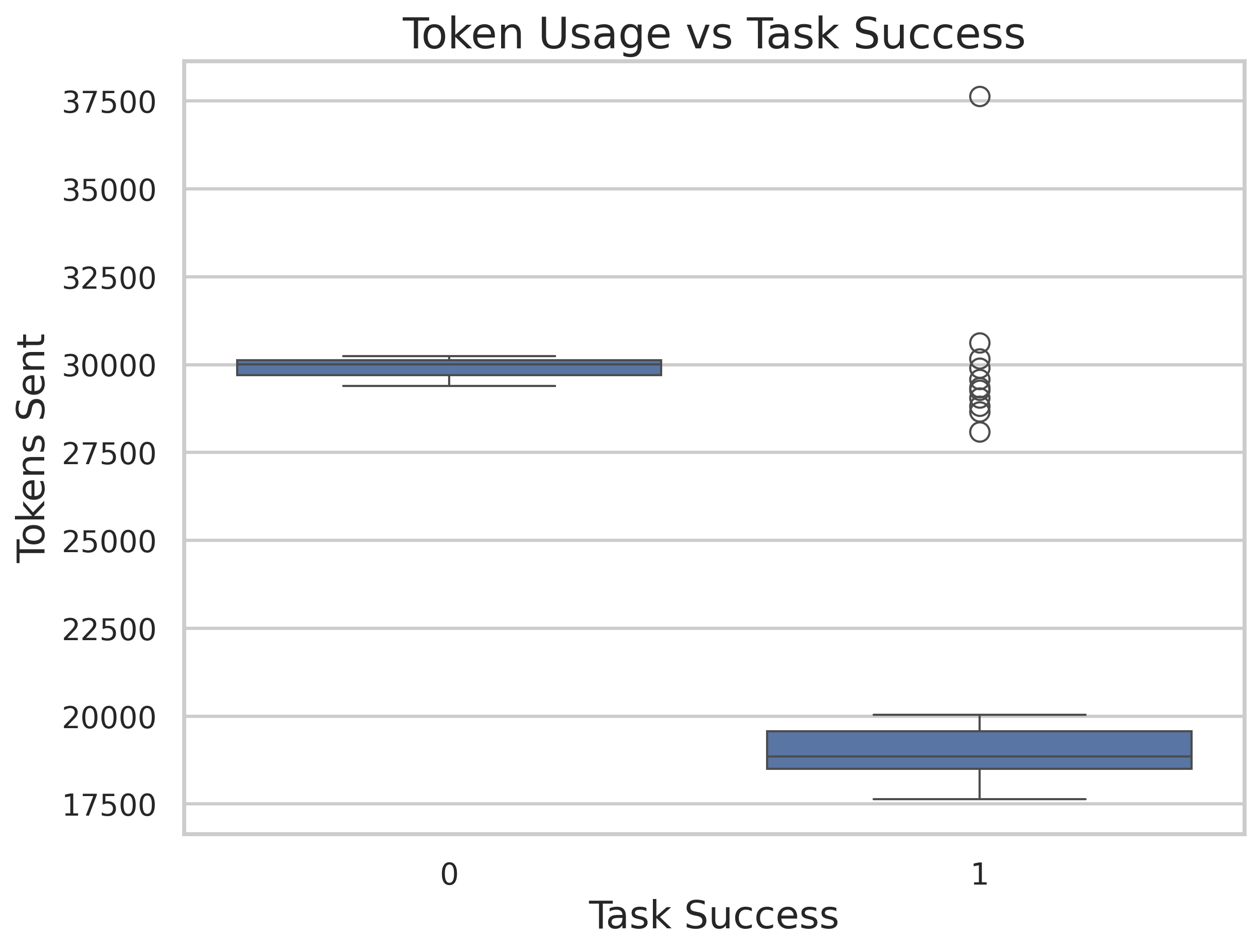}
\caption{Token usage vs task success. Successful executions 
use fewer tokens.}
\label{fig:tokens_success}
\end{figure}

Figures~\ref{fig:entropy_success}, \ref{fig:steps_success}, \ref{fig:tokens_success} 
further illustrate these relationships. Successful 
executions are shorter, use fewer tokens, and exhibit lower 
tool entropy, indicating more focused and structured 
behavior. Failed executions, in contrast, are longer, more 
exploratory, and less consistent.

These results show that the evaluation agent does not 
reward verbosity, execution length, or tool usage 
complexity. Instead, success is associated with concise and 
controlled execution patterns. We find no evidence that 
superficial trajectory characteristics systematically 
influence evaluation outcomes. This analysis confirms that 
the evaluation agent is robust and not biased toward 
specific implementation styles, closing the loop between 
observed evaluation concerns and empirical validation.

% =============================================================================
\section{Dimension 8: Clustering Methodology (Supporting Appendix for Dimension 6)}
\label{app:clustering}
\begin{figure}[t!]
\centering
\includegraphics[width=\textwidth]{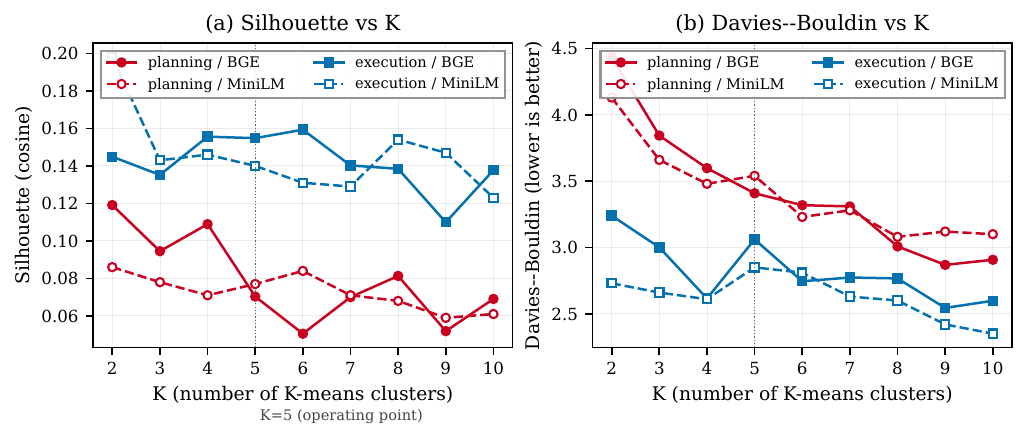}
\caption{K-means cluster-quality sweeps. Solid lines:
\textsc{BGE}. Dashed lines: \textsc{MiniLM}. Execution (blue) has
consistently higher silhouette and lower Davies--Bouldin than
planning (red) at every $K$ and under both encoders. Dotted vertical
line marks the operating point $K{=}5$ used in
Figure~\ref{fig:archetype_bars}.}
\label{fig:kmeans_sweep}
\end{figure}

\begin{figure}[h!]
\centering
\includegraphics[width=\textwidth]{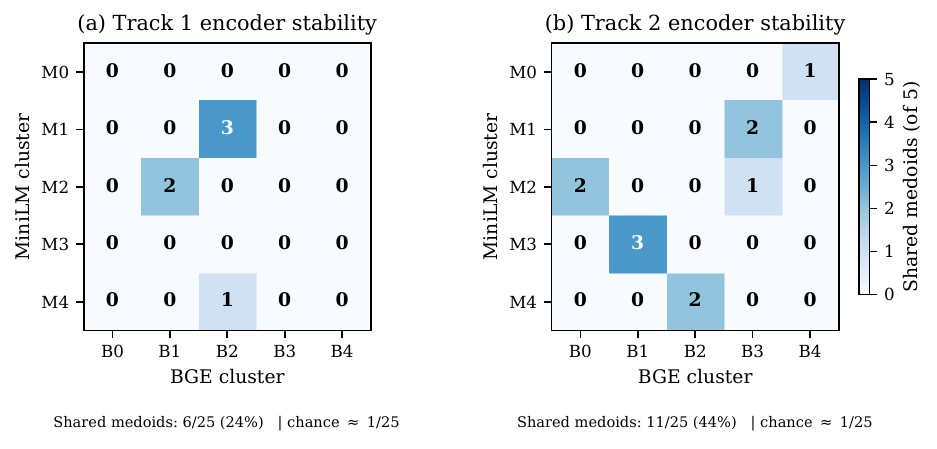}
\caption{Encoder-stability heatmap: shared top-$5$ medoids per
matched cluster pair between \textsc{MiniLM} (rows) and \textsc{BGE}
(columns) at $K{=}5$. Chance overlap under an independent random
assignment is $\approx 1$ medoid per pair. Track~2 (right) has
nearly twice the planning-side stability ($11$ vs.\ $6$
shared medoids).}
\label{fig:medoid_overlap}
\end{figure}

% =============================================================================
Submission-level skill descriptions were generated by prompting a
fixed LLM to read each participant's source and summarise its
strategy along the four planning editable blocks (module variables,
agent descriptions, post-processing, prompt template) or two
execution editable blocks (task-revision helper, dynamic-workflow
run-loop). The resulting Markdown was boilerplate-stripped (shared
section headers, code fences, HTML comments), embedded with both
\textsc{all-MiniLM-L6-v2} (384-dim) and \textsc{BAAI/bge-base-en-v1.5}
(768-dim), L2-normalised, and clustered per track with K-means
($K\!\in\![2,10]$) and HDBSCAN on a 20-dim UMAP of the embeddings.
Cluster quality was evaluated with silhouette (cosine),
Davies--Bouldin, and Calinski--Harabasz indices; per-cluster
interpretation used class-TF--IDF~\citep{grootendorst2022bertopic}
over the boilerplate-stripped text. Encoder stability was quantified as the number of shared medoids (top-$5$
closest-to-centroid submissions) between matched clusters across encoders; matching used Hungarian assignment on the medoid-overlap cost matrix (Figures \ref{fig:kmeans_sweep}, \ref{fig:medoid_overlap}).

We release all intermediate artefacts with this paper: 
per-submission embeddings, K-means sweeps, HDBSCAN labels, 
UMAP projections, medoid manifests, class-TF--IDF top-term 
tables, and canonical composite implementations derived 
from each cluster's medoids.

\section{Dimension 9: Failure Mode Distribution and Taxonomy Analysis of Submissions}
\label{sec:appendix_failure_modes}

We analyzed \textbf{36,884 annotated failure instances} of the submissions, comprising \textbf{999 unique failure-mode titles} and \textbf{3,183 unique descriptions}. The average title length is \textbf{26.06} characters, while descriptions average \textbf{136.37} characters. This indicates a highly diverse but semantically overlapping label space, motivating clustering and taxonomy construction.

\subsection{Frequency Distribution of Failure Modes}
The distribution of failure-mode labels is highly skewed. A small number of dominant failure modes account for a large portion of the dataset, as shown in Table~\ref{tab:top_failure_modes}. We notice that the failures are dominated by issues related to \textit{task completion}, \textit{error handling}, and \textit{context understanding}, rather than domain-specific errors.

\begin{table}[!ht]
\centering
\caption{Top failure-mode labels by frequency.}
\small
\begin{tabular}{l r}
\toprule
\textbf{Failure Mode} & \textbf{Count} \\
\midrule
Lack of Final Answer & 3588 \\
Inadequate Error Handling & 2827 \\
Lack of Contextual Understanding & 1551 \\
Inaction & 1089 \\
Redundant Information Retrieval & 792 \\
Insufficient Data Handling & 605 \\
Lack of Proactive Action & 583 \\
Incomplete Task Execution & 539 \\
Lack of Progression & 418 \\
Insufficient Task Progression & 374 \\
\bottomrule
\end{tabular}
\label{tab:top_failure_modes}
\end{table}

\subsection{Common Patterns in Failure Descriptions}
Frequent phrases extracted from descriptions reveal consistent failure patterns, including,
\textit{``final answer''} (7,181 occurrences)
\textit{``indicating potential''} (3,868)
\textit{``trace shows''} (3,637)
\textit{``failed handle''} (2,882)
\textit{``agent failed''} (2,614). Thus, failures are consistently associated with incomplete execution, lack of convergence, and insufficient robustness, suggesting that errors are primarily \textit{process-level} rather than output-level.

\subsection{Failure Mode Clustering}
\label{sec:clustering}

\begin{figure}[!ht]
    \centering
    \includegraphics[width=\linewidth]{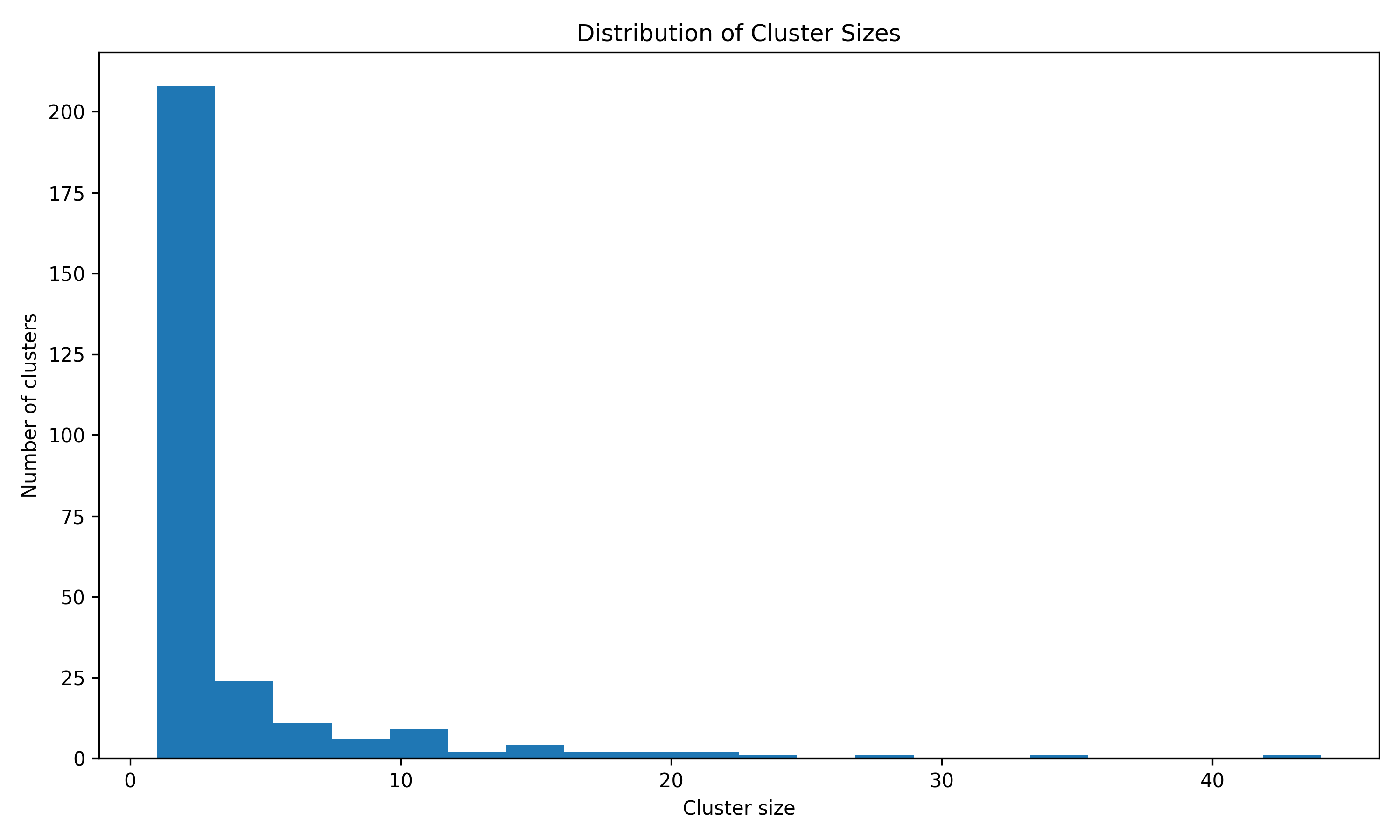}
    \caption{Distribution of cluster sizes across failure-mode labels. Most clusters are small (size 1--3), while a few large clusters capture dominant failure patterns.}
    \label{fig:cluster_distribution}
\end{figure}

To consolidate semantically similar failure-mode labels, we perform unsupervised clustering over the set of unique titles. Each failure-mode title is encoded into a dense vector representation using the Sentence-BERT model \texttt{all-MiniLM-L6-v2}. This model maps semantically similar phrases (e.g., \textit{``Lack of Final Answer''} and \textit{``Missing Final Answer''}) to nearby points in embedding space. We compute pairwise cosine distances between all title embeddings. Cosine distance is chosen due to its effectiveness in capturing semantic similarity in sentence embeddings. We apply agglomerative hierarchical clustering with average linkage. The clustering is performed using a distance threshold of $0.35$, without pre-specifying the number of clusters. Starting from singleton clusters, pairs of clusters are iteratively merged until the inter-cluster distance exceeds the threshold. Clustering results are summarized in Table~\ref{tab:clusters}, which presents the largest clusters along with representative labels and descriptions, and in Figure~\ref{fig:cluster_distribution}, which shows the distribution of cluster sizes.

This approach allows grouping of semantically equivalent or closely related failure modes while preserving fine-grained distinctions between unrelated categories. The use of a distance threshold avoids the need to predefine the number of clusters. The resulting cluster distribution is highly skewed: most clusters are small (size 1--3), while a few large clusters capture dominant failure patterns such as \textit{lack of final answer} and \textit{inadequate error handling}. This suggests that while surface-level label diversity is high, the underlying failure modes are semantically concentrated. The clustering process is sensitive to the chosen distance threshold and embedding model. Additionally, clustering is performed only on titles, which may omit contextual information present in descriptions. Future work could incorporate description-level embeddings or supervised clustering approaches.

\subsection{Failure Mode Taxonomy}

We organize failure modes into a hierarchical taxonomy of parent categories and subcategories.

\subsubsection{Parent Category Distribution}

\begin{table}[!ht]
\centering
\caption{High-level distribution of failure categories.}
\small
\begin{tabular}{l r}
\toprule
\textbf{Parent Category} & \textbf{Relative Frequency} \\
\midrule
Answer Completion / Task Closure & Dominant ($>$75\%) \\
Redundancy / Repetition & Moderate \\
Error Handling / Robustness & Low \\
Context / Memory & Low \\
Data Availability / Access & Sparse \\
Others & Sparse \\
\bottomrule
\end{tabular}
\label{tab:parent_categories}
\end{table}

\paragraph{Key Insight.}
Failure modes are overwhelmingly dominated by \textit{inability to complete tasks}, rather than isolated technical issues.

\subsubsection{Top Subcategories}

\begin{table}[!ht]
\centering
\caption{Representative failure-mode subcategories.}
\small
\begin{tabular}{l p{5.5cm}}
\toprule
\textbf{Subcategory} & \textbf{Description} \\
\midrule
Lack of Final Answer & No conclusive output produced \\
Inadequate Error Handling & Failure to recover from errors \\
Lack of Contextual Understanding & Misinterpretation of context \\
Insufficient Task Progression & Partial execution without completion \\
Redundant Information Retrieval & Repeated retrieval without progress \\
\bottomrule
\end{tabular}
\label{tab:subcategories}
\end{table}

\subsection{Representative Clusters}

The largest clusters highlight dominant failure behaviors, as shown in Table~\ref{tab:clusters}.

\begin{table}[!ht]
\centering
\caption{Largest failure-mode clusters.}
\small
\begin{tabular}{l l r}
\toprule
\textbf{Cluster} & \textbf{Subcategory} & \textbf{Size} \\
\midrule
C7 & Lack of Final Answer & 3885 \\
C8 & Inadequate Error Handling & 3605 \\
C13 & Contextual Understanding & 2101 \\
C0 & Task Completion Indicator & 1807 \\
C2 & Redundant Retrieval & 1342 \\
\bottomrule
\end{tabular}
\label{tab:clusters}
\end{table}

\paragraph{Observation.}
These clusters frequently co-occur, suggesting compound failure modes rather than isolated issues.

\subsection{Key Findings}

\begin{itemize}
    \item \textbf{Task completion is the primary bottleneck.} Most failures involve inability to produce a final answer.
    \item \textbf{Redundancy indicates poor control.} Systems frequently repeat actions without progress.
    \item \textbf{Context handling is insufficient.} Failures often arise from incorrect assumptions about the environment.
    \item \textbf{Error handling lacks robustness.} Systems fail to adapt to missing or inconsistent data.
    \item \textbf{Semantic consolidation is necessary.} Despite 999 unique labels, failures collapse into a small set of core categories.
\end{itemize}

These findings suggest that improving improving system performance primarily requires stronger termination criteria to ensure reliable task completion, along with robust error recovery mechanisms to handle failures effectively. Also, enhancing context tracking and memory consistency, while reducing redundant action loops, is essential for achieving more coherent and efficient execution.

%\clearpage
%\newpage
%\input{checklist.tex}

\end{document}